\documentclass[10pt]{article} 
\usepackage[accepted]{tmlr}

\usepackage{amsmath,amsfonts,bm}

\def\eqref#1{equation~\ref{#1}}

\def\1{\bm{1}}

\def\vs{{\bm{s}}}

\DeclareMathAlphabet{\mathsfit}{\encodingdefault}{\sfdefault}{m}{sl}
\SetMathAlphabet{\mathsfit}{bold}{\encodingdefault}{\sfdefault}{bx}{n}

\newcommand{\R}{\mathbb{R}}

\usepackage[hyperfootnotes=false]{hyperref}
\usepackage{url}
\usepackage[utf8]{inputenc} 
\usepackage[T1]{fontenc}    
\usepackage{booktabs}       
\usepackage{amsfonts}       
\usepackage{nicefrac}       
\usepackage{microtype}      
\usepackage{xcolor}         
\usepackage{graphicx}
\usepackage{wrapfig}
\usepackage{multirow}
\usepackage{epsfig}
\usepackage{amsmath}
\usepackage{amssymb}
\usepackage{subfigure}
\usepackage{marvosym}
\usepackage{soul,color}
\usepackage{enumitem}
\usepackage{threeparttable}
\usepackage{algorithm}
\usepackage{algpseudocode}
\usepackage{setspace}
\usepackage{amsthm}
\usepackage{dutchcal}
\usepackage{makecell}
\usepackage{pifont}
\usepackage{xspace}
\usepackage{colortbl}
\usepackage{siunitx}

\newcommand{\boldres}[1]{\textcolor{red}{\textbf{#1}}}
\newcommand{\secondres}[1]{\textcolor{blue}{\textbf{#1}}}

\usepackage[most]{tcolorbox}

\newtcolorbox{notesbox}{%
  title = Draft,
  colback = yellow!10,   
  colframe = black,      
  fonttitle = \bfseries, 
  sharp corners,
  before skip = 10pt,
  after skip  = 10pt,
}

\definecolor{fbApp}{HTML}{ffe4e3}
\definecolor{tabhighlight}{HTML}{e5e5e5}
\definecolor{pink}{rgb}{1, 0, 0.5}
\definecolor{darkgrey}{rgb}{0.53,0.53,0.53}
\definecolor{mygrey}{rgb}{0.9,0.9,0.9}
\definecolor{purple}{RGB}{230, 227, 254}
\definecolor{lightgreen}{RGB}{238, 252, 241}
\definecolor{lightred}{RGB}{231, 187, 187}
\definecolor{darkred}{RGB}{198, 129, 129}
\definecolor{tabhighlight}{HTML}{e5e5e5}
\definecolor{someorange}{rgb}{0.773,0.353,0.067}
\definecolor{someblue}{rgb}{0.27, 0.35, 0.760}
\definecolor{codegreen}{rgb}{0,0.5,0}
\definecolor{codeblue}{rgb}{0.25,0.5,0.5}
\definecolor{codegray}{rgb}{0.6,0.6,0.6}

\usepackage{lineno}

\title{XCTFormer: Leveraging Cross-Channel and Cross-Time \\Dependencies for Enhanced Time-Series Analysis}

\author{\name Israel Zexer \email zexer@post.bgu.ac.il \\
      \addr The Stein Faculty of Computer and Information Science\\
        Ben-Gurion University of the Negev
      \AND
      \name Omri Azencot \email azencot@bgu.ac.il \\
      \addr The Stein Faculty of Computer and Information Science\\
        Ben-Gurion University of the Negev}

\def\month{03}  
\def\year{2026} 
\def\openreview{\url{https://openreview.net/forum?id=TEfyR4t0Tw}} 

\begin{document}

\maketitle

\begin{abstract}
Multivariate time-series analysis involves extracting informative representations from sequences of multiple interdependent variables, supporting tasks such as forecasting, imputation, and anomaly detection. In real-world scenarios, these variables are typically collected from a shared context or underlying phenomenon, suggesting the presence of latent dependencies across time and channels that can be leveraged to improve performance. However, recent findings show that channel-independent (CI) models, which assume no inter-variable dependencies, often outperform channel-dependent (CD) models that explicitly model such relationships. This surprising result indicates that current CD models may not fully exploit their potential due to limitations in how dependencies are captured. Recent studies have revisited channel dependence modeling with various approaches; however, these methods often employ indirect modeling strategies, which can lead to meaningful dependencies being overlooked. To address this issue, we introduce \textbf{XCTFormer}, a transformer-based channel-dependent (CD) model that explicitly captures cross-temporal and cross-channel dependencies via an enhanced attention mechanism. The model operates in a \emph{token-to-token} fashion, modeling pairwise dependencies between every pair of tokens across time and channels. The architecture comprises (i) a data processing module, (ii) a novel Cross-Relational Attention Block (CRAB) that increases capacity and expressiveness, and (iii) an optional Dependency Compression Plugin (DeCoP) that improves scalability. Through extensive experiments on three time-series benchmarks, we show that \textbf{XCTFormer} achieves strong results compared to widely recognized baselines; in particular, it attains state-of-the-art performance on the imputation task, outperforming the second-best method by an average of 
 20.8\% in MSE and 15.3\% in MAE. Our code is publicly available at \url{https://github.com/azencot-group/XCTFormer}.
\end{abstract}

\section{Introduction}

Forecasting, anomaly detection, and imputation are critical tasks across a wide range of real-world domains \citep{Jin2024GNN4TS}. For instance, forecasting is utilized in energy management, weather prediction, healthcare, and more \citep{mystakidis2024energy, brunet2023advancing, duarte2021comparison}. Time-series analysis plays a vital role in extracting key information from sequential data to facilitate these tasks. The effectiveness of this information extraction is crucial, as it directly impacts the performance of subsequent time-series tasks \citep{trirat2024universal}. Accurate time-series analysis enables organizations to enhance decision-making and optimize resource allocation \citep{bui2018time, wang2024ai}, highlighting the importance of the information extraction component as a key area of research.

Time-series data can be modeled using two main approaches \citep{han2024capacity}. Univariate approaches treat each channel independently, disregarding any potential relationships between them. In contrast, \emph{multivariate} approaches take into account not only the temporal behavior within each channel but also potential dependencies across channels. In real-world scenarios, multivariate datasets are often derived from a common underlying process or phenomenon, which typically leads to dependencies among the features \citep{chen2024structured}. Incorporating relevant signals enhances representation quality and improves accuracy in downstream tasks \citep{isik2025scaling, Domingos2012FewUsefulThings}. As a result, multivariate models are generally expected to outperform univariate models by leveraging both cross-channel dependencies and temporal dynamics. Therefore, time-series analysis can benefit significantly from richer representations when cross-channel dependencies are utilized.

However, recent work in time-series forecasting by \citet{han2024capacity} challenged this assumption by showing that channel-independent (CI) models, which treat multivariate time-series as separate univariate channels and ignore potential inter-channel correlations, outperform channel-dependent (CD) models that explicitly model such dependencies. They attribute this surprising outcome to a trade-off between capacity, defined as a model’s ability to fit complex patterns, and robustness, defined as its ability to remain accurate in the presence of noise, input variation, or distribution shifts. While CD models gain capacity by incorporating cross-channel information, this often comes at the expense of robustness, making them more sensitive to distribution shifts. In contrast, CI models sacrifice some capacity by ignoring cross-channel dependencies, thereby improving robustness and generalization accuracy. These findings challenge the common belief that adding relevant information typically improves representation quality and accuracy, revealing a gap between channel-dependent methods and their unrealized potential. Motivated by these findings, we seek in this paper to address the following question:
\begin{tcolorbox}[notitle, rounded corners, colframe=gray, colback=white, boxrule=2pt, boxsep=0pt, left=0.15cm, right=0.17cm, enhanced, shadow={2.5pt}{-2.5pt}{0pt}{opacity=5,mygrey},toprule=2pt, before skip=0.65em, after skip=0.75em 
  ]
\emph{
  {
    \centering 
  {
    \fontsize{8.5pt}{13.2pt}\selectfont 
    How should we model sequential cross-channel information to realize its potential? 
  }
  \\
  }
  }
\end{tcolorbox}

Recent research has revisited channel dependence with cross-channel modeling approaches that often outperform channel-independent (CI) baselines. iTransformer \citep{liu2023itransformer} targets cross-channel dependencies by treating each channel as a token and applying a Transformer on the token sequence. CrossFormer \citep{zhang2023crossformer} and CARD \citep{wang2024card} address both cross-channel and cross-time relationships, by employing a two-stage pipeline for sequence modeling and channel processing.
Despite recent advancements, most methods model dependencies across different channels and time \emph{indirectly}, thereby potentially overlooking important interactions. Additionally, cross-channel dependencies are often unknown in advance, as the underlying generative process is typically unknown. These dependencies may also change over time \citep{zhao2024rethinking}, raising the need for simultaneous cross-channel cross-time modeling. 
To address these challenges, we propose a \emph{direct} modeling strategy with a \emph{token-to-token} approach that explicitly captures each token’s pairwise dependencies across all channels and time-steps. This potentially minimizes essential information loss associated with existing indirect models. To accomplish this, we introduce \textbf{XCTFormer}, a Transformer-based framework that models all pairwise dependencies directly within a single attention block, token-to-token, effectively identifying the most relevant dependencies for downstream tasks.

The backbone of the \textbf{XCTFormer} consists of three novel components: (i) a data processing unit, (ii) the Cross-Relational Attention Block (CRAB), and (iii) the Dependency Compression Plugin (DeCoP). First, we independently patch each channel and tokenize the data. Next, we flatten the channel and time dimensions, which allows CRAB and DeCoP to capture all pairwise dependencies in a token-to-token manner.
CRAB extends the standard attention block \citep{vaswani2017attention} with two key modifications to improve expressivity and robustness. First, it introduces a learnable, non-boolean masking mechanism that supplements conventional binary masks by weighting dependencies according to their learned importance. This allows the model to focus on the most crucial dependencies for the downstream task. Second, CRAB replaces the standard softmax function with a new normalization technique that retains the properties needed for attention activation \citep{saratchandran2025rethinking} while allowing negative weights. This extension increases the model's expressiveness by enabling it to capture a wider range of relationships, as suggested by \citet{lv2024more}. Lastly, DeCoP is an optional  module that partly modifies CRAB with the addition of a learnable compression mechanism, designed to enhance scalability for datasets with numerous channels. It addresses the memory limitations imposed by the transformer's quadratic attention mechanism. DeCoP compresses the quadratic attention into a linear form while minimizing information loss through a learnable compression transformation. We evaluated XCTFormer against various baseline models on multiple downstream tasks, including forecasting, anomaly detection, and imputation, demonstrating strong results. Our main contributions are:
\begin{enumerate}
    \item We identify a key limitation in the current literature on time-series modeling: while analysis methods have advanced substantially, little emphasis has been placed on explicitly capturing both cross-channel and cross-time dependencies in a unified manner. Most existing approaches either model temporal patterns or inter-channel relations separately, which restricts their ability to exploit the full structure of multivariate time-series data.

    \item To address this gap, we propose XCTFormer, a general-purpose framework that models all pairwise cross-channel and cross-time dependencies directly through token-to-token mappings. XCTFormer integrates two complementary components: (i) CRAB, which enhances expressiveness by learning importance-aware attention masks and allowing signed attention activations, and (ii) DeCoP, which mitigates scalability bottlenecks on high-dimensional data through learnable compression while minimizing information loss.

    \item We evaluate our approach on three core time series tasks, forecasting, anomaly detection, and imputation, obtaining consistent improvements or competitive performance against strong baselines. In particular, we achieve state-of-the-art (SoTA) performance in the imputation task, with average reductions in MSE and MAE of 20.8\% and 15.3\%, respectively. We also observe notable gains in forecasting accuracy and anomaly detection performance.
    Furthermore, we introduce a synthetic dataset, evaluated on the forecasting task, to evaluate our model on capturing cross-variate and cross-time relationships and to test its robustness to spurious correlations (Appendix~\ref{appendix:synthetic_dataset}).

\end{enumerate}

\section{Related Work}
\label{sec:related}

\paragraph{From Classical Methods to Deep Architectures.} Multivariate time-series analysis has progressed from traditional statistical models like ARIMA~\citep{box_jenkins_1970}, which often struggle to capture nonlinear dynamics, to deep neural approaches such as LSTM~\citep{hochreiter1997long} and TCN~\citep{Franceschi2018TCN}. While these deep models improve expressiveness, they may still fall short in modeling very long-range dependencies. More recently, time-series tasks have utilized both simple MLP-based architectures~\citep{zeng2023transformers, wang2024timemixer, nochumsohn2025multi} and Transformer-based models~\citep{zhou2021informer, nie2023time, liu2023itransformer, zhang2023crossformer, wang2024card}. Broadly, these models adopt either a channel-independent (CI) strategy, treating each variable separately, or a channel-dependent (CD) approach that explicitly leverages cross-variable structure. 

\paragraph{Early CD designs: temporal focus with implicit cross-channel modeling.} Early CD models emphasized efficient temporal modeling and attention computation. These methods implicitly incorporated cross-channel information by generating tokens representing all channels at the same or nearby time-steps, typically using 1D convolutions, before applying cross-time attention~\citep{li2019enhancing,zhou2021informer,wu2021autoformer,liu2021pyraformer,zhou2022fedformer}. However, since inter-channel relationships were not explicitly embedded, these approaches failed to fully leverage cross-channel dependencies~\citep{zhang2023crossformer}. Consequently, the attention mechanism struggled to recover missing structure, leading to suboptimal representations~\citep{liu2023itransformer}.

\paragraph{CI baselines and channel as token formulations.} On the CI side, PatchTST partitions each channel into overlapping time patches, treating these patches as tokens. These channel tokens are then passed to a stacked transformer architecture that exclusively models cross-time dependencies~\citep{nie2023time}. Linear models, when applied independently to each channel, have also demonstrated competitive performance~\citep{zeng2023transformers,das2023long}. MTLinear \citep{nochumsohn2025multi} is CI: it first clusters channels and then trains a predictor for each cluster to mitigate conflicts in the multi-task objective, but cross-channel dependencies are not explicitly modeled. To reintroduce cross-channel interactions, iTransformer represents each entire channel as a single token, enabling self-attention to operate across variables~\citep{liu2023itransformer}. LEDDAM ~\citep{leddam} takes a different approach by decomposing each series into trend and seasonal components, processing the seasonal part via parallel cross-channel and cross-time pathways before combining them. However, it still lacks a unified mechanism that jointly models both dimensions within its attention module.

\paragraph{Two-stage explicit cross-time and cross-channel modeling.} CrossFormer addresses the limitations of earlier models by dividing each channel into equal-length segments and embedding these segments individually to better preserve semantic information~\citep{zhang2023crossformer}. This approach, along with CARD~\citep{wang2024card}, utilizes a two-stage attention scheme: first attending along the temporal dimension, then explicitly across channels. While this sequential treatment is effective, it captures cross-channel temporal dependencies only \emph{indirectly}, which may result in limited expressiveness.

\paragraph{Time-series foundation models.} Time-series foundation models (TSFMs) have attracted growing interest as unified architectures for zero-shot and few-shot forecasting across multiple datasets. They are pretrained on diverse time-series corpora to learn general-purpose temporal patterns that transfer across domains. Most TSFMs follow channel-independent designs \citep{timesfm, chronos1, timemoe, tirex, nochumsohn2025super}, handling multivariate inputs by processing each variable independently as a univariate series. This choice improves scalability and helps pretraining remain broadly applicable across datasets with varying numbers and types of variables. But it may fail to fully leverage cross-variable dependencies that are crucial in many real-world multivariate systems. Recent efforts have begun to address this cross-channel challenge. For example, Chronos 2 \citep{chronos2} introduces group attention to share information within sets of related series, while Moirai-1 \citep{moirai1} proposes an any-variate architecture that flattens multivariate time series into a single token sequence, allowing it to handle an arbitrary number of variables and jointly model cross-channel structure. Overall, these works highlight that effectively modeling cross-channel structure remains a key challenge, also within the time-series foundation model paradigm.

\section{Vanilla Transformer Attention} \label{background:attention_vanilla}

To facilitate a clear understanding of our proposed modifications, we first outline the standard transformer attention mechanism~\citep{vaswani2017attention}. Consider an input sequence $X\in\mathbb{R}^{N\times D_i}$, where $N$ denotes the sequence length and $D_i$ the per-token input feature dimension. In our case, the same sequence serves to form queries, keys, and values. The attention block projects $X$ with learnable matrices (with $D_m$ being the per-head attention dimension):
\[
W_q,W_k,W_v\in\mathbb{R}^{D_i\times D_m},\qquad
Q=XW_q,\; K=XW_k,\; V=XW_v,\quad Q,K,V\in\mathbb{R}^{N\times D_m}.
\]

Scaled dot-product scores quantify pairwise query–key affinity:
\[
A=\frac{QK^\top}{\sqrt{D_m}}\in\mathbb{R}^{N\times N}.
\]

An optional mask $M\in\mathbb{R}^{N\times N}$ encodes disallowed positions (e.g., padding or future time-steps) via
\[
M_{ij}=\begin{cases}
0, & \text{allowed}\\
-\infty, & \text{blocked}
\end{cases}
\]

We then convert scores into attention weights row-wise and aggregate values accordingly:
\[
W=\mathrm{Softmax}(A+M)\in\mathbb{R}^{N\times N},\qquad
O=WV\in\mathbb{R}^{N\times D_m}.
\]

Before normalization, the optional mask $M$ is added to the score matrix $A$. Applying a row-wise softmax to $A+M$ effectively assigns zero weight to blocked entries; hence, the mask serves as a selection mechanism that suppresses specific relationships (e.g., to prevent information leakage from future time steps). The resulting attention matrix $W$ is nonnegative with each row summing to one, yielding a probability-like distribution over keys for each query. Consequently, the output $O$ is a row-wise weighted combination of the value vectors, governed by these attention weights, representing the attention block's output.

\section{XCTFormer} 
\label{sec:method}

\begin{figure}[ht]
  \centering
  \includegraphics[page=1,width=\textwidth]{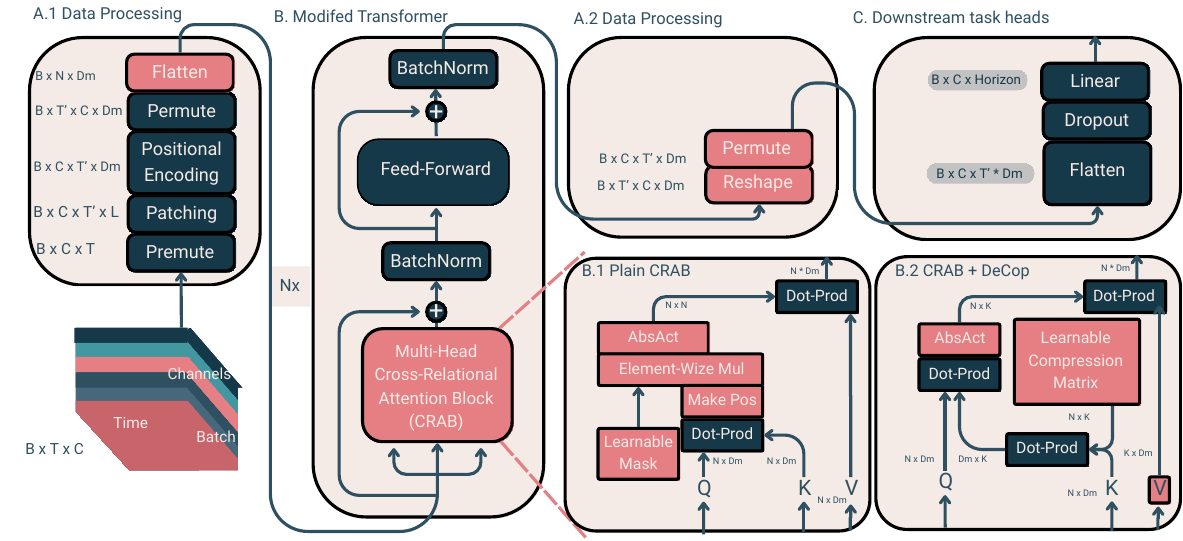}
  \vspace{-5mm}
  \caption{XCTFormer model overview. Multivariate inputs are divided into patches per channel, tokenized, and then passed as a flattened time-and-channel sequence through stacked Transformers with CRAB attention. CRAB utilizes a learnable mask and a signed, non-softmax normalization to model direct token-to-token dependencies. The optional DeCoP  module enhances scalability by compressing pairwise attention into a compact representation, reducing memory and compute requirements; since the compressed attention loses the original pairwise structure, the learnable mask is not applied when DeCoP is used. In the diagram,  panel~B.1 illustrates the CRAB flow and panel~B.2 illustrates the CRAB with DeCoP flow.}
  \label{fig:XCTFormer_Method_Overview}
\end{figure}

To model all pairwise dependencies through \emph{direct token-to-token} modeling, we present \textbf{XCTFormer}, a Transformer-based, general-purpose, encoder-only time-series model. XCTFormer comprises a universal backbone and a task-specific head. The backbone has three components: (i) a data-processing unit, (ii) a Cross-Relational Attention Block (CRAB), and (iii) a Dependency Compression Plugin (DeCoP). Figure~\ref{fig:XCTFormer_Method_Overview} summarizes the pipeline: panel A tokenizes the input and flattens across channels and patches to form a unified sequence that exposes all pairwise dependencies to the Transformer; Panel B applies a stack of our modified Transformer equipped with CRAB (Sec.~\ref{method:CRAB}) and the optional DeCoP module (Sec.~\ref{method:DeCoP}); Panel C maps the resulting representations to predictions via a task-specific head.

\subsection{Data Processing} \label{method:data_processing}

\begin{figure}[t]
  \centering
  \setlength{\abovecaptionskip}{2pt}
  \setlength{\belowcaptionskip}{6pt}
  \includegraphics[width=0.5\textwidth]{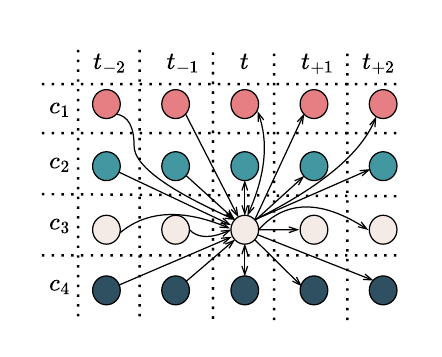}
  \caption{Potential cross-channel and temporal dependencies for token at channel 3 at time $t$.}
  \label{method:all_possible_dependencies}
\end{figure}

To effectively capture the diverse and unknown dependency structures present in multivariate time-series, XCTFormer is designed to explicitly model all pairwise cross-channel and cross-temporal relationships. For each token, we define potential pairwise dependencies across channels and time points throughout the entire time-series. These dependencies can take one of four potential forms: (i) self-lag relationships, where past values of the same channel may influence future states; (ii) cross-channel synchronous relationships, where channels at identical time points may influence one another; (iii) cross-channel lagged relationships, where other channels may exert temporal influence through their historical values; and (iv) forward-in-time relationships, where current values may influence subsequent values within the same or different channels. For visual representation of these dependencies, see Figure \ref{method:all_possible_dependencies}.

Modeling dependencies at the level of individual measurements is both computationally expensive and impractical, as single measurements lack meaning without temporal context \citep{zeng2023transformers}. To address this, we adopt a patching strategy \citep{nie2023time}, segmenting each channel independently into short temporal patches that capture local patterns. We project each patch through a learnable linear layer and add a learnable positional encoding along the time axis for each channel, generating tokens. Finally, we permute the data dimensions so that the patch sequence comes first, then flatten the tokens across the patch and channel dimensions to create a unified sequence. This enables the Transformer to model all pairwise dependencies (see Panel A, Figure~\ref{fig:XCTFormer_Method_Overview}). We apply this permutation to simplify the structure of the attention mask, making it easier to analyze further (see App. \ref{appendix:mask_analysis}).

\subsection{Cross-Relational Attention Block (CRAB)} 
\label{method:CRAB}

CRAB modifies the standard attention~\citep{vaswani2017attention} block with two complementary components designed to increase models' expressivity: (i) a learnable, non-Boolean relational mask designed to learn important dependencies, and (ii) a signed, absolute-sum normalization activation function that replaces the original softmax, designed to increase expressiveness by allowing negative values, inspired by \cite{lv2024more} findings. For a visual representation of the model, refer to Panel B.1, Fig.~\ref{fig:XCTFormer_Method_Overview}.

Our \textbf{learnable non-Boolean mask} is designed to learn the most dominant dependencies between tokens. We apply this mask to a \emph{positive-transformed} attention score matrix. Starting from the score matrix $A\in\mathbb{R}^{N\times N}$, we remove sign information via a global shift,
\[
A_{+}=A-\min(A) \ .
\]
Then, we initialize a learnable real-valued mask $M\in\mathbb{R}^{N\times N}$ with zero mean and standard deviation $\sqrt{2/N}$ following He initialization~\citep{he2015delving}. We apply $M$ in an element-wise fashion:
\[
A \;=\; M \circ A_{+} \ .
\]

Thus, this shift removes sign while preserving the \emph{relative ranking} and pairwise differences of the original scores, while the learnable mask $M$ sets their signs and reweighs magnitudes. The produced output is then passed to the activation function.

Our \textbf{modified attention activation function} replaces softmax with a row-wise normalization that yields \emph{signed} attention weights. Allowing negative weights can increase the model's expressive power~\citep{lv2024more}. To ensure stable training, our proposed activation must preserve the stability property that underlies softmax's success~\citep{saratchandran2025rethinking}: maintaining a bounded Frobenius norm of $\sqrt{N}$ for the produced activation matrix,
\[
    \bigl\|\operatorname{Activation}(A)\bigr\|_F \;\le\; \sqrt{N} \ .
\]

For an attention-score matrix $A \in \mathbb{R}^{N \times N}$, we define our activation function as a normalization of values by the absolute sum of the corresponding row. The AbsAct function is defined as:

\begin{equation}
\text{AbsAct}(A_{ij}) = \frac{A_{ij} + \varepsilon}{\sum_{k=1}^{N} |A_{ik} + \varepsilon| + \delta}, \quad \forall i,j \in \{1,\ldots,N\}
\end{equation}

where $\varepsilon=\num{1e-4}$ and $\delta=\num{1e-8}$ are numerical stabilizers. The parameter $\varepsilon$ shifts each attention score before normalization, while $\delta$ adds a positive margin to ensure the denominator remains non-zero. Our activation function satisfies the bounded-norm constraint, ensuring stable training (see Appendix~\ref{appendix:activation_validity} for a formal proof). Additionally, allowing negative attention weights enables the model to capture a wider range of dependencies, thereby increasing its expressiveness (see Appendix~\ref{appendix:signed_attention_mask_analysis} for additional analysis). Note that since AbsAct divides each entry by its row's absolute-value sum, the standard $1/\sqrt{d_k}$ scaling of attention scores cancels out and is therefore omitted (see Appendix~\ref{appendix:scaling_invariance} for a formal derivation).

\subsection{Dependency Compression Plugin (DeCoP)} 
\label{method:DeCoP}

Since we model all pairwise cross-channel and cross-time dependencies (Figure~\ref{method:all_possible_dependencies}) using the attention mechanism, memory and compute scale as \(\mathcal{O}(N^2)\), where \(N\) is the total number of modeled relations. Applying such attention to datasets with many channels can exceed hardware capacity, potentially leading to running failures. To address this limitation, we introduce \emph{DeCoP}, a plugin which compresses each row of the attention matrix \(A\in\mathbb{R}^{N\times N}\) with a learnable transformation, yielding a compressed matrix \(A_c\in\mathbb{R}^{N\times k}\) with \(k\ll N\). This reduces the dominant cost from quadratic to linear in \(N\) for fixed \(k\), improving scalability regardless of dataset size.  For a visual plot of the model, we refer to Panel B.2, Fig.~\ref{fig:XCTFormer_Method_Overview}. DeCoP is defined as follows:

DeCoP introduces a learnable compression transform whose parameters are initialized with He initialization~\citep{he2015delving} and fine-tuned during training. Our compressor is a matrix, $C \in \mathbb{R}^{N\times k}$ such that $k\ll N$. Let \(Q=XW_q,\ K=XW_k\) with \(Q,K\in\mathbb{R}^{N\times D_m}\). We utilize $C$ in the attention computation as follows:
\[
    A_c = Q\,(K^{\top}C)\in\mathbb{R}^{N\times k} \ .
\]

Due to the associative property of matrix multiplication, computing $Q(K^{\top}C)$ is equivalent to computing $(QK^{\top})C$, enabling us to obtain a compressed version of the full token-to-token attention without materializing the quadratic $N \times N$ attention matrix. While vanilla attention incurs $O(N^2 D_m)$ cost for computing $QK^\top$, DeCoP's reordered computation achieves $O(N D_m k)$ complexity, scaling linearly in $N$ when $k\ll N$ while preserving essential attention relationships through the compressed representation. For a full complexity analysis, see Appendix~\ref{appendix:DeCoP_complexity_analysis}.

Finally, we also need to modify $V$ calculations as the attention dimension is reduced to $k$. The modification is defined as follows:
\[
    W_v\in\mathbb{R}^{k\times N} \ , \qquad V=W_vX,\quad V\in\mathbb{R}^{k\times D_m} \ .
\]
The new $V$ represents the values corresponding to the compressed attention dependencies. The final output is calculated as follows:
\[
    W_c=\mathrm{AbsAct}(A_c)\in\mathbb{R}^{N\times k} \ , \qquad O = W_c V\in\mathbb{R}^{N\times D_m} \ .
\]

Note that when DeCoP is applied, the learnable mask $M$ from CRAB (Sec.~\ref{method:CRAB}) is not used. Since the compressed attention $A_c \in \mathbb{R}^{N \times k}$ does not preserve the direct pairwise structure of the full $N \times N$ matrix, element-wise masking of individual token-to-token relationships is no longer applicable; instead, AbsAct is applied directly to the compressed scores $A_c$.

\section{Experiments}
We evaluate the proposed XCTFormer across three fundamental time-series tasks: long-term forecasting, imputation, and anomaly detection. Our experiments use well-established benchmark datasets commonly used in prior work to ensure a fair comparison with existing approaches. Across all experiments, we apply DeCoP (Sec.~\ref{method:DeCoP}) to datasets with more than 60 channels; datasets with 60 or fewer channels use the plain CRAB module without DeCoP (Sec.~\ref{method:CRAB}). For each task, we present the experimental setup and datasets and report comparative results against strong baselines. The following subsections detail our experiments for each time-series task. App.~\ref{appendix:experiments_detail} provides formal task formulations, extended training and evaluation details, including hyperparameter search protocol and values.

\subsection{Long-Term Forecasting}

Time-series forecasting aims to predict future values from historical observations. We evaluate our model on seven widely used multivariate datasets, comprising four ETT subsets (ETTm1, ETTm2, ETTh1, ETTh2), Weather, Electricity (ECL), and Traffic, following Autoformer \citep{wu2021autoformer}. We adopt the TimesNet setup \citep{wu2022timesnet} with a lookback window of 96 time-steps and forecasting horizons \( \{96, 192, 336, 720\} \). We compare against twelve widely recognized forecasting models:
(i) Transformer-based: Autoformer \citep{wu2021autoformer}, FEDformer \citep{zhou2022fedformer}, Crossformer \citep{zhang2023crossformer}, PatchTST \citep{nie2023time}, iTransformer \citep{liu2023itransformer};
(ii) Linear/MLP-based: DLinear \citep{zeng2023transformers}, TiDE \citep{das2023long}, TimeMixer \citep{wang2024timemixer}, MTLinear \citep{nochumsohn2025multi};
(iii) Hybrid Transformer and Linear: LeDDAM \citep{leddam};
(iv) TCN-based: SCINet \citep{liu2022scinet}, TimesNet \citep{wu2022timesnet}.
We also include a synthetic dataset whose target channel is constructed from lagged cross-variate signals. The dataset also includes distractor channels (random walks) to test robustness to spurious correlations. Models are evaluated in the multivariate-to-single (MS) setting, which allows to specifically test each model's capacity to capture cross-channel and cross-time dependencies. To ensure a fair comparison, all models are trained using the same hyperparameters as ETTm1 (Appendix~\ref{appendix:synthetic_dataset}).

\begin{table}[htbp]
  \caption{Average long-term forecasting results comparison. Synthetic$^\dagger$ is evaluated under multivariate-to-single (MS) setting while all other datasets use multivariate (M). We compare extensive competitive models under different prediction lengths. \emph{Avg} is averaged from all four prediction lengths, that $\{96, 192, 336, 720\} $.}\label{tab:avg_forecasting_results}
  \vskip 0.05in
  \centering
  \resizebox{1.0\columnwidth}{!}{
  \begin{threeparttable}
  \begin{small}
  \renewcommand{\multirowsetup}{\centering}
  \setlength{\tabcolsep}{1pt}
  \begin{tabular}{c|cc|cc|cc|cc|cc|cc|cc|cc|cc|cc|cc|cc|cc|cc|}
    \toprule
    \multicolumn{1}{c}{\multirow{2}{*}{Models}} &
    \multicolumn{2}{c}{\rotatebox{0}{\scalebox{0.8}{\textbf{XCTFormer}}}} &
    \multicolumn{2}{c}{\rotatebox{0}{\scalebox{0.8}{TimeMixer++}}} &
    \multicolumn{2}{c}{\rotatebox{0}{\scalebox{0.8}{MTLinear}}} &
    \multicolumn{2}{c}{\rotatebox{0}{\scalebox{0.8}{Leddam}}} &
    \multicolumn{2}{c}{\rotatebox{0}{\scalebox{0.8}{TimeMixer}}} &
    \multicolumn{2}{c}{\rotatebox{0}{\scalebox{0.8}{iTransformer}}} &
    \multicolumn{2}{c}{\rotatebox{0}{\scalebox{0.8}{PatchTST}}} &
    \multicolumn{2}{c}{\rotatebox{0}{\scalebox{0.8}{Crossformer}}} &
    \multicolumn{2}{c}{\rotatebox{0}{\scalebox{0.8}{TiDE}}} &
    \multicolumn{2}{c}{\rotatebox{0}{\scalebox{0.8}{TimesNet}}} &
    \multicolumn{2}{c}{\rotatebox{0}{\scalebox{0.8}{DLinear}}} &
    \multicolumn{2}{c}{\rotatebox{0}{\scalebox{0.8}{SCINet}}} &
    \multicolumn{2}{c}{\rotatebox{0}{\scalebox{0.8}{FEDformer}}} &
    \multicolumn{2}{c}{\rotatebox{0}{\scalebox{0.8}{Autoformer}}} \\
    \multicolumn{1}{c}{} &
    \multicolumn{2}{c}{\scalebox{0.8}{(\textbf{Ours})}} &
    \multicolumn{2}{c}{\scalebox{0.8}{(ICLR 2025)}} &
    \multicolumn{2}{c}{\scalebox{0.8}{(AISTATS 2025)}} &
    \multicolumn{2}{c}{\scalebox{0.8}{(ICML 2024)}} &
    \multicolumn{2}{c}{\scalebox{0.8}{(ICLR 2024)}} &
    \multicolumn{2}{c}{\scalebox{0.8}{(ICLR 2024)}} &
    \multicolumn{2}{c}{\scalebox{0.8}{(ICLR 2023)}} &
    \multicolumn{2}{c}{\scalebox{0.8}{(ICLR 2023)}} &
    \multicolumn{2}{c}{\scalebox{0.8}{(TMLR 2023)}} &
    \multicolumn{2}{c}{\scalebox{0.8}{(ICLR 2023)}} &
    \multicolumn{2}{c}{\scalebox{0.8}{(AAAI 2023)}} &
    \multicolumn{2}{c}{\scalebox{0.8}{(NeurIPS 2022)}} &
    \multicolumn{2}{c}{\scalebox{0.8}{(ICML 2022)}} &
    \multicolumn{2}{c}{\scalebox{0.8}{(NeurIPS 2021)}} \\
    
    \cmidrule(lr){2-3} \cmidrule(lr){4-5} \cmidrule(lr){6-7} \cmidrule(lr){8-9} \cmidrule(lr){10-11} \cmidrule(lr){12-13} \cmidrule(lr){14-15} \cmidrule(lr){16-17} \cmidrule(lr){18-19} \cmidrule(lr){20-21} \cmidrule(lr){22-23} \cmidrule(lr){24-25} \cmidrule(lr){26-27} \cmidrule(lr){28-29}
    \multicolumn{1}{c}{Metric} & \scalebox{0.68}{MSE} & \scalebox{0.68}{MAE} & \scalebox{0.68}{MSE} & \scalebox{0.68}{MAE} & \scalebox{0.68}{MSE} & \scalebox{0.68}{MAE} & \scalebox{0.68}{MSE} & \scalebox{0.68}{MAE} & \scalebox{0.68}{MSE} & \scalebox{0.68}{MAE} & \scalebox{0.68}{MSE} & \scalebox{0.68}{MAE} & \scalebox{0.68}{MSE} & \scalebox{0.68}{MAE} & \scalebox{0.68}{MSE} & \scalebox{0.68}{MAE} & \scalebox{0.68}{MSE} & \scalebox{0.68}{MAE} & \scalebox{0.68}{MSE} & \scalebox{0.68}{MAE} & \scalebox{0.68}{MSE} & \scalebox{0.68}{MAE} & \scalebox{0.68}{MSE} & \scalebox{0.68}{MAE} & \scalebox{0.68}{MSE} & \scalebox{0.68}{MAE} & \scalebox{0.68}{MSE} & \scalebox{0.68}{MAE}\\
    \toprule
    \scalebox{0.85}{ETT(Avg)} & \secondres{\scalebox{0.85}{0.364}} & \secondres{\scalebox{0.85}{0.386}} & \boldres{\scalebox{0.85}{0.348}} & \boldres{\scalebox{0.85}{0.377}} & \scalebox{0.85}{0.373} & \scalebox{0.85}{0.387} & \scalebox{0.85}{0.367} & \scalebox{0.85}{0.387} & \scalebox{0.85}{0.367} & \scalebox{0.85}{0.389} & \scalebox{0.85}{0.384} & \scalebox{0.85}{0.404} & \scalebox{0.85}{0.397} & \scalebox{0.85}{0.406} & \scalebox{0.85}{0.685} & \scalebox{0.85}{0.578} & \scalebox{0.85}{0.482} & \scalebox{0.85}{0.470} & \scalebox{0.85}{0.391} & \scalebox{0.85}{0.404} & \scalebox{0.85}{0.446} & \scalebox{0.85}{0.447} & \scalebox{0.85}{0.689} & \scalebox{0.85}{0.597} & \scalebox{0.85}{0.421} & \scalebox{0.85}{0.433} & \scalebox{0.85}{0.465} & \scalebox{0.85}{0.459} \\
    \midrule
    \scalebox{0.85}{Weather} & \secondres{\scalebox{0.85}{0.237}} & \secondres{\scalebox{0.85}{0.267}} & \boldres{\scalebox{0.85}{0.226}} & \boldres{\scalebox{0.85}{0.262}} & \scalebox{0.85}{0.238} & \scalebox{0.85}{0.276} & \scalebox{0.85}{0.242} & \scalebox{0.85}{0.272} & \scalebox{0.85}{0.240} & \scalebox{0.85}{0.272} & \scalebox{0.85}{0.258} & \scalebox{0.85}{0.278} & \scalebox{0.85}{0.265} & \scalebox{0.85}{0.285} & \scalebox{0.85}{0.264} & \scalebox{0.85}{0.320} & \scalebox{0.85}{0.270} & \scalebox{0.85}{0.320} & \scalebox{0.85}{0.259} & \scalebox{0.85}{0.286} & \scalebox{0.85}{0.265} & \scalebox{0.85}{0.315} & \scalebox{0.85}{0.292} & \scalebox{0.85}{0.363} & \scalebox{0.85}{0.309} & \scalebox{0.85}{0.360} & \scalebox{0.85}{0.338} & \scalebox{0.85}{0.382} \\
    \midrule
    \scalebox{0.85}{ECL\tnote{p}} & \secondres{\scalebox{0.85}{0.166}} & \secondres{\scalebox{0.85}{0.263}} & \boldres{\scalebox{0.85}{0.165}} & \boldres{\scalebox{0.85}{0.253}} & \scalebox{0.85}{0.198} & \scalebox{0.85}{0.283} & \scalebox{0.85}{0.168} & \secondres{\scalebox{0.85}{0.263}} & \scalebox{0.85}{0.182} & \scalebox{0.85}{0.273} & \scalebox{0.85}{0.178} & \scalebox{0.85}{0.270} & \scalebox{0.85}{0.216} & \scalebox{0.85}{0.318} & \scalebox{0.85}{0.244} & \scalebox{0.85}{0.334} & \scalebox{0.85}{0.252} & \scalebox{0.85}{0.344} & \scalebox{0.85}{0.193} & \scalebox{0.85}{0.304} & \scalebox{0.85}{0.225} & \scalebox{0.85}{0.319} & \scalebox{0.85}{0.268} & \scalebox{0.85}{0.365} & \scalebox{0.85}{0.213} & \scalebox{0.85}{0.327} & \scalebox{0.85}{0.227} & \scalebox{0.85}{0.364} \\
    \midrule
    \scalebox{0.85}{Traffic\tnote{p}} & \scalebox{0.85}{0.435} & \scalebox{0.85}{0.287} & \boldres{\scalebox{0.85}{0.416}} & \boldres{\scalebox{0.85}{0.264}} & \scalebox{0.85}{0.621} & \scalebox{0.85}{0.372} & \scalebox{0.85}{0.467} & \scalebox{0.85}{0.294} & \scalebox{0.85}{0.485} & \scalebox{0.85}{0.297} & \secondres{\scalebox{0.85}{0.428}} & \secondres{\scalebox{0.85}{0.282}} & \scalebox{0.85}{0.529} & \scalebox{0.85}{0.341} & \scalebox{0.85}{0.667} & \scalebox{0.85}{0.426} & \scalebox{0.85}{0.760} & \scalebox{0.85}{0.473} & \scalebox{0.85}{0.620} & \scalebox{0.85}{0.336} & \scalebox{0.85}{0.625} & \scalebox{0.85}{0.383} & \scalebox{0.85}{0.804} & \scalebox{0.85}{0.509} & \scalebox{0.85}{0.609} & \scalebox{0.85}{0.376} & \scalebox{0.85}{0.628} & \scalebox{0.85}{0.379} \\
    \midrule
    \scalebox{0.85}{Synthetic$^\dagger$} & \secondres{\scalebox{0.85}{0.040}} & \secondres{\scalebox{0.85}{0.155}} & \scalebox{0.85}{--} & \scalebox{0.85}{--} & \scalebox{0.85}{1.099} & \scalebox{0.85}{0.835} & \scalebox{0.85}{0.079} & \scalebox{0.85}{0.211} & \scalebox{0.85}{0.070} & \scalebox{0.85}{0.197} & \scalebox{0.85}{0.254} & \scalebox{0.85}{0.390} & \scalebox{0.85}{0.277} & \scalebox{0.85}{0.408} & \boldres{\scalebox{0.85}{0.025}} & \boldres{\scalebox{0.85}{0.124}} & \scalebox{0.85}{1.196} & \scalebox{0.85}{0.893} & \scalebox{0.85}{0.445} & \scalebox{0.85}{0.523} & \scalebox{0.85}{0.480} & \scalebox{0.85}{0.559} & \scalebox{0.85}{0.097} & \scalebox{0.85}{0.239} & \scalebox{0.85}{2.476} & \scalebox{0.85}{1.265} & \scalebox{0.85}{1.419} & \scalebox{0.85}{0.949} \\
    \midrule
    \scalebox{0.85}{$1^{st}$ Count} & \scalebox{0.85}{0} & \scalebox{0.85}{0} & \scalebox{0.85}{4} & \scalebox{0.85}{4} & \scalebox{0.85}{0} & \scalebox{0.85}{0} & \scalebox{0.85}{0} & \scalebox{0.85}{0} & \scalebox{0.85}{0} & \scalebox{0.85}{0} & \scalebox{0.85}{0} & \scalebox{0.85}{0} & \scalebox{0.85}{0} & \scalebox{0.85}{0} & \scalebox{0.85}{1} & \scalebox{0.85}{1} & \scalebox{0.85}{0} & \scalebox{0.85}{0} & \scalebox{0.85}{0} & \scalebox{0.85}{0} & \scalebox{0.85}{0} & \scalebox{0.85}{0} & \scalebox{0.85}{0} & \scalebox{0.85}{0} & \scalebox{0.85}{0} & \scalebox{0.85}{0} & \scalebox{0.85}{0} & \scalebox{0.85}{0} \\
    \midrule
    \midrule
    \scalebox{0.85}{Avg FLOPs} & \multicolumn{2}{c|}{\scalebox{0.85}{3.33E+09}} & \multicolumn{2}{c|}{\scalebox{0.85}{--}} & \multicolumn{2}{c|}{\scalebox{0.85}{7.59E+06}} & \multicolumn{2}{c|}{\scalebox{0.85}{8.64E+08}} & \multicolumn{2}{c|}{\scalebox{0.85}{8.72E+08}} & \multicolumn{2}{c|}{\scalebox{0.85}{1.34E+09}} & \multicolumn{2}{c|}{\scalebox{0.85}{8.28E+08}} & \multicolumn{2}{c|}{\scalebox{0.85}{2.31E+09}} & \multicolumn{2}{c|}{\scalebox{0.85}{1.31E+09}} & \multicolumn{2}{c|}{\scalebox{0.85}{1.23E+11}} & \multicolumn{2}{c|}{\scalebox{0.85}{1.01E+07}} & \multicolumn{2}{c|}{\scalebox{0.85}{1.41E+09}} & \multicolumn{2}{c|}{\scalebox{0.85}{2.01E+09}} & \multicolumn{2}{c|}{\scalebox{0.85}{5.26E+07}} \\
    \midrule
    \scalebox{0.85}{Avg Params} & \multicolumn{2}{c|}{\scalebox{0.85}{5.08E+06}} & \multicolumn{2}{c|}{\scalebox{0.85}{--}} & \multicolumn{2}{c|}{\scalebox{0.85}{7.71E+05}} & \multicolumn{2}{c|}{\scalebox{0.85}{2.85E+06}} & \multicolumn{2}{c|}{\scalebox{0.85}{1.34E+05}} & \multicolumn{2}{c|}{\scalebox{0.85}{2.30E+06}} & \multicolumn{2}{c|}{\scalebox{0.85}{7.10E+05}} & \multicolumn{2}{c|}{\scalebox{0.85}{7.87E+06}} & \multicolumn{2}{c|}{\scalebox{0.85}{8.53E+06}} & \multicolumn{2}{c|}{\scalebox{0.85}{5.74E+07}} & \multicolumn{2}{c|}{\scalebox{0.85}{6.50E+04}} & \multicolumn{2}{c|}{\scalebox{0.85}{7.14E+07}} & \multicolumn{2}{c|}{\scalebox{0.85}{1.69E+07}} & \multicolumn{2}{c|}{\scalebox{0.85}{2.15E+05}} \\
    \bottomrule
  \end{tabular}
    \end{small}
  \begin{tablenotes}[flushleft]
    \footnotesize
    \item[$\dagger$] Evaluated under multivariate-to-single (MS) setting; hyperparameters chosen identical to ETTm1 for all models. See Appendix~\ref{appendix:synthetic_dataset} for dataset construction details.
    \item[1] Reported MTLinear results reflect the per-dataset best of MTNLinear and MTDLinear.
    \item[p] DeCoP was enabled for XCTFormer on this dataset.
  \end{tablenotes}
  \end{threeparttable}
  }
\end{table}

\paragraph{Results.} As shown in Table~\ref{tab:avg_forecasting_results}, XCTFormer delivers strong forecasting results compared to widely recognized baselines across diverse benchmarks, achieving second-best performance on 8 out of 10 evaluation metrics. These results highlight the model’s effectiveness in capturing cross-channel dependencies in the long-term forecasting task.
On the synthetic dataset, where the target is generated from lagged cross-variate signals, XCTFormer ranks second (Avg MSE\,=\,0.040) while using $55{\times}$ fewer parameters and $364{\times}$ fewer FLOPs than the top-ranked model. These results suggest that XCTFormer models cross-channel and temporal dependencies effectively in this setting, while being less affected by distractor channels, indicating robustness to spurious correlations. Full results and additional details are provided in Appendix~\ref{appendix:synthetic_dataset}.

\begin{table}[t]
  \caption{To evaluate our model performance on imputation, we randomly mask $\{12.5\%, 25\%, 37.5\%, 50\%\}$ of the time points in a time series of length 1024. The final results are averaged across these 4 different masking ratios. \textcolor{red}{Red} indicates best performance (lowest error), \textcolor{blue}{blue} indicates second best.
  }
\renewcommand\arraystretch{0.7}
  \vskip 0.05in
  \centering
  \resizebox{\columnwidth}{!}{
  \begin{threeparttable}
  \begin{small}
  \renewcommand{\multirowsetup}{\centering}
  \label{tab:imputation_avg}
  \setlength{\tabcolsep}{0.7pt}
  \begin{tabular}{c|cc|cc|cc|cc|cc|cc|cc|cc|cc|cc|cc|cc}
    \toprule
    \multicolumn{1}{c}{\multirow{2}{*}{Models}} & 
    \multicolumn{2}{c}{\rotatebox{0}{\scalebox{0.68}{\textbf{XCTFormer}}}} &
    \multicolumn{2}{c}{\rotatebox{0}{\scalebox{0.68}{TimeMixer++}}} &
    \multicolumn{2}{c}{\rotatebox{0}{\scalebox{0.68}{TimeMixer}}} &
    \multicolumn{2}{c}{\rotatebox{0}{\scalebox{0.68}{iTransformer}}} &
    \multicolumn{2}{c}{\rotatebox{0}{\scalebox{0.68}{PatchTST}}} &
    \multicolumn{2}{c}{\rotatebox{0}{\scalebox{0.68}{Crossformer}}} &
    \multicolumn{2}{c}{\rotatebox{0}{\scalebox{0.68}{FEDformer}}} &
    \multicolumn{2}{c}{\rotatebox{0}{\scalebox{0.68}{TIDE}}} &
    \multicolumn{2}{c}{\rotatebox{0}{\scalebox{0.68}{DLinear}}} &
    \multicolumn{2}{c}{\rotatebox{0}{\scalebox{0.68}{TimesNet}}} &
    \multicolumn{2}{c}{\rotatebox{0}{\scalebox{0.68}{MICN}}} &
    \multicolumn{2}{c}{\rotatebox{0}{\scalebox{0.68}{Autoformer}}}
    \\
    \multicolumn{1}{c}{} & \multicolumn{2}{c}{\scalebox{0.68}{(\textbf{Ours})}} & \multicolumn{2}{c}{\scalebox{0.68}{(ICLR 2025)}} & \multicolumn{2}{c}{\scalebox{0.68}{(ICLR 2024)}} & \multicolumn{2}{c}{\scalebox{0.68}{(ICLR 2024)}} & \multicolumn{2}{c}{\scalebox{0.68}{(ICLR 2023)}} & \multicolumn{2}{c}{\scalebox{0.68}{(ICLR 2023)}} & \multicolumn{2}{c}{\scalebox{0.68}{(ICML 2022)}} & \multicolumn{2}{c}{\scalebox{0.68}{(TMLR 2023)}} & \multicolumn{2}{c}{\scalebox{0.68}{(AAAI 2023)}} & \multicolumn{2}{c}{\scalebox{0.68}{(ICLR 2023)}} & \multicolumn{2}{c}{\scalebox{0.68}{(ICLR 2023)}} & \multicolumn{2}{c}{\scalebox{0.68}{(NeurIPS 2021)}} 
    \\
    \cmidrule(lr){2-3} \cmidrule(lr){4-5} \cmidrule(lr){6-7} \cmidrule(lr){8-9} \cmidrule(lr){10-11} \cmidrule(lr){12-13} \cmidrule(lr){14-15} \cmidrule(lr){16-17} \cmidrule(lr){18-19} \cmidrule(lr){20-21} \cmidrule(lr){22-23} \cmidrule(lr){24-25}
    \multicolumn{1}{c}{\scalebox{0.68}{Metric}} & \scalebox{0.68}{MSE} & \scalebox{0.68}{MAE} & \scalebox{0.68}{MSE} & \scalebox{0.68}{MAE} & \scalebox{0.68}{MSE} & \scalebox{0.68}{MAE} & \scalebox{0.68}{MSE} & \scalebox{0.68}{MAE} & \scalebox{0.68}{MSE} & \scalebox{0.68}{MAE} & \scalebox{0.68}{MSE} & \scalebox{0.68}{MAE} & \scalebox{0.68}{MSE} & \scalebox{0.68}{MAE} & \scalebox{0.68}{MSE} & \scalebox{0.68}{MAE} & \scalebox{0.68}{MSE} & \scalebox{0.68}{MAE} & \scalebox{0.68}{MSE} & \scalebox{0.68}{MAE} & \scalebox{0.68}{MSE} & \scalebox{0.68}{MAE} & \scalebox{0.68}{MSE} & \scalebox{0.68}{MAE}\\
    \toprule
    \scalebox{0.68}{ETT(Avg)} &\boldres{\scalebox{0.68}{0.050}} &\boldres{\scalebox{0.68}{0.141}} & \secondres{\scalebox{0.68}{0.055}} &\secondres{\scalebox{0.68}{0.154}} & \scalebox{0.68}{0.097} &\scalebox{0.68}{0.220} & \scalebox{0.68}{0.096} &\scalebox{0.68}{0.205} & \scalebox{0.68}{0.120} &\scalebox{0.68}{0.225} & \scalebox{0.68}{0.150} &\scalebox{0.68}{0.258} & \scalebox{0.68}{0.124} &\scalebox{0.68}{0.230} & \scalebox{0.68}{0.314} &\scalebox{0.68}{0.366} & \scalebox{0.68}{0.115} &\scalebox{0.68}{0.229} & \scalebox{0.68}{0.079} &\scalebox{0.68}{0.182} & \scalebox{0.68}{0.119} &\scalebox{0.68}{0.234} & \scalebox{0.68}{0.104} &\scalebox{0.68}{0.215}\\
    \midrule
    \scalebox{0.68}{Weather} &\boldres{\scalebox{0.68}{0.032}} &\boldres{\scalebox{0.68}{0.049}} & \secondres{\scalebox{0.68}{0.049}} &\secondres{\scalebox{0.68}{0.078}} & \scalebox{0.68}{0.091} &\scalebox{0.68}{0.114} & \scalebox{0.68}{0.095} &\scalebox{0.68}{0.102} & \scalebox{0.68}{0.082} &\scalebox{0.68}{0.149} & \scalebox{0.68}{0.150} &\scalebox{0.68}{0.111} & \scalebox{0.68}{0.064} &\scalebox{0.68}{0.139} & \scalebox{0.68}{0.063} &\scalebox{0.68}{0.131} & \scalebox{0.68}{0.071} &\scalebox{0.68}{0.107} & \scalebox{0.68}{0.061} &\scalebox{0.68}{0.098} & \scalebox{0.68}{0.075} &\scalebox{0.68}{0.126} & \scalebox{0.68}{0.066} &\scalebox{0.68}{0.107}\\
    \midrule
    \scalebox{0.68}{ECL\tnote{p}} &\boldres{\scalebox{0.68}{0.046}} &\boldres{\scalebox{0.68}{0.140}} & \scalebox{0.68}{0.109} &\secondres{\scalebox{0.68}{0.197}} & \scalebox{0.68}{0.142} &\scalebox{0.68}{0.261} & \scalebox{0.68}{0.140} &\scalebox{0.68}{0.223} & \scalebox{0.68}{0.129} &\scalebox{0.68}{0.198} & \scalebox{0.68}{0.125} &\scalebox{0.68}{0.204} & \scalebox{0.68}{0.181} &\scalebox{0.68}{0.314} & \scalebox{0.68}{0.182} &\scalebox{0.68}{0.202} & \secondres{\scalebox{0.68}{0.080}} &\scalebox{0.68}{0.200} & \scalebox{0.68}{0.135} &\scalebox{0.68}{0.255} & \scalebox{0.68}{0.138} &\scalebox{0.68}{0.246} & \scalebox{0.68}{0.141} &\scalebox{0.68}{0.234}\\
    \bottomrule
  \end{tabular}
    \end{small}
      \begin{tablenotes}[flushleft]
      \footnotesize
      \item[p] DeCoP was enabled for XCTFormer on this dataset.
    \end{tablenotes}
  \end{threeparttable}
  }
  \vspace{-10pt}
\end{table}

\subsection{Imputation}

Time-series imputation reconstructs missing values from observed data. We evaluate our model on six widely used multivariate datasets: four ETT subsets (ETTm1, ETTm2, ETTh1, ETTh2) \citep{zhou2021informer}, Electricity (ECL), and Weather. We adopt the TimeMixer++ setup, using a lookback window of 1024 time steps and applying random missing-mask rates of \( \{12.5\%, 25\%, 37.5\%, 50\%\} \). We compare against eleven widely recognized models:
(i) Transformer-based: Autoformer \citep{wu2021autoformer}, FEDformer \citep{zhou2022fedformer}, Crossformer \citep{zhang2023crossformer}, PatchTST \citep{nie2023time}, iTransformer \citep{liu2023itransformer};
(ii) MLP-based: DLinear \citep{zeng2023transformers}, TiDE \citep{das2023long}, TimeMixer \citep{wang2024timemixer};
(iii) Convolutional-based: SCINet \citep{liu2022scinet}, TimesNet \citep{wu2022timesnet}, MICN \citep{wang2023micn}.

\paragraph{Results.} As shown in Table~\ref{tab:imputation_avg}, XCTFormer delivers state-of-the-art (SoTA) imputation results in comparison to competing baselines across diverse benchmarks. With the best performance on all 6 evaluation metrics, particularly, our approach outperforms the second-best baseline by an average of 20.8\% on MSE and 15.3\% on MAE across all datasets, highlighting the model’s effectiveness in capturing cross-channel dependencies.

\subsection{Anomaly Detection}

Anomaly detection seeks to identify unusual or abnormal patterns in time-series, often corresponding to faults, attacks, or rare operational modes. We evaluate on five widely used benchmarks: SMD (Server Machine Dataset, \citep{su2019omnianomaly}), SWaT (Secure Water Treatment, \citep{mathur2016swat}), PSM (Pooled Server Metrics, \citep{abdulaal2021psm}), and NASA telemetry datasets MSL and SMAP \citep{hundman2018nasa}. We compare against nineteen widely used models: (i) RNN/TCN: LSTM \citep{hochreiter1997long}, TCN \citep{Franceschi2018TCN};
(ii) Transformer-based: Transformer \citep{vaswani2017attention}, LogTrans \citep{2019Enhancing}, Reformer \citep{kitaev2020reformer}, Informer \citep{zhou2021informer}, Pyraformer \citep{liu2021pyraformer}, Autoformer \citep{wu2021autoformer}, FEDformer \citep{zhou2022fedformer}, ETSformer \citep{woo2022etsformer}, Stationary (Non-stationary Transformer) \citep{liu2022non}, Anomaly Transformer \citep{xu2021anomaly}, LightTS \citep{lightts}, iTransformer \citep{liu2023itransformer};
(iii) State-space: LSSL \citep{gu2022efficiently};
(iv) Linear/MLP: DLinear \citep{zeng2023transformers}, TiDE \citep{das2023long};
(v) Convolutional/Mixer: TimesNet \citep{wu2022timesnet}, TimeMixer++ \citep{wang2024timemixer++}.

\begin{table}[t]
  \caption{F1 scores (\%) for anomaly detection across five benchmark datasets. \textcolor{red}{Red} indicates best (highest), \textcolor{blue}{blue} indicates second best.}
  \label{tab:anomaly_detection_f1}
  \vskip 0.05in
  \centering
  \resizebox{\columnwidth}{!}{
  \begin{threeparttable}
  \begin{small}
  \renewcommand{\multirowsetup}{\centering}
  \setlength{\tabcolsep}{1pt}
  \begin{tabular}{c|c|c|c|c|c|c|c|c|c|c|c|c|c|c|c|c|c|c|c|c}
    \toprule
    \multicolumn{1}{c}{\multirow{2}{*}{Models}} &
    \multicolumn{1}{c}{\rotatebox{0}{\scalebox{0.8}{\textbf{XCTFormer}}}} &
    \multicolumn{1}{c}{\rotatebox{0}{\scalebox{0.8}{TimeMixer++}}} &
    \multicolumn{1}{c}{\rotatebox{0}{\scalebox{0.8}{iTransformer}}} &
    \multicolumn{1}{c}{\rotatebox{0}{\scalebox{0.8}{TiDE}}} &
    \multicolumn{1}{c}{\rotatebox{0}{\scalebox{0.8}{TimesNet}}} &
    \multicolumn{1}{c}{\rotatebox{0}{\scalebox{0.8}{FEDformer}}} &
    \multicolumn{1}{c}{\rotatebox{0}{\scalebox{0.8}{LightTS}}} &
    \multicolumn{1}{c}{\rotatebox{0}{\scalebox{0.8}{ETSformer}}} &
    \multicolumn{1}{c}{\rotatebox{0}{\scalebox{0.8}{DLinear}}} &
    \multicolumn{1}{c}{\rotatebox{0}{\scalebox{0.8}{Stationary}}} &
    \multicolumn{1}{c}{\rotatebox{0}{\scalebox{0.8}{LSSL}}} &
    \multicolumn{1}{c}{\rotatebox{0}{\scalebox{0.8}{Autoformer}}} &
    \multicolumn{1}{c}{\rotatebox{0}{\scalebox{0.8}{Pyraformer}}} &
    \multicolumn{1}{c}{\rotatebox{0}{\scalebox{0.8}{Anomaly\tnote{$\ast$}}}} &
    \multicolumn{1}{c}{\rotatebox{0}{\scalebox{0.8}{Informer}}} &
    \multicolumn{1}{c}{\rotatebox{0}{\scalebox{0.8}{Reformer}}} &
    \multicolumn{1}{c}{\rotatebox{0}{\scalebox{0.8}{TCN}}} &
    \multicolumn{1}{c}{\rotatebox{0}{\scalebox{0.8}{LogTrans}}} &
    \multicolumn{1}{c}{\rotatebox{0}{\scalebox{0.8}{Transformer}}} &
    \multicolumn{1}{c}{\rotatebox{0}{\scalebox{0.8}{LSTM}}}
    \\
    \multicolumn{1}{c}{} & \multicolumn{1}{c}{\scalebox{0.8}{(\textbf{Ours})}} & \multicolumn{1}{c}{\scalebox{0.8}{(ICLR 2025)}} & \multicolumn{1}{c}{\scalebox{0.8}{(ICLR 2024)}} & \multicolumn{1}{c}{\scalebox{0.8}{(TMLR 2023)}} & \multicolumn{1}{c}{\scalebox{0.8}{(ICLR 2023)}} & \multicolumn{1}{c}{\scalebox{0.8}{(ICML 2022)}} & \multicolumn{1}{c}{\scalebox{0.8}{(AAAI 2022)}} & \multicolumn{1}{c}{\scalebox{0.8}{(ICML 2022)}} & \multicolumn{1}{c}{\scalebox{0.8}{(AAAI 2023)}} & \multicolumn{1}{c}{\scalebox{0.8}{(NeurIPS 2022)}} & \multicolumn{1}{c}{\scalebox{0.8}{(NeurIPS 2022)}} & \multicolumn{1}{c}{\scalebox{0.8}{(NeurIPS 2021)}} & \multicolumn{1}{c}{\scalebox{0.8}{(ICLR 2022)}} & \multicolumn{1}{c}{\scalebox{0.8}{(ICLR 2022)}} & \multicolumn{1}{c}{\scalebox{0.8}{(AAAI 2021)}} & \multicolumn{1}{c}{\scalebox{0.8}{(ICLR 2020)}} & \multicolumn{1}{c}{\scalebox{0.8}{(2018)}} & \multicolumn{1}{c}{\scalebox{0.8}{(NeurIPS 2019)}} & \multicolumn{1}{c}{\scalebox{0.8}{(NeurIPS 2017)}} & \multicolumn{1}{c}{\scalebox{0.8}{(1997)}}
    \\
    \cmidrule(lr){2-2} \cmidrule(lr){3-3} \cmidrule(lr){4-4} \cmidrule(lr){5-5} \cmidrule(lr){6-6} \cmidrule(lr){7-7} \cmidrule(lr){8-8} \cmidrule(lr){9-9} \cmidrule(lr){10-10} \cmidrule(lr){11-11} \cmidrule(lr){12-12} \cmidrule(lr){13-13} \cmidrule(lr){14-14} \cmidrule(lr){15-15} \cmidrule(lr){16-16} \cmidrule(lr){17-17} \cmidrule(lr){18-18} \cmidrule(lr){19-19} \cmidrule(lr){20-20} \cmidrule(lr){21-21}
    \multicolumn{1}{c}{\scalebox{0.68}{Metric}} & \scalebox{0.68}{F1} & \scalebox{0.68}{F1} & \scalebox{0.68}{F1} & \scalebox{0.68}{F1} & \scalebox{0.68}{F1} & \scalebox{0.68}{F1} & \scalebox{0.68}{F1} & \scalebox{0.68}{F1} & \scalebox{0.68}{F1} & \scalebox{0.68}{F1} & \scalebox{0.68}{F1} & \scalebox{0.68}{F1} & \scalebox{0.68}{F1} & \scalebox{0.68}{F1} & \scalebox{0.68}{F1} & \scalebox{0.68}{F1} & \scalebox{0.68}{F1} & \scalebox{0.68}{F1} & \scalebox{0.68}{F1} & \scalebox{0.68}{F1}\\
    \toprule
    \scalebox{0.95}{SMD}  & \scalebox{1.2}{84.21} & \boldres{\scalebox{1.2}{86.50}} & \scalebox{1.2}{71.15} & \scalebox{1.2}{68.91} & \secondres{\scalebox{1.2}{85.81}} & \scalebox{1.2}{85.08} & \scalebox{1.2}{82.53} & \scalebox{1.2}{83.13} & \scalebox{1.2}{77.10} & \scalebox{1.2}{84.62} & \scalebox{1.2}{71.31} & \scalebox{1.2}{85.11} & \scalebox{1.2}{83.04} & \scalebox{1.2}{85.49} & \scalebox{1.2}{81.65} & \scalebox{1.2}{75.32} & \scalebox{1.2}{81.49} & \scalebox{1.2}{76.21} & \scalebox{1.2}{79.56} & \scalebox{1.2}{71.41} \\
    \midrule
    \scalebox{0.95}{MSL} & \scalebox{1.2}{79.05} & \boldres{\scalebox{1.2}{85.82}} & \scalebox{1.2}{72.54} & \scalebox{1.2}{70.18} & \secondres{\scalebox{1.2}{85.15}} & \scalebox{1.2}{78.57} & \scalebox{1.2}{78.95} & \scalebox{1.2}{85.03} & \scalebox{1.2}{84.88} & \scalebox{1.2}{77.50} & \scalebox{1.2}{82.53} & \scalebox{1.2}{79.05} & \scalebox{1.2}{84.86} & \scalebox{1.2}{83.31} & \scalebox{1.2}{84.06} & \scalebox{1.2}{84.40} & \scalebox{1.2}{78.60} & \scalebox{1.2}{79.57} & \scalebox{1.2}{78.68} & \scalebox{1.2}{81.93} \\
    \midrule
    \scalebox{0.95}{SMAP} & \boldres{\scalebox{1.2}{86.68}} & \secondres{\scalebox{1.2}{73.10}} & \scalebox{1.2}{66.87} & \scalebox{1.2}{64.00} & \scalebox{1.2}{71.52} & \scalebox{1.2}{70.76} & \scalebox{1.2}{69.21} & \scalebox{1.2}{69.50} & \scalebox{1.2}{69.26} & \scalebox{1.2}{71.09} & \scalebox{1.2}{66.90} & \scalebox{1.2}{71.12} & \scalebox{1.2}{71.09} & \scalebox{1.2}{71.18} & \scalebox{1.2}{69.92} & \scalebox{1.2}{70.40} & \scalebox{1.2}{70.45} & \scalebox{1.2}{69.97} & \scalebox{1.2}{69.70} & \scalebox{1.2}{70.48} \\
    \midrule
    \scalebox{0.95}{SWaT} & \scalebox{1.2}{92.60} & \boldres{\scalebox{1.2}{94.64}} & \scalebox{1.2}{79.18} & \scalebox{1.2}{76.73} & \scalebox{1.2}{91.74} & \scalebox{1.2}{93.19} & \secondres{\scalebox{1.2}{93.33}} & \scalebox{1.2}{84.91} & \scalebox{1.2}{87.52} & \scalebox{1.2}{79.88} & \scalebox{1.2}{85.76} & \scalebox{1.2}{92.74} & \scalebox{1.2}{91.78} & \scalebox{1.2}{83.10} & \scalebox{1.2}{81.43} & \scalebox{1.2}{82.80} & \scalebox{1.2}{85.09} & \scalebox{1.2}{80.52} & \scalebox{1.2}{80.37} & \scalebox{1.2}{84.34} \\
    \midrule
    \scalebox{0.95}{PSM} & \scalebox{1.2}{95.30} & \boldres{\scalebox{1.2}{97.60}} & \scalebox{1.2}{95.17} & \scalebox{1.2}{92.50} & \secondres{\scalebox{1.2}{97.47}} & \scalebox{1.2}{97.23} & \scalebox{1.2}{97.15} & \scalebox{1.2}{91.76} & \scalebox{1.2}{93.55} & \scalebox{1.2}{97.29} & \scalebox{1.2}{77.20} & \scalebox{1.2}{93.29} & \scalebox{1.2}{82.08} & \scalebox{1.2}{79.40} & \scalebox{1.2}{77.10} & \scalebox{1.2}{73.61} & \scalebox{1.2}{70.57} & \scalebox{1.2}{76.74} & \scalebox{1.2}{76.07} & \scalebox{1.2}{81.67} \\
    \midrule
    \scalebox{0.95}{\textbf{Avg}} & \boldres{\scalebox{1.2}{87.57}} & \secondres{\scalebox{1.2}{87.47}} & \scalebox{1.2}{76.98} & \scalebox{1.2}{74.46} & \scalebox{1.2}{86.34} & \scalebox{1.2}{84.97} & \scalebox{1.2}{84.23} & \scalebox{1.2}{82.87} & \scalebox{1.2}{82.46} & \scalebox{1.2}{82.08} & \scalebox{1.2}{76.74} & \scalebox{1.2}{84.26} & \scalebox{1.2}{82.57} & \scalebox{1.2}{80.50} & \scalebox{1.2}{78.83} & \scalebox{1.2}{77.31} & \scalebox{1.2}{77.24} & \scalebox{1.2}{76.60} & \scalebox{1.2}{76.88} & \scalebox{1.2}{77.97} \\
    \bottomrule
  \end{tabular}
    \end{small}
      \begin{tablenotes}[flushleft]
      \footnotesize
      \item[$\ast$] Anomaly Transformer \citep{xu2021anomaly}; for fair comparison, only reconstruction error is used as the anomaly criterion.
    \end{tablenotes}
  \end{threeparttable}
  }
  \vspace{-10pt}
\end{table}

\paragraph{Results.} As shown in Table~\ref{tab:anomaly_detection_f1}, XCTFormer performs competitively against the considered strong baselines (for detailed comparison table refer to Appendix~\ref{appendix:anomaly_detection_full_results}).  Our model achieves a high $F_1$ score, suggesting it captures cross-channel dependencies effectively for anomaly detection.

\newpage

\section{Analysis}

\paragraph{Ablation Study.}
To evaluate each component's contribution, we conducted an ablation study across three fundamental time-series tasks: long-term forecasting, imputation, and anomaly detection. 
Our methodology involved two categories of experiments: (i) \textit{component-wise analysis}, where we systematically removed or altered individual architectural modifications introduced to the vanilla Transformer to isolate each component's impact. Modifications include: removing the learnable mask, reverting the activation function from our proposed approach back to the standard softmax and both;  (ii) \textit{dependency modeling analysis}, where we examined variants that model only cross-channel dependencies (inspired by iTransformer~\citep{liu2023itransformer}) or only temporal relationships (similar to PatchTST~\citep{nie2023time}) to validate the necessity of our integrated cross-channel and cross-time modeling strategy.
Results presented in Table~\ref{tab:ablation_study_xctformer} report the average performance metrics across all datasets and experimental configurations specific to each task for every model variation. The full XCTFormer consistently outperforms all variants across all three tasks and evaluation metrics. These findings validate our architectural design choices and provide empirical evidence that each proposed component contributes meaningfully to the model's overall performance across diverse time-series applications. For detailed experimental configurations and full results, refer to Appendix~\ref{appendix:ablation_details}.

\begin{table}[htp]
    \caption{Ablation study results across different tasks, evaluated with different XCTFormer variations.}
    \label{tab:ablation_study_xctformer}
    \centering
    \resizebox{\columnwidth}{!}{
    \begin{tabular}{lcccccccc}
        \toprule
            & \multicolumn{2}{c}{Long-term Forecasting} & \multicolumn{2}{c}{Imputation} & \multicolumn{3}{c}{Anomaly Detection} & \begin{tabular}[c]{@{}c@{}}XCTFormer\\vs Others\end{tabular} \\
            \cmidrule(lr){2-3} \cmidrule(lr){4-5} \cmidrule(lr){6-8} \cmidrule(lr){9-9}
            & MSE & MAE & MSE & MAE & Precision & Recall & F-Score & (\%) \\
        \midrule
           \rowcolor{tabhighlight}
            XCTFormer (Original) & \textbf{0.328} & \textbf{0.337} & \textbf{0.044} & \textbf{0.124} & \textbf{92.1} & \textbf{83.7} & \textbf{87.6} & - \\
            \hspace{2em} W/o mask$^\dagger$ &  0.351 &  0.369 &  0.051 &  0.131 & 90.6 & 74.5 & 81.1 &  8.0\% \\
            \hspace{2em} Original softmax activation & 0.359 & 0.364 & 0.053 & 0.143 & 90.3 & 75.4 & 81.5 & 9.7\% \\
            \hspace{2em} Vanilla transformer & 0.361 & 0.364 & 0.060 & 0.149 & 90.9 & 76.1 & 82.0 & 11.2\% \\
            \hspace{2em} Sequence modeling & 0.341 & \underline{0.343} & 0.052 & \underline{0.131} & \underline{91.2} & \underline{78.8} & \underline{83.9} & 5.6\% \\
            \hspace{2em} Channel modeling & 0.341 & 0.348 & 0.081 & 0.174 & 91.2 & 76.1 & 82.3 & 14.2\% \\
        \bottomrule
    \end{tabular}
   }
   \vspace{0.3em}
   \footnotesize{$^\dagger$ W/o mask averages exclude DeCoP datasets (ECL, Traffic) where the learnable mask is not applicable.}
\end{table}
\paragraph{Robustness Across Random Seeds.}
To evaluate the stability and reliability of XCTFormer, we assessed its performance across different random initializations. Neural models are often sensitive to parameter initialization randomness and the order of training samples, leading to variability in results. To address this, we trained XCTFormer using the optimal hyperparameters selected by validation on five distinct random seeds (2021 to 2025). For each of the three primary time-series tasks: long-term forecasting, anomaly detection, and imputation, we report both the mean and standard deviation of the relevant performance metric, providing a more robust estimate of model effectiveness. We further quantify robustness using a confidence score, calculated from the coefficient of variation  \citep{Reed2002CV}, which reflects the model’s precision and repeatability. In this context, a lower standard deviation indicates greater consistency and, therefore, higher reliability. Summarized seed robustness results are presented in Table \ref{tab:summarized_seed_results}. For more information on confidence score calculation and the complete analysis tables, refer to Appendix \ref{appendix:full_seed_analysis}.
\begin{table}[H]
  \caption{Averaged metrics of trained models, evaluated on five different seeds (2021–2025) across all datasets, are reported for each metric and time-series task, along with the corresponding confidence interval.}
  \label{tab:summarized_seed_results}
  \vskip 0.05in
  \centering
  \begin{threeparttable}
  \begin{small}
  \setlength{\tabcolsep}{6pt}
  \begin{tabular*}{\textwidth}{@{\extracolsep{\fill}} l l c c}
    \toprule
    Task & Metric & Mean $\pm$ Avg.\ Std & Confidence Score (\%) \\
    \midrule
    Long-Term Forecasting & MSE & $0.330 \pm 0.003$ & 99.05\% \\
    Long-Term Forecasting & MAE & $0.339 \pm 0.002$ & 99.28\%  \\
    \midrule
    Imputation & MSE & $0.047 \pm 0.005$ & 89.14\%  \\
    Imputation & MAE & $0.128 \pm 0.009$  & 93.10\%  \\
    \midrule
    Anomaly Detection & Precision & $91.382 \pm 0.868$   & 99.05\%  \\
    Anomaly Detection & Recall    & $79.660 \pm 4.298$   & 94.60\%  \\
    Anomaly Detection & F1        & $84.600 \pm 3.100$   & 96.34\%  \\
    \bottomrule
  \end{tabular*}
  \end{small}
  \end{threeparttable}
\end{table}

\section{Limitations}

XCTFormer explicitly models all pairwise channel-time dependencies via a unified attention block, improving expressiveness and delivering strong performance relative to well-established baselines.  However, this design also introduces practical limitations. Flattening time channel tokens makes attention quadratic in the number of tokens, increasing memory and runtime as the lookback length and channel count grow. DeCoP mitigates this cost by compressing attention to a linear form, but it still scales with sequence length and dimensionality and adds a decent parameter overhead. Finally, the gains are not uniform across datasets and tasks, with some settings showing smaller improvements or higher variance, suggesting that the presented pairwise modeling strategy is sensitive to the underlying dependency structure and may offer limited benefits when cross-channel relations are weak or difficult to capture.

\section{Conclusion}
\label{sec:conclusion}

In this paper, we address a fundamental paradox in multivariate time-series analysis: although leveraging cross-channel structure should improve performance, recent findings show that channel-independent models often outperform channel-dependent models. This counterintuitive result suggests that existing channel-dependent methods do not fully exploit cross-channel information. We argue that this limitation arises from current approaches that model cross-channel and cross-time dependencies \emph{indirectly}, thereby overlooking interactions. To bridge this gap, we introduce \textbf{XCTFormer}, which revisits channel dependence through direct, token-by-token modeling. Instead of treating channels and time steps as separate entities processed through multi-stage pipelines, XCTFormer treats each channel-time data point as an individual token and models all pairwise dependencies within a unified attention mechanism, which is important for capturing time-evolving dependencies. Through the Cross-Relational Attention Block (CRAB) with  learnable masking (when DeCoP is not applied) and an enhanced attention activation function, XCTFormer improves expressivity while maintaining robustness, and the optional Dependency Compression Plugin (DeCoP) supports scalability on high-dimensional datasets. Extensive evaluation across forecasting, anomaly detection, and imputation highlights XCTFormer’s effectiveness: it delivers state-of-the-art imputation accuracy, with average error reductions of 20.8\% in MSE and 15.3\% in MAE, while also achieving strong performance gains in forecasting and anomaly detection. Additional evaluation on a synthetic dataset, where the target depends on lagged cross-variate signals with distractor channels, suggests that XCTFormer effectively captures cross-channel and cross-time dependencies while showing robustness to spurious correlations. At the same time, this direct modeling strategy introduces practical limitations: Unified token-to-token attention scales quadratically with the number of time-channel tokens, and while DeCoP reduces this cost, the parameter count still grows linearly and remains non-negligible. In addition, gains are not uniform across datasets and tasks, with some settings showing smaller improvements or higher variance, suggesting sensitivity to the underlying dependency structure. This motivates further research into more robust channel-time modeling strategies that balance expressiveness, efficiency, and consistency across diverse datasets and tasks. Despite the limitations presented, our proposed direct modeling approach represents a substantial step toward a more comprehensive capture of dependencies and toward realizing the full modeling potential of multivariate time-series data.

\section*{Acknowledgments}

This research was partially supported by the Lynn and William Frankel Center of The Stein Faculty of Computer and Information Science, Ben-Gurion University of the Negev, ISF grants 668/21 and 1299/25, an ISF equipment grant, and by the Israeli Council for Higher Education (CHE) via the Data Science Research Center, Ben-Gurion University of the Negev, Israel.

\newpage
\bibliography{tmlr}

@inproceedings{zeng2023transformers,
  author       = {Ailing Zeng and
                  Muxi Chen and
                  Lei Zhang and
                  Qiang Xu},
  title        = {Are Transformers Effective for Time Series Forecasting?},
  booktitle    = {Thirty-Seventh Conference on Artificial Intelligence},
  publisher    = {{AAAI}},
  year         = {2023},
}

@inproceedings{liu2023itransformer,
  author       = {Yong Liu and
                  Tengge Hu and
                  Haoran Zhang and
                  Haixu Wu and
                  Shiyu Wang and
                  Lintao Ma and
                  Mingsheng Long},
  title        = {{iTransformer}: Inverted Transformers Are Effective for Time Series
                  Forecasting},
  booktitle    = {The Twelfth International Conference on Learning Representations,
                  {ICLR}},
  year         = {2024},
}

@inproceedings{wu2021autoformer,
  author       = {Haixu Wu and
                  Jiehui Xu and
                  Jianmin Wang and
                  Mingsheng Long},
  title        = {Autoformer: Decomposition Transformers with Auto-Correlation for Long-Term
                  Series Forecasting},
  booktitle    = {Annual Conference
                  on Neural Information Processing Systems 2021, NeurIPS},
  year         = {2021},
}

@inproceedings{nie2023time,
  author       = {Yuqi Nie and
                  Nam H. Nguyen and
                  Phanwadee Sinthong and
                  Jayant Kalagnanam},
  title        = {A {Time Series is Worth 64 Words: Long-term Forecasting with Transformers}},
  booktitle    = {The Eleventh International Conference on Learning Representations,
                  {ICLR}},
  year         = {2023},
}

@inproceedings{vaswani2017attention,
  author       = {Ashish Vaswani and
                  Noam Shazeer and
                  Niki Parmar and
                  Jakob Uszkoreit and
                  Llion Jones and
                  Aidan N. Gomez and
                  Lukasz Kaiser and
                  Illia Polosukhin},
  title        = {Attention is All you Need},
  booktitle    = {Annual Conference
                  on Neural Information Processing Systems, NeurIPS},
  year         = {2017},
}

@book{box_jenkins_1970,
  author    = {Box, George E. P. and Jenkins, Gwilym M.},
  title     = {Time Series Analysis: Forecasting and Control},
  publisher = {Holden-Day},
  year      = {1970},
  edition   = {First},
}

@article{hochreiter1997long,
  title={Long short-term memory},
  author={Hochreiter, Sepp and Schmidhuber, J{\"u}rgen},
  journal={Neural computation},
  year={1997},
  publisher={MIT Press}
}

@inproceedings{Franceschi2018TCN,
  author       = {Jean{-}Yves Franceschi and
                  Aymeric Dieuleveut and
                  Martin Jaggi},
  title        = {{Unsupervised Scalable Representation Learning for Multivariate Time
                  Series}},
  booktitle    = {Annual Conference
                  on Neural Information Processing Systems, NeurIPS},
  year         = {2019},
}

@inproceedings{leddam,
  author       = {Guoqi Yu and
                  Jing Zou and
                  Xiaowei Hu and
                  Angelica I. Avil{\'{e}}s{-}Rivero and
                  Jing Qin and
                  Shujun Wang},
  title        = {{Revitalizing Multivariate Time Series Forecasting: Learnable Decomposition
                  with Inter-Series Dependencies and Intra-Series Variations Modeling}},
  booktitle    = {Forty-first International Conference on Machine Learning, {ICML}},
  year         = {2024},
}

@article{han2024capacity,
  author       = {Lu Han and
                  Han{-}Jia Ye and
                  De{-}Chuan Zhan},
  title        = {{The Capacity and Robustness Trade-Off: Revisiting the Channel Independent
                  Strategy for Multivariate Time Series Forecasting}},
  journal      = {{IEEE} Trans. Knowl. Data Eng.},
  volume       = {36},
  number       = {11},
  year         = {2024},
}

@inproceedings{zhou2021informer,
  author       = {Haoyi Zhou and
                  Shanghang Zhang and
                  Jieqi Peng and
                  Shuai Zhang and
                  Jianxin Li and
                  Hui Xiong and
                  Wancai Zhang},
  title        = {{Informer: Beyond Efficient Transformer for Long Sequence Time-Series
                  Forecasting}},
  booktitle    = {Thirty-Fifth Conference on Artificial Intelligence},
  publisher    = {AAAI},
  year         = {2021},
}

@inproceedings{li2019enhancing,
  author       = {Shiyang Li and
                  Xiaoyong Jin and
                  Yao Xuan and
                  Xiyou Zhou and
                  Wenhu Chen and
                  Yu{-}Xiang Wang and
                  Xifeng Yan},
  title        = {{Enhancing the Locality and Breaking the Memory Bottleneck of Transformer
                  on Time Series Forecasting}},
  booktitle    = {Annual Conference
                  on Neural Information Processing Systems, NeurIPS },
  year         = {2019},
}

@inproceedings{zhou2022fedformer,
  author       = {Tian Zhou and
                  Ziqing Ma and
                  Qingsong Wen and
                  Xue Wang and
                  Liang Sun and
                  Rong Jin},
  title        = {{FEDformer: Frequency Enhanced Decomposed Transformer for Long-term
                  Series Forecasting}},
  booktitle    = {International Conference on Machine Learning, {ICML}},
  year         = {2022},
}

@inproceedings{
wang2024card,
title={{CARD: Channel Aligned Robust Blend Transformer for Time Series Forecasting}},
author={Xue Wang and Tian Zhou and Qingsong Wen and Jinyang Gao and Bolin Ding and Rong Jin},
booktitle={The Twelfth International Conference on Learning Representations, ICLR},
year={2024},
}

@article{wang2024ai,
  title   = {{AI-Empowered Methods for Smart Energy Consumption: A Review of Load Forecasting, Anomaly Detection and Demand Response}},
  author  = {Wang, Xinlin and Wang, Hao and Bhandari, Binayak and Cheng, Leming},
  journal = {International Journal of Precision Engineering and Manufacturing-Green Technology},
  year    = {2024},
}

@inproceedings{bui2018time,
  title={{Time series forecasting for healthcare diagnosis and prognostics with the focus on cardiovascular diseases}},
  author={Bui, C and Pham, N and Vo, A and Tran, A and Nguyen, A and Le, T},
  booktitle={6th International Conference on the Development of Biomedical Engineering in Vietnam (BME6)},
  year={2018},
  organization={Springer}
}

@article{mystakidis2024energy,
  title={{Energy Forecasting: A Comprehensive Review of Techniques and Technologies}},
  author={Mystakidis, Aristeidis and Koukaras, Paraskevas and Tsalikidis, Nikolaos and Ioannidis, Dimosthenis and Tjortjis, Christos},
  journal={Energies},
  year={2024},
  publisher={MDPI}
}

@article{duarte2021comparison,
  title={{A comparison of time-series predictions for healthcare emergency department indicators and the impact of COVID-19}},
  author={Duarte, Diego and Walshaw, Chris and Ramesh, Nadarajah},
  journal={Applied Sciences},
  year={2021},
  publisher={MDPI}
}

@article{brunet2023advancing,
  title={{Advancing weather and climate forecasting for our changing world}},
  author={Brunet, Gilbert and Parsons, David B and Ivanov, Dimitar and Lee, Boram and Bauer, Peter and Bernier, Natacha B and Bouchet, Veronique and Brown, Andy and Busalacchi, Antonio and Flatter, Georgina Campbell and others},
  journal={Bulletin of the American Meteorological Society},
  year={2023},
  publisher={American Meteorological Society}
}

@inproceedings{liu2021pyraformer,
  author       = {Shizhan Liu and
                  Hang Yu and
                  Cong Liao and
                  Jianguo Li and
                  Weiyao Lin and
                  Alex X. Liu and
                  Schahram Dustdar},
  title        = {{Pyraformer: Low-Complexity Pyramidal Attention for Long-Range Time
                  Series Modeling and Forecasting}},
  booktitle    = {The Tenth International Conference on Learning Representations, {ICLR}},
  year         = {2022},
}

@inproceedings{zhang2023crossformer,
  author       = {Yunhao Zhang and
                  Junchi Yan},
  title        = {{Crossformer: Transformer Utilizing Cross-Dimension Dependency for
                  Multivariate Time Series Forecasting}},
  booktitle    = {The Eleventh International Conference on Learning Representations,
                  {ICLR}},
  year         = {2023},
}

@article{das2023long,
  author       = {Abhimanyu Das and
                  Weihao Kong and
                  Andrew Leach and
                  Shaan Mathur and
                  Rajat Sen and
                  Rose Yu},
  title        = {{Long-term Forecasting with TiDE: Time-series Dense Encoder}},
  journal      = {Transactions on Machine Learning Research},
  year         = {2023},
}

@inproceedings{he2015delving,
  author       = {Kaiming He and
                  Xiangyu Zhang and
                  Shaoqing Ren and
                  Jian Sun},
  title        = {{Delving Deep into Rectifiers: Surpassing Human-Level Performance on
                  ImageNet Classification}},
  booktitle    = {{IEEE} International Conference on Computer Vision, {ICCV}},
  year         = {2015},
}

@inproceedings{wang2024timemixer,
  author       = {Shiyu Wang and
                  Haixu Wu and
                  Xiaoming Shi and
                  Tengge Hu and
                  Huakun Luo and
                  Lintao Ma and
                  James Y. Zhang and
                  Jun Zhou},
  title        = {{TimeMixer: Decomposable Multiscale Mixing for Time Series Forecasting}},
  booktitle    = {The Twelfth International Conference on Learning Representations,
                  {ICLR} },
  year         = {2024},
}

@inproceedings{wang2024timemixer++,
  author       = {Shiyu Wang and
                  Jiawei Li and
                  Xiaoming Shi and
                  Zhou Ye and
                  Baichuan Mo and
                  Wenze Lin and
                  Shengtong Ju and
                  Zhixuan Chu and
                  Ming Jin},
  title        = {{TimeMixer++: A General Time Series Pattern Machine for Universal
                  Predictive Analysis}},
  booktitle    = {The Thirteenth International Conference on Learning Representations,
                  {ICLR}},
  year         = {2025},
}

@inproceedings{kim2022reversible,
  author       = {Taesung Kim and
                  Jinhee Kim and
                  Yunwon Tae and
                  Cheonbok Park and
                  Jang{-}Ho Choi and
                  Jaegul Choo},
  title        = {{Reversible Instance Normalization for Accurate Time-Series Forecasting
                  against Distribution Shift}},
  booktitle    = {The Tenth International Conference on Learning Representations, {ICLR}},
  year         = {2022},
}

@article{trirat2024universal,
  author       = {Patara Trirat and
                  Yooju Shin and
                  Junhyeok Kang and
                  Youngeun Nam and
                  Jihye Na and
                  Minyoung Bae and
                  Joeun Kim and
                  Byunghyun Kim and
                  Jae{-}Gil Lee},
  title        = {{Universal Time-Series Representation Learning: A Survey}},
  journal      = {arXiv},
  year         = {2024},
  eprint       = {2401.03717},
}

@inproceedings{isik2025scaling,
  author       = {Berivan Isik and
                  Natalia Ponomareva and
                  Hussein Hazimeh and
                  Dimitris Paparas and
                  Sergei Vassilvitskii and
                  Sanmi Koyejo},
  title        = {{Scaling Laws for Downstream Task Performance in Machine Translation}},
  booktitle    = {The Thirteenth International Conference on Learning Representations,
                  {ICLR}},
  year         = {2025},
}

@article{Domingos2012FewUsefulThings,
  author  = {Pedro Domingos},
  title   = {{A Few Useful Things to Know about Machine Learning}},
  journal = {Communications of the ACM},
  year    = {2012},
}

@article{Jin2024GNN4TS,
title = {{A Survey on Graph Neural Networks for Time Series: Forecasting, Classification, Imputation, and Anomaly Detection}},
author = {Ming Jin and Huan Yee Koh and Qingsong Wen and Daniele Zambon and Cesare Alippi and Geoffrey I. Webb and Irwin King and Shirui Pan},
journal = {IEEE TPAMI},
year = {2024}}

@article{lv2024more,
  title={{More Expressive Attention with Negative Weights}},
  author={Lv, Ang and Xie, Ruobing and Li, Shuaipeng and Liao, Jiayi and Sun, Xingwu and Kang, Zhanhui and Wang, Di and Yan, Rui},
  journal={arXiv preprint arXiv:2411.07176},
  year={2024}
}

@inproceedings{wang2023micn,
  author       = {Huiqiang Wang and
                  Jian Peng and
                  Feihu Huang and
                  Jince Wang and
                  Junhui Chen and
                  Yifei Xiao},
  title        = {{MICN: Multi-scale Local and Global Context Modeling for Long-term
                  Series Forecasting}},
  booktitle    = {The Eleventh International Conference on Learning Representations,
                  {ICLR}},
  year         = {2023},
}

@inproceedings{liu2022scinet,
  author       = {Minhao Liu and
                  Ailing Zeng and
                  Muxi Chen and
                  Zhijian Xu and
                  Qiuxia Lai and
                  Lingna Ma and
                  Qiang Xu},
  title        = {{SCINet: Time Series Modeling and Forecasting with Sample Convolution
                  and Interaction}},
  booktitle    = {Annual Conference
                  on Neural Information Processing Systems, NeurIPS },
  year         = {2022},
}

@inproceedings{wu2022timesnet,
  author       = {Haixu Wu and
                  Tengge Hu and
                  Yong Liu and
                  Hang Zhou and
                  Jianmin Wang and
                  Mingsheng Long},
  title        = {{TimesNet: Temporal 2D-Variation Modeling for General Time Series Analysis}},
  booktitle    = {The Eleventh International Conference on Learning Representations,
                  {ICLR}},
  year         = {2023},
}

@misc{saratchandran2025rethinking,
      title={{Rethinking Attention: Polynomial Alternatives to Softmax in Transformers}}, 
      author={Hemanth Saratchandran and Jianqiao Zheng and Yiping Ji and Wenbo Zhang and Simon Lucey},
      year={2025},
      eprint={2410.18613},
      archivePrefix={arXiv},
}

@inproceedings{zhao2024rethinking,
  author       = {Lifan Zhao and
                  Yanyan Shen},
  title        = {{Rethinking Channel Dependence for Multivariate Time Series Forecasting:
                  Learning from Leading Indicators}},
  booktitle    = {The Twelfth International Conference on Learning Representations,
                  {ICLR}},
  year         = {2024},
}

@inproceedings{chen2024structured,
  author       = {Xiaodan Chen and
                  Xiucheng Li and
                  Xinyang Chen and
                  Zhijun Li},
  title        = {{Structured Matrix Basis for Multivariate Time Series Forecasting with
                  Interpretable Dynamics}},
  booktitle    = {Annual Conference
                  on Neural Information Processing Systems, NeurIPS 
                  },
  year         = {2024},
}

@inproceedings{nochumsohn2025multi,
  author       = {Liran Nochumsohn and
                  Hedi Zisling and
                  Omri Azencot},
  title        = {{A Multi-Task Learning Approach to Linear Multivariate Forecasting}},
  booktitle    = {International Conference on Artificial Intelligence and Statistics,
                  {AISTATS} },
  year         = {2025},
}

@inproceedings{paszke2019pytorch,
  author       = {Adam Paszke and
                  Sam Gross and
                  Francisco Massa and
                  Adam Lerer and
                  James Bradbury and
                  Gregory Chanan and
                  Trevor Killeen and
                  Zeming Lin and
                  Natalia Gimelshein and
                  Luca Antiga and
                  Alban Desmaison and
                  Andreas K{\"{o}}pf and
                  Edward Z. Yang and
                  Zachary DeVito and
                  Martin Raison and
                  Alykhan Tejani and
                  Sasank Chilamkurthy and
                  Benoit Steiner and
                  Lu Fang and
                  Junjie Bai and
                  Soumith Chintala},
  title        = {{PyTorch: An Imperative Style, High-Performance Deep Learning Library}},
  booktitle    = {Annual Conference
                  on Neural Information Processing Systems, NeurIPS },
  year         = {2019},
}

@inproceedings{kingma2015adam,
  author       = {Diederik P. Kingma and
                  Jimmy Ba},
  title        = {{Adam: A Method for Stochastic Optimization}},
  booktitle    = {3rd International Conference on Learning Representations, {ICLR}},
  year         = {2015},
}

@inproceedings{2019Enhancing,
  title     = {{Enhancing the Locality and Breaking the Memory Bottleneck of Transformer on Time Series Forecasting}},
  author    = {Li, Shiyang and Jin, Xiaoyong and Xuan, Yao and Zhou, Xiyou and Chen, Wenhu and Wang, Yu-Xiang and Yan, Xifeng},
  booktitle = {Advances in Neural Information Processing Systems (NeurIPS)},
  year      = {2019}
}

@inproceedings{kitaev2020reformer,
  title     = {{Reformer: The Efficient Transformer}},
  author    = {Kitaev, Nikita and Kaiser, {\L}ukasz and Levskaya, Anselm},
  booktitle = {International Conference on Learning Representations (ICLR)},
  year      = {2020}
}

@inproceedings{xu2021anomaly,
  title     = {{Anomaly Transformer: Time Series Anomaly Detection with Association Discrepancy}},
  author    = {Xu, Jiehui and Wu, Haixu and Wang, Jianmin and Long, Mingsheng},
  booktitle = {International Conference on Learning Representations (ICLR)},
  year      = {2022}
}

@inproceedings{gu2022efficiently,
  title     = {{Efficiently Modeling Long Sequences with Structured State Spaces}},
  author    = {Gu, Albert and Goel, Karan and R{\'e}, Christopher},
  booktitle = {International Conference on Learning Representations (ICLR)},
  year      = {2022}
}

@inproceedings{liu2022non,
  title     = {Non-stationary Transformers: Exploring the Stationarity in Time Series Forecasting},
  author    = {Liu, Yong and Wu, Haixu and Wang, Jianmin and Long, Mingsheng},
  booktitle = {Advances in Neural Information Processing Systems (NeurIPS)},
  year      = {2022}
}

@article{woo2022etsformer,
  title   = {{ETSformer: Exponential Smoothing Transformers for Time-series Forecasting}},
  author  = {Woo, Gerald and Liu, Chenghao and Sahoo, Doyen and Kumar, Akshat and Hoi, Steven},
  journal = {arXiv preprint arXiv:2202.01381},
  year    = {2022}
}

@article{lightts,
  title   = {{Less Is More: Fast Multivariate Time Series Forecasting with Light Sampling-oriented MLP Structures}},
  author  = {Zhang, Tianping and Zhang, Yizhuo and Cao, Wei and Bian, Jiang and Yi, Xiaohan and Zheng, Shun and Li, Jian},
  journal = {arXiv preprint arXiv:2207.01186},
  year    = {2022}
}

@article{Reed2002CV,
  author  = {Reed, Gerald F. and Lynn, Frances and Meade, Brian D.},
  title   = {Use of the Coefficient of Variation in Assessing Variability in Quantitative Assays},
  journal = {Clinical and Diagnostic Laboratory Immunology},
  year    = {2002},
}

@inproceedings{su2019omnianomaly,
  title     = {{Robust Anomaly Detection for Multivariate Time Series through Stochastic Recurrent Neural Network}},
  author    = {Su, Ya and Zhao, Youjian and Niu, Chenhao and Liu, Rong and Sun, Wei and Pei, Dan},
  booktitle = {Proceedings of the 25th ACM SIGKDD International Conference on Knowledge Discovery and Data Mining},
  year      = {2019},
}

@inproceedings{mathur2016swat,
  title     = {{SWaT: A Water Treatment Testbed for Research and Training on ICS Security}},
  author    = {Mathur, Aditya P. and Tippenhauer, Nils Ole},
  booktitle = {Proceedings of the International Workshop on Cyber-Physical Systems for Smart Water Networks},
  publisher = {IEEE},
  year      = {2016}
}

@inproceedings{abdulaal2021psm,
  title     = {{Practical Approach to Asynchronous Multivariate Time Series Anomaly Detection and Localization}},
  author    = {Abdulaal, Ahmed and Liu, Zhuanghua and Lancewicki, Tomer},
  booktitle = {Proceedings of the 27th ACM SIGKDD Conference on Knowledge Discovery and Data Mining},
  year      = {2021},
}

@inproceedings{hundman2018nasa,
  title     = {Detecting Spacecraft Anomalies Using LSTMs and Nonparametric Dynamic Thresholding},
  author    = {Hundman, Kyle and Constantinou, Valentino and Laporte, Christopher and Colwell, Ian and S{\"o}derstr{\"o}m, Tom},
  booktitle = {Proceedings of the 24th ACM SIGKDD International Conference on Knowledge Discovery and Data Mining},
  year      = {2018},
  note      = {Datasets: SMAP and MSL},
}

@inproceedings{timesfm,
  author       = {Abhimanyu Das and
                  Weihao Kong and
                  Rajat Sen and
                  Yichen Zhou},
  title        = {A decoder-only foundation model for time-series forecasting},
  booktitle    = {Forty-first International Conference on Machine Learning, {ICML} 2024},
  year         = {2024},
}

@article{chronos1,
  author       = {Abdul Fatir Ansari and
                  Lorenzo Stella and
                  Ali Caner T{\"{u}}rkmen and
                  Xiyuan Zhang and
                  Pedro Mercado and
                  Huibin Shen and
                  Oleksandr Shchur and
                  Syama Sundar Rangapuram and
                  Sebastian Pineda{-}Arango and
                  Shubham Kapoor and
                  Jasper Zschiegner and
                  Danielle C. Maddix and
                  Hao Wang and
                  Michael W. Mahoney and
                  Kari Torkkola and
                  Andrew Gordon Wilson and
                  Michael Bohlke{-}Schneider and
                  Bernie Wang},
  title        = {{Chronos}: Learning the Language of Time Series},
  journal      = {Trans. Mach. Learn. Res.},
  year         = {2024},
}

@inproceedings{timemoe,
  author       = {Xiaoming Shi and
                  Shiyu Wang and
                  Yuqi Nie and
                  Dianqi Li and
                  Zhou Ye and
                  Qingsong Wen and
                  Ming Jin},
  title        = {{Time-MoE}: Billion-Scale Time Series Foundation Models with Mixture
                  of Experts},
  booktitle    = {The Thirteenth International Conference on Learning Representations,
                  {ICLR} 2025},
  year         = {2025},
}

@article{chronos2,
  author       = {Abdul Fatir Ansari and
                  Oleksandr Shchur and
                  Jaris K{\"{u}}ken and
                  Andreas Auer and
                  Boran Han and
                  Pedro Mercado and
                  Syama Sundar Rangapuram and
                  Huibin Shen and
                  Lorenzo Stella and
                  Xiyuan Zhang and
                  Mononito Goswami and
                  Shubham Kapoor and
                  Danielle C. Maddix and
                  Pablo Guerron and
                  Tony Hu and
                  Junming Yin and
                  Nick Erickson and
                  Prateek Mutalik Desai and
                  Hao Wang and
                  Huzefa Rangwala and
                  George Karypis and
                  Yuyang Wang and
                  Michael Bohlke{-}Schneider},
  title        = {{Chronos-2}: From Univariate to Universal Forecasting},
  journal      = {CoRR},
  year         = {2025},
}

@inproceedings{moirai1,
  author       = {Gerald Woo and
                  Chenghao Liu and
                  Akshat Kumar and
                  Caiming Xiong and
                  Silvio Savarese and
                  Doyen Sahoo},
  title        = {Unified Training of Universal Time Series Forecasting Transformers},
  booktitle    = {Forty-first International Conference on Machine Learning, {ICML} 2024},
  year         = {2024},
}

@article{tirex,
  author       = {Andreas Auer and
                  Patrick Podest and
                  Daniel Klotz and
                  Sebastian B{\"{o}}ck and
                  G{\"{u}}nter Klambauer and
                  Sepp Hochreiter},
  title        = {TiRex: Zero-Shot Forecasting Across Long and Short Horizons with Enhanced
                  In-Context Learning},
  journal      = {CoRR},
  year         = {2025},
}

@misc{fvcore,
  author       = {Facebook Research},
  title        = {fvcore: {FAIR} Computer Vision Library},
  year         = {2021},
  howpublished = {\url{https://github.com/facebookresearch/fvcore}},
}

@article{nochumsohn2025super,
  title={Super-Linear: A Lightweight Pretrained Mixture of Linear Experts for Time Series Forecasting},
  author={Nochumsohn, Liran and Marshanski, Raz and Zisling, Hedi and Azencot, Omri},
  journal={arXiv preprint arXiv:2509.15105},
  year={2025}
}

@article{nochumsohn2024beyond,
  title={Beyond data scarcity: A frequency-driven framework for zero-shot forecasting},
  author={Nochumsohn, Liran and Moshkovitz, Michal and Avner, Orly and Di Castro, Dotan and Azencot, Omri},
  journal={arXiv preprint arXiv:2411.15743},
  year={2024}
}

@article{oreshkin2026zero,
  title={Zero-shot Forecasting by Simulation Alone},
  author={Oreshkin, Boris N and Jauhari, Mayank and Selvam, Ravi Kiran and Wolff, Malcolm and Pan, Wenhao and Ramasubramanian, Shankar and Olivares, Kin G and Konstantinova, Tatiana and Potapczynski, Andres and Cao, Mengfei and others},
  journal={arXiv preprint arXiv:2601.00970},
  year={2026}
}

@article{naiman2024utilizing,
  title={Utilizing image transforms and diffusion models for generative modeling of short and long time series},
  author={Naiman, Ilan and Berman, Nimrod and Pemper, Itai and Arbiv, Idan and Fadlon, Gal and Azencot, Omri},
  journal={Advances in Neural Information Processing Systems},
  volume={37},
  pages={121699--121730},
  year={2024}
}

@inproceedings{fadlon025diffusion,
  author={Fadlon, Gal and Arbiv, Idan and Berman, Nimrod and Azencot, Omri},
  title={A Diffusion Model for Regular Time Series Generation from Irregular Data with Completion and Masking},
  booktitle = {Advances in Neural Information Processing Systems (NeurIPS) 39},
  year = {2025}
}

@inproceedings{gonen2025time,
  author={Gonen, Tal and Pemper, Itai and Naiman, Ilan and Berman, Nimrod and Azencot, Omri},
  title={Time Series Generation Under Data Scarcity: A Unified Generative Modeling Approach},
  booktitle = {Advances in Neural Information Processing Systems (NeurIPS) 39},
  year = {2025}
}

@inproceedings{naiman2024generative,
title={Generative Modeling of Regular and Irregular Time Series Data via {Koopman} {VAE}s},
author={Ilan Naiman and N. Benjamin Erichson and Pu Ren and Michael W. Mahoney and Omri Azencot},
booktitle={The Twelfth International Conference on Learning Representations},
year={2024}
}

@article{nochumsohn2025data,
  author       = {Liran Nochumsohn and Omri Azencot},
  title        = {Data Augmentation Policy Search for Long-Term Forecasting},
  journal      = {Trans. Mach. Learn. Res.},
  volume       = {2025},
  year         = {2025}
}
\bibliographystyle{tmlr}

\newpage

\appendix

\section{Appendix: Extended Notes on XCTFormer}

\subsection{Theoretical validity of the proposed activation function}\label{appendix:activation_validity}

\begin{proof}

We show that the proposed AbsAct activation function satisfies the sufficient stability criterion of \citet{saratchandran2025rethinking}, namely that the Frobenius norm of the produced matrix is bounded by the square root of the number of rows:
\[
\|\mathbf{Activation}(A)\|_F \;\le\; \sqrt{N}
\]

Given a matrix \(A=(a_{ij})\in \R^{N\times N}\), and using the activation defined in Section~\ref{sec:method}, we have:
 
 \begin{equation}
\mathbf{AbsAct}\bigg(
\begin{bmatrix}
a_{11} & \cdots & a_{1n}\\
\vdots & \ddots & \vdots \\
a_{n1} & \cdots & a_{nn}
\end{bmatrix}
\bigg)
=
\begin{bmatrix}
\displaystyle \frac{a_{11}}{\sum_{j=1}^n \lvert a_{1j}\rvert} & \cdots &
\displaystyle \frac{a_{1n}}{\sum_{j=1}^n \lvert a_{1j}\rvert}\\[10pt]
\vdots & \ddots & \vdots \\[2pt]
\displaystyle \frac{a_{n1}}{\sum_{j=1}^n \lvert a_{nj}\rvert} & \cdots &
\displaystyle \frac{a_{nn}}{\sum_{j=1}^n \lvert a_{nj}\rvert}
\end{bmatrix}.
\end{equation}

\noindent\textit{Note.} In practice, we add a small positive offset and a denominator stabilizer to prevent division by zero. We set $\tilde a_{ij}=a_{ij}+10^{-4}$ element-wise, then normalize each row by $\sum_{j=1}^n \lvert \tilde a_{ij}\rvert + 10^{-8}$, i.e., use
$\tilde a_{ij}\big/\!\left(\sum_{k=1}^n \lvert \tilde a_{ik}\rvert + 10^{-8}\right)$.
These constants ($10^{-4}$ and $10^{-8}$) are included only for numerical stability and are omitted from the proof for simplicity.

By definition of the Frobenius Norm: 

\begin{align}
\|\mathbf{AbsAct}(A)\|_F^2
&= \sum_{i=1}^{N}\sum_{j=1}^{n}\!\left(\frac{a_{ij}}{\sum_{k=1}^{n}\lvert a_{ik}\rvert}\right)^{\!2} \\
&= \sum_{i=1}^{N} \frac{\sum_{j=1}^{n} a_{ij}^2}{\left(\sum_{k=1}^{n}\lvert a_{ik}\rvert\right)^2} \\
&\le \sum_{i=1}^{N} \frac{\left(\sum_{j=1}^{n}\lvert a_{ij}\rvert\right)^2}{\left(\sum_{k=1}^{n}\lvert a_{ik}\rvert\right)^2}
\quad\text{(since }\sum_j x_j^2 \le (\sum_j |x_j|)^2\text{)} \\
&= \sum_{i=1}^{N} 1 \;=\; N.
\end{align}

Hence, $\|\mathbf{AbsAct}(A)\|_F \le \sqrt{N}$. \\
\textit{Note.} The same bound holds when DeCoP is applied: the matrix still has $N$ rows, and after rowwise $\ell_1$ normalization each row’s Frobenius norm is strictly less than one. Consequently, the sum of the Frobenius norm across all rows does not exceed $N$.

\end{proof}

\subsection{Scaling invariance of AbsAct}\label{appendix:scaling_invariance}

In standard attention, the score matrix is scaled by $1/\sqrt{d_k}$ before applying softmax to keep dot-product magnitudes controlled and avoid overly sharp attention distributions. Under AbsAct, this scaling is unnecessary. Let $\alpha > 0$ be any positive scalar (e.g.\ $\alpha = 1/\sqrt{d_k}$). For any row $i$ with $\sum_{k=1}^{N} |A_{ik}| > 0$, we have for every entry $(i,j)$:
\begin{align}
\text{AbsAct}(\alpha A)_{ij}
&= \frac{\alpha A_{ij}}{\sum_{k=1}^{N} |\alpha A_{ik}|}
= \frac{\alpha A_{ij}}{\alpha \sum_{k=1}^{N} |A_{ik}|}
= \frac{A_{ij}}{\sum_{k=1}^{N} |A_{ik}|}
= \text{AbsAct}(A)_{ij}.
\end{align}

Hence, AbsAct is homogeneous of degree zero for positive scalars:
\[
\text{AbsAct}(\alpha A) = \text{AbsAct}(A), \qquad \forall \alpha > 0,
\]
whenever the row-wise denominator is nonzero. Because the $1/\sqrt{d_k}$ factor is a positive scalar applied uniformly to every entry, it cancels exactly and can be omitted without affecting the output.

\subsection{DeCoP: Complexity Analysis}\label{appendix:DeCoP_complexity_analysis}

We show that DeCoP reduces the attention cost from \emph{quadratic} in the sequence length \(N\) to \emph{linear} (for fixed \(k\ll N\)), thereby lowering both compute and memory costs.

\paragraph{Setup.}
Let \(X\in\mathbb{R}^{N\times D_i}\) and \(W_q,W_k\in\mathbb{R}^{D_i\times D_m}\). Define
\[
Q = X W_q \in \mathbb{R}^{N\times D_m}, \qquad
K = X W_k \in \mathbb{R}^{N\times D_m}.
\]
DeCoP introduces a learnable compression matrix \(C \in \mathbb{R}^{N\times k}\) with \(k\ll N\).

\paragraph{DeCoP computation (no \(N\times N\) intermediate).}
\[
S \;=\; K^\top C \in \mathbb{R}^{D_m\times k},\qquad
A \;=\; Q S \;\in\; \mathbb{R}^{N\times k}.
\]

When DeCoP is applied, the learnable non-boolean mask from CRAB (Sec.~\ref{method:CRAB}) is not used. Since the compressed attention $A_c \in \mathbb{R}^{N \times k}$ does not retain the direct pairwise structure of the full $N \times N$ attention matrix, element-wise masking of individual token-to-token relationships is no longer applicable.

\paragraph{Total complexity (after).}
\begin{align*}
\mathrm{Total}
&= O(N D_i D_m) && \text{(form $Q$)} \\
&\quad + O(N D_i D_m) && \text{(form $K$)} \\
&\quad + O(N D_m k)   && \text{($S = K^\top C$)} \\
&\quad + O(N D_m k)   && \text{($A = Q S$)} \\
&= O\!\big(N[\,2 D_i D_m + 2 D_m k\,]\big)
= O\!\big(2 N D_i D_m + 2 N D_m k\big) \ \\
&\overset{(D_i=D_m)}{=} \boxed{\,O\!\big(N (D_m^2 + D_m k)\big)\,}.
\end{align*}

\paragraph{Growth in \(N\).}
For fixed \(D_i, D_m, k\) with \(k \ll N\), the cost is linear in \(N\).

\paragraph{Memory.}
Store \(Q, K\): \(O(N D_m)\); \(A\) and \(A_{+}\): \(O(N k)\); temporary \(S\): \(O(D_m k)\);
parameters \(C\): \(O(N k)\), \(M\): \(O(N k)\).
No \(N\times N\) matrix is materialized.

\newpage
\section{Appendix: Implementation Details}\label{appendix:experiments_detail}

\subsection{Time-Series Tasks Formulation and Metrics}

This section presents the mathematical formulation and evaluation metrics for three fundamental time-series tasks: long-term forecasting, anomaly detection, and imputation. Each task addresses distinct challenges in temporal data analysis while sharing common architectural foundations.

\subsubsection{Long-Term Forecasting}

Long-term forecasting aims to predict future values of a multivariate time-series given historical observations. Given a historical sequence $\mathbf{X} \in \mathbb{R}^{L \times C}$ where $L$ is the lookback window length and $C$ is the number of channels, the objective is to predict the future sequence $\mathbf{Y} \in \mathbb{R}^{H \times C}$ where $H$ is the prediction horizon.

\textbf{Training Objective:}
The model minimizes the Mean Squared Error (MSE) loss:
\begin{equation}
\mathcal{L}_{forecast} = \frac{1}{N} \sum_{i=1}^{N} ||\mathbf{Y}_i - \hat{\mathbf{Y}}_i||^2
\end{equation}
where $N$ is the number of training samples, $\mathbf{Y}_i$ is the ground truth, and $\hat{\mathbf{Y}}_i$ is the predicted sequence.

\textbf{Evaluation Metrics:}
Performance is assessed using multiple regression metrics:
\begin{itemize}
\item \textbf{Mean Absolute Error (MAE):} $MAE = \frac{1}{NH} \sum_{i=1}^{N} \sum_{t=1}^{H} |\mathbf{Y}_{i,t} - \hat{\mathbf{Y}}_{i,t}|$
\item \textbf{Mean Squared Error (MSE):} $MSE = \frac{1}{NH} \sum_{i=1}^{N} \sum_{t=1}^{H} (\mathbf{Y}_{i,t} - \hat{\mathbf{Y}}_{i,t})^2$
\end{itemize}

\subsubsection{Imputation}

Time-series imputation reconstructs missing values in partially observed sequences. Let $\mathbf{X}\in\mathbb{R}^{L\times C}$ be the ground-truth sequence and $\mathbf{M}\in\{0,1\}^{L\times C}$ a binary mask where $M_{t,c}=0$ marks a missing entry; the observed input is obtained via an element-wise product $\mathbf{X}^{\mathrm{obs}}=\mathbf{X}\odot\mathbf{M}$ .

\textbf{Problem Formulation.}
An imputation model with parameters $\psi$ reconstructs the complete sequence:
\begin{equation}
\widehat{\mathbf{X}}=\operatorname{Imputer}_{\psi}\!\left(\mathbf{X}^{\mathrm{obs}},\,\mathbf{M}\right).
\end{equation}

\textbf{Training Objective.}
During training, artificial masks are sampled with mask rate $p$. The reconstruction loss is computed \emph{only on masked positions}:
\begin{equation}
\mathcal{L}_{\mathrm{impute}}
=\frac{1}{|\mathcal{M}|}\sum_{(t,c)\in\mathcal{M}}\bigl(\mathbf{X}_{t,c}-\widehat{\mathbf{X}}_{t,c}\bigr)^2,
\qquad
\mathcal{M}=\{(t,c): M_{t,c}=0\}.
\end{equation}

\textbf{Evaluation Protocol.}
\begin{enumerate}
\item Apply a random mask (rate $p$) to test sequences to obtain $\mathbf{X}^{\mathrm{obs}}$ and $\mathbf{M}$.
\item Impute the missing entries:
$\mathbf{X}^{\mathrm{filled}}=\mathbf{X}^{\mathrm{obs}}\odot\mathbf{M}+\widehat{\mathbf{X}}\odot(1-\mathbf{M})$.
\item Compute metrics \emph{exclusively on the masked set} $\mathcal{M}$.
\end{enumerate}

\textbf{Evaluation Metrics (masked-only).}
All metrics are calculated \emph{only on masked values} $(t,c)\in\mathcal{M}$:
\begin{align}
\mathrm{MAE}_{\mathrm{mask}}&=\frac{1}{|\mathcal{M}|}\sum_{(t,c)\in\mathcal{M}}\bigl|\mathbf{X}_{t,c}-\widehat{\mathbf{X}}_{t,c}\bigr|,\\
\mathrm{MSE}_{\mathrm{mask}}&=\frac{1}{|\mathcal{M}|}\sum_{(t,c)\in\mathcal{M}}\bigl(\mathbf{X}_{t,c}-\widehat{\mathbf{X}}_{t,c}\bigr)^2.
\end{align}

\subsubsection{Anomaly Detection}

Anomaly detection identifies unusual patterns or outliers in time-series data using a reconstruction-based approach. The model learns to reconstruct normal patterns and flags samples with high reconstruction errors as anomalous.

\textbf{Problem Formulation:}
Given a time-series $\mathbf{X} \in \mathbb{R}^{L \times C}$, the model $g_\phi$ learns to reconstruct the input:
\begin{equation}
\hat{\mathbf{X}} = g_\phi(\mathbf{X})
\end{equation}
The anomaly score is computed as the reconstruction error: $s = ||\mathbf{X} - \hat{\mathbf{X}}||^2$

\textbf{Training Objective:}
The model is trained exclusively on normal data using reconstruction loss:
\begin{equation}
\mathcal{L}_{recon} = \frac{1}{N} \sum_{i=1}^{N} ||\mathbf{X}_i - g_\phi(\mathbf{X}_i)||^2
\end{equation}

\textbf{Detection Mechanism:}
We use a percentile-based threshold over the pooled anomaly-score distribution. Let
\(\mathcal{S}=\{s_{\text{train}}\}\cup\{s_{\text{test}}\}\).
For a target anomaly rate \(\alpha\), the threshold is
\[
\tau=\operatorname{quantile}(\mathcal{S},\,1-\alpha).
\]
In our experiments we set \(\alpha=0.01\) (1\%) for all datasets, except SMD \citep{su2019omnianomaly}, where \(\alpha=0.005\) (0.5\%).
A sample with score \(s\) is labeled as
\[
\text{label}=
\begin{cases}
1, & \text{if } s>\tau \ \text{(anomaly)},\\
0, & \text{if } s\le \tau \ \text{(normal)}.
\end{cases}
\]

\textbf{Evaluation Metrics:}
Performance is measured using binary classification metrics:
\begin{itemize}
\item \textbf{Precision:} $P = \frac{TP}{TP + FP}$ (proportion of correctly identified anomalies)
\item \textbf{Recall:} $R = \frac{TP}{TP + FN}$ (proportion of actual anomalies detected)
\item \textbf{F1-Score:} $F_1 = 2 \cdot \frac{P \times R}{P + R}$ (harmonic mean of precision and recall)
\item \textbf{Accuracy:} $Acc = \frac{TP + TN}{TP + TN + FP + FN}$ (overall classification correctness)
\end{itemize}
where $TP$, $TN$, $FP$, and $FN$ represent true positives, true negatives, false positives, and false negatives, respectively.

\subsection{Experiment Datasets And Evaluation Setups}

We evaluate long-term forecasting on seven widely used multivariate datasets (Weather, Electricity, Traffic, ETTh1, ETTh2, ETTm1, ETTm2). For forecasting, we follow the TimesNet \citep{wu2022timesnet} setup with a look-back window $L=96$ and horizons $H\in\{96,192,336,720\}$; dataset specifications appear in Table \ref{tab:long_term_forecasting_dataset}. For time-series imputation, we use the same datasets as forecasting, except for Traffic, and follow the TimeMixer++ \citep{wang2024timemixer++} setup with $L=1024$ and masking ratios $p\in\{12.5\%,25\%,37.5\%,50\%\}$, refer to Table \ref{tab:imputation_datasets}. Anomaly detection focuses on identifying fine-grained patterns. To assess this, we selected the following datasets: SMD (Server Machine Dataset, \citep{su2019omnianomaly}), SWaT (Secure Water Treatment, \citep{mathur2016swat}), PSM (Pooled Server Metrics, \citep{abdulaal2021psm}), and NASA telemetry datasets MSL and SMAP \citep{hundman2018nasa}. The details of the datasets used for anomaly detection are provided in Table \ref{tab:anomaly_detection_datasets}. For long-term forecasting, we additionally evaluate on a synthetic dataset where the target is derived from lagged cross-channel signals; full construction details appear in Appendix~\ref{appendix:synthetic_dataset}.

\begin{table}[thbp]
  \caption{Benchmark datasets and evaluation settings for long-term forecasting.}\label{tab:long_term_forecasting_dataset}
  \vskip 0.1in
  \centering
  \resizebox{1.0\columnwidth}{!}{
  \begin{threeparttable}
  \begin{small}
  \renewcommand{\multirowsetup}{\centering}
  \setlength{\tabcolsep}{3.8pt}
  \begin{tabular}{l|c|c|c|c|c|c}
    \toprule
    Dataset & Dim & Look-back & Prediction Horizons & Dataset Size & Frequency & \scalebox{0.8}{Information} \\
    \midrule
    ETTm1       & 7   & \scalebox{0.8}{96}  & \scalebox{0.8}{\{96, 192, 336, 720\}} & (34465, 11521, 11521) & 15 min & \scalebox{0.8}{Temperature} \\
    ETTm2       & 7   & \scalebox{0.8}{96}  & \scalebox{0.8}{\{96, 192, 336, 720\}} & (34465, 11521, 11521) & 15 min & \scalebox{0.8}{Temperature} \\
    ETTh1       & 7   & \scalebox{0.8}{96}  & \scalebox{0.8}{\{96, 192, 336, 720\}} & (8545, 2881, 2881)    & 15 min & \scalebox{0.8}{Temperature} \\
    ETTh2       & 7   & \scalebox{0.8}{96}  & \scalebox{0.8}{\{96, 192, 336, 720\}} & (8545, 2881, 2881)    & 15 min & \scalebox{0.8}{Temperature} \\
    Weather     & 21  & \scalebox{0.8}{96}  & \scalebox{0.8}{\{96, 192, 336, 720\}} & (36792, 5271, 10540)  & 10 min & \scalebox{0.8}{Weather} \\
    ECL         & 321 & \scalebox{0.8}{96}  & \scalebox{0.8}{\{96, 192, 336, 720\}} & (18317, 2633, 5261)   & Hourly & \scalebox{0.8}{Electricity} \\
    Traffic     & 862 & \scalebox{0.8}{96}  & \scalebox{0.8}{\{96, 192, 336, 720\}} & (12185, 1757, 3509)   & Hourly & \scalebox{0.8}{Transportation} \\
    \bottomrule
  \end{tabular}
  \end{small}
  \end{threeparttable}
  }
\end{table}

\begin{table}[thbp]
  \caption{Benchmark datasets and evaluation settings for time-series imputation.}\label{tab:imputation_datasets}
  \vskip 0.1in
  \centering
  \resizebox{1.0\columnwidth}{!}{
  \begin{threeparttable}
  \begin{small}
  \renewcommand{\multirowsetup}{\centering}
  \setlength{\tabcolsep}{3.8pt}
  \begin{tabular}{l|c|c|c|c|c|c}
    \toprule
    Dataset & Dim & Look-back & \scalebox{0.8}{Imputation Mask Ratios} & Dataset Size & Frequency & \scalebox{0.8}{Information} \\
    \midrule
    ETTm1       & 7   & \scalebox{0.8}{1024} & \scalebox{0.8}{[12.5\%, 25\%, 37.5\%, 50\%]} & (34465, 11521, 11521) & 15 min & \scalebox{0.8}{Temperature} \\
    ETTm2       & 7   & \scalebox{0.8}{1024} & \scalebox{0.8}{[12.5\%, 25\%, 37.5\%, 50\%]} & (34465, 11521, 11521) & 15 min & \scalebox{0.8}{Temperature} \\
    ETTh1       & 7   & \scalebox{0.8}{1024} & \scalebox{0.8}{[12.5\%, 25\%, 37.5\%, 50\%]} & (8545, 2881, 2881)    & 15 min & \scalebox{0.8}{Temperature} \\
    ETTh2       & 7   & \scalebox{0.8}{1024} & \scalebox{0.8}{[12.5\%, 25\%, 37.5\%, 50\%]} & (8545, 2881, 2881)    & 15 min & \scalebox{0.8}{Temperature} \\
    Weather     & 21  & \scalebox{0.8}{1024} & \scalebox{0.8}{[12.5\%, 25\%, 37.5\%, 50\%]} & (36792, 5271, 10540)  & 10 min & \scalebox{0.8}{Weather} \\
    ECL         & 321 & \scalebox{0.8}{1024} & \scalebox{0.8}{[12.5\%, 25\%, 37.5\%, 50\%]} & (18317, 2633, 5261)   & Hourly & \scalebox{0.8}{Electricity} \\
    \bottomrule
  \end{tabular}
  \end{small}
  \end{threeparttable}
  }
\end{table}

\begin{table}[thbp]
  \caption{Dataset detailed descriptions for anomaly detection. The dataset size is organized in (Train, Validation, Test).}\label{tab:anomaly_detection_datasets}
  \vskip 0.1in
  \centering
  \resizebox{1.0\columnwidth}{!}{
  \begin{threeparttable}
  \begin{small}
  \renewcommand{\multirowsetup}{\centering}
  \setlength{\tabcolsep}{3.8pt}
  \begin{tabular}{l|c|c|c|c}
    \toprule
    Dataset & Dim & Series Length & Dataset Size & \scalebox{0.8}{Information} \\
    \midrule
    SMD  & 38 & \scalebox{0.8}{100} & (566724, 141681, 708420) & \scalebox{0.8}{Server machines} \\
    MSL  & 55 & \scalebox{0.8}{100} & (44653, 11664, 73729)    & \scalebox{0.8}{Spacecraft telemetry (Mars)} \\
    SMAP & 25 & \scalebox{0.8}{100} & (108146, 27037, 427617)  & \scalebox{0.8}{Spacecraft telemetry} \\
    SWaT & 51 & \scalebox{0.8}{100} & (396000, 99000, 449919)  & \scalebox{0.8}{Water treatment ICS} \\
    PSM  & 25 & \scalebox{0.8}{100} & (105984, 26497, 87841)   & \scalebox{0.8}{Server metrics} \\
    \bottomrule
  \end{tabular}
  \end{small}
  \end{threeparttable}
  }
\end{table}

\subsection{Training Details.} \label{appendix:hyperparameters}
All experiments were implemented in PyTorch \citep{paszke2019pytorch} and run on NVIDIA RTX 3090 GPUs. We fix the random seed to 2021. We applied a RevIN transformation \citep{kim2022reversible} to mitigate distributional shifts in the data. Models are trained with the Adam optimizer \citep{kingma2015adam} using mean squared error (MSE) loss, together with a \texttt{OneCycleLR} scheduler. At a high level, \texttt{OneCycleLR} first increases the learning rate from its initial value to a peak (while inversely adjusting momentum), and then gradually anneals it to a small value for the remainder of training; this promotes fast early progress and stable late-stage convergence. We set \texttt{pct\_start} \(= 0.4\), allocating 40\% of the total training steps to the warm-up/increase phase and 60\% to the annealing phase. For each run, we select the checkpoint with the lowest validation MSE and report the corresponding test performance in the tables. 

\paragraph{Hyper-Parameter Search.} We tuned hyperparameters with Optuna and chose the configuration yielding the lowest validation \textbf{MSE}. The selected configuration is tuned per dataset and task, and then kept fixed across all forecasting horizons or imputation mask ratios, as reported in Table~\ref{tab:hyperparameters}.

\begin{table}[htbp]
\centering
\caption{Hyperparameter settings for \textbf{XCTFormer} per dataset per time-series task}
\label{tab:hyperparameters}
\resizebox{\textwidth}{!}{%
\begin{tabular}{|l|c|c|c|c|c|c|c|c|c|c|c|c|c|}
\hline
\multicolumn{1}{|l|}{} &
\multicolumn{2}{c|}{\textbf{Data Processing}} &
\multicolumn{6}{c|}{\textbf{Transformer}} &
\multicolumn{2}{c|}{\textbf{XCTFormer}} &
\multicolumn{3}{c|}{\textbf{Training}} \\
\hline
\textbf{Dataset} &
\textbf{patch\_len} &
\textbf{stride} &
\textbf{e\_layers} &
\textbf{n\_heads} &
\textbf{d\_model} &
\textbf{d\_ff} &
\textbf{dropout} &
\textbf{fc\_dropout} &
\textbf{attn\_dropout} &
\textbf{k} &
\textbf{batch\_size} &
\textbf{learning\_rate} &
\textbf{epochs} \\
\hline
\multicolumn{14}{|c|}{\textbf{Long-term time-series Forecasting}} \\
\hline
ETTh1       & 16 & 8  & 1 & 1 & 8   & 16  & 0.2 & 0.3  & 0.6 & -   & 32 & 0.001  & 10 \\
ETTh2       & 16 & 8  & 3 & 1 & 30  & 60  & 0.1 & 0.2  & 0.8 & -   & 32 & 0.01   & 10 \\
ETTm1       & 16 & 8  & 2 & 4 & 32  & 64  & 0.1 & 0.05 & 0.8 & -   & 32 & 0.005  & 10 \\
ETTm2       & 16 & 8  & 1 & 1 & 224 & 448 & 0.1 & 0.05 & 0.8 & -   & 32 & 0.005  & 10 \\
Weather     & 16 & 8  & 3 & 2 & 248 & 496 & 0.1 & 0.05 & 0.8 & -   & 32 & 0.0005 & 10 \\
Traffic     & 16 & 8  & 3 & 4 & 248 & 496 & 0.1 & 0.05 & 0.6 & 192 & 8  & 0.001  & 10 \\
ECL         & 16 & 8  & 3 & 1 & 248 & 496 & 0.1 & 0.05 & 0.5 & 64  & 32 & 0.005  & 10 \\
\hline
\multicolumn{14}{|c|}{\textbf{Imputation}} \\
\hline
ETTh1       & 16 & 8  & 2 & 1 & 64  & 128 & 0.1 & 0.05 & 0.5 & -   & 32 & 0.01  & 10 \\
ETTh2       & 64 & 32 & 3 & 1 & 160 & 320 & 0.1 & 0.05 & 0.3 & -   & 32 & 0.005 & 10 \\
ETTm1       & 16 & 8  & 3 & 4 & 96  & 192 & 0.1 & 0.05 & 0.1 & -   & 32 & 0.005 & 10 \\
ETTm2       & 16 & 8  & 2 & 1 & 128 & 256 & 0.1 & 0.05 & 0.5 & -   & 32 & 0.001 & 10 \\
Weather     & 16 & 8  & 3 & 1 & 192 & 384 & 0.1 & 0.05 & 0.8 & -   & 32 & 0.001 & 10 \\
ECL         & 64 & 32 & 2 & 2 & 192 & 384 & 0.1 & 0.05 & 0.7 & 128 & 32 & 0.005 & 10 \\
\hline
\multicolumn{14}{|c|}{\textbf{Anomaly Detection}} \\
\hline
MSL  & 16 & 8  & 2 & 4 & 256 & 512 & 0.1 & 0.05 & 0.7 & - & 128 & 0.01   & 10 \\
PSM  & 16 & 8  & 2 & 1 & 256 & 512 & 0.1 & 0.05 & 0.8 & - & 128 & 0.001  & 10 \\
SMAP & 16 & 8  & 3 & 1 & 256 & 128 & 0.1 & 0.05 & 0.3 & - & 128 & 0.005  & 10 \\
SMD  & 16 & 8  & 2 & 1 & 168 & 336 & 0.1 & 0.05 & 0.3 & - & 128 & 0.001  & 10 \\
SWaT & 16 & 8  & 1 & 2 & 216 & 432 & 0.1 & 0.05 & 0.4 & - & 128 & 0.0005 & 10 \\
\hline
\end{tabular}%
}
\end{table}

\paragraph{Hyper-Parameter Search Space.} We searched over a bounded hyperparameter space per dataset and task, while fixing a few coupled settings to reduce degrees of freedom (we set stride = patch\_len/2 and d\_ff = 2 × d\_model). For ETTh1/ETTh2 in forecasting, where we observed stronger overfitting, we narrowed the d\_model range and explicitly tuned dropout to improve generalization. The search area is represented in Table \ref{tab:hyperparameter_search_space}.

\begin{table}[htbp]
\centering
\caption{Hyperparameter Search Space}
\label{tab:hyperparameter_search_space}
\resizebox{\textwidth}{!}{%
\begin{tabular}{ll|cccc|ccccc}
\toprule
\textbf{Task} & \textbf{Datasets} & \textbf{lr} & \textbf{att\_dropout} & \textbf{n\_heads} & \textbf{e\_layers} & \textbf{d\_model} & \textbf{patch\_len} & \textbf{dropout} & \textbf{fc\_dropout} & \textbf{k} \\
\midrule
\multirow{2}{*}{Long-term Forecast} 
    & ETTh1, ETTh2 & \multirow{4}{*}{\shortstack{$\{5\text{e-}4, 1\text{e-}3,$\\$5\text{e-}3, 1\text{e-}2\}$}} & \multirow{4}{*}{$[0.1, 0.8]_{0.1}$} & \multirow{4}{*}{$\{1, 2, 4\}$} & \multirow{4}{*}{$[1, 3]$} & $[4, 64]_2$ & -- & $[0.1, 0.3]_{0.05}$ & $[0.05, 0.3]_{0.05}$ & -- \\
    & Others &  &  &  &  & $[8, 256]_8$ & -- & -- & -- & $[64, 256]_{64}^\dagger$ \\
\cmidrule{1-2} \cmidrule{7-11}
Imputation & All &  &  &  &  & $[32, 256]_{32}$ & $\{16, 64, 128\}$ & -- & -- & $[64, 256]_{64}^\dagger$ \\
\cmidrule{1-2} \cmidrule{7-11}
Anomaly Detection & All &  &  &  &  & $[8, 256]_8$ & -- & -- & -- & -- \\
\bottomrule
\end{tabular}%
}
\vspace{0.5em}
\footnotesize
\textbf{Notation:} $[a, b]_s$ denotes integer/float range from $a$ to $b$ with step $s$; $\{...\}$ denotes categorical choices; -- indicates parameter not used.
$^\dagger$\textbf{k} was searched only for ECL and Traffic datasets.
\end{table}

\subsection{Baseline Implementation Details}
For baselines evaluated under the same experimental setting as our main study, we directly used the reported results from the TimeMixer++ \cite{wang2024timemixer++} paper when available, and otherwise reported from the corresponding original papers.

For the synthetic dataset, all compared models were trained using the ETTm1 hyperparameters. Each model was run using the code and hyperparameters from its official repository where available; otherwise, the implementation was taken from \url{https://github.com/thuml/Time-Series-Library}.

\subsection{Technical Evaluation Note}
We compute final metrics for imputation and forecasting as weighted averages across batches to account for varying batch sizes during evaluation. This adjustment is necessary because the last batch in an epoch may contain fewer samples than the standard batch size. When computing performance metrics by simply averaging across batches without considering batch sizes, smaller batches receive disproportionate weight in the final metric calculation, leading to biased performance estimates that do not accurately reflect true model performance across the entire dataset. In many older works, researchers addressed this problem by setting the drop\_last=True parameter in PyTorch's DataLoader, which discards the final incomplete batch to ensure identical batch sizes. However, this approach wastes data and can be particularly problematic for smaller datasets, where discarding samples reduces available training or evaluation data. In recent works, it is more common to solve this problem by setting drop\_last=False and computing weighted averages, where each batch's metric contribution is weighted by its actual size, ensuring that the final averaged metric accurately represents performance across all samples in the dataset without discarding any data.

\subsection{Computational Profiling}\label{appendix:computational_profiling}
To assess the computational cost of each model, we measure both floating-point operations (FLOPs) and trainable parameter counts across all datasets. Full per-dataset results are reported in Tables~\ref{tab:flops} and~\ref{tab:params}.

\paragraph{FLOPs.}
For all real-world datasets, FLOPs are computed using \texttt{fvcore}'s \texttt{FlopCountAnalysis}~\cite{fvcore}, which performs a symbolic forward pass to trace the model's computation graph and counts multiply-accumulate operations (MACs) across all layers, including custom modules.
\paragraph{Trainable parameters.}
Parameter counts are obtained via PyTorch's native enumeration: \texttt{sum(p.numel() for p in model.parameters() if p.requires\_grad)}, ensuring all learnable weights are captured regardless of layer type.

\paragraph{Excluded models.}
TimeMixer++ is excluded because its source code is not publicly available. Crossformer values on ETTh2 and ETTm2 are unavailable (N/A) due to configuration incompatibility with those datasets.

\newpage
\section{Appendix: Extended Analysis }
\subsection{Interpretable Learned Masks Analysis.} \label{appendix:mask_analysis}

This analysis applies to configurations without DeCoP, where the learnable mask is active (see Sec.~\ref{method:DeCoP}).

CRAB (Sec. \ref{method:CRAB}) introduces a learnable non-boolean mask that learns the most dominant cross-channel and temporal dependencies. The mask learns dominant dependencies by directly modulating the strength of attention values during training. Specifically, the mask multiplies attention weights element-wise, with higher absolute values amplifying the corresponding attention relationships and values near zero suppressing them. Through gradient-based optimization, the mask automatically discovers which cross-channel and temporal interactions are most informative for the downstream task, effectively learning a data-driven weighting scheme that prioritizes the most predictive dependencies.

Examining the learned mask can therefore provide data-specific insights about these dependency structures, as the mask values directly reflect relationship dominance, with higher absolute values indicating stronger learned dependencies. In this section, we explain how to interpret the attention mask as a foundation for further analysis so it can be leveraged for different analytical needs. 

\begin{figure}[ht]
  \centering
  \includegraphics[page=1,width=\textwidth]{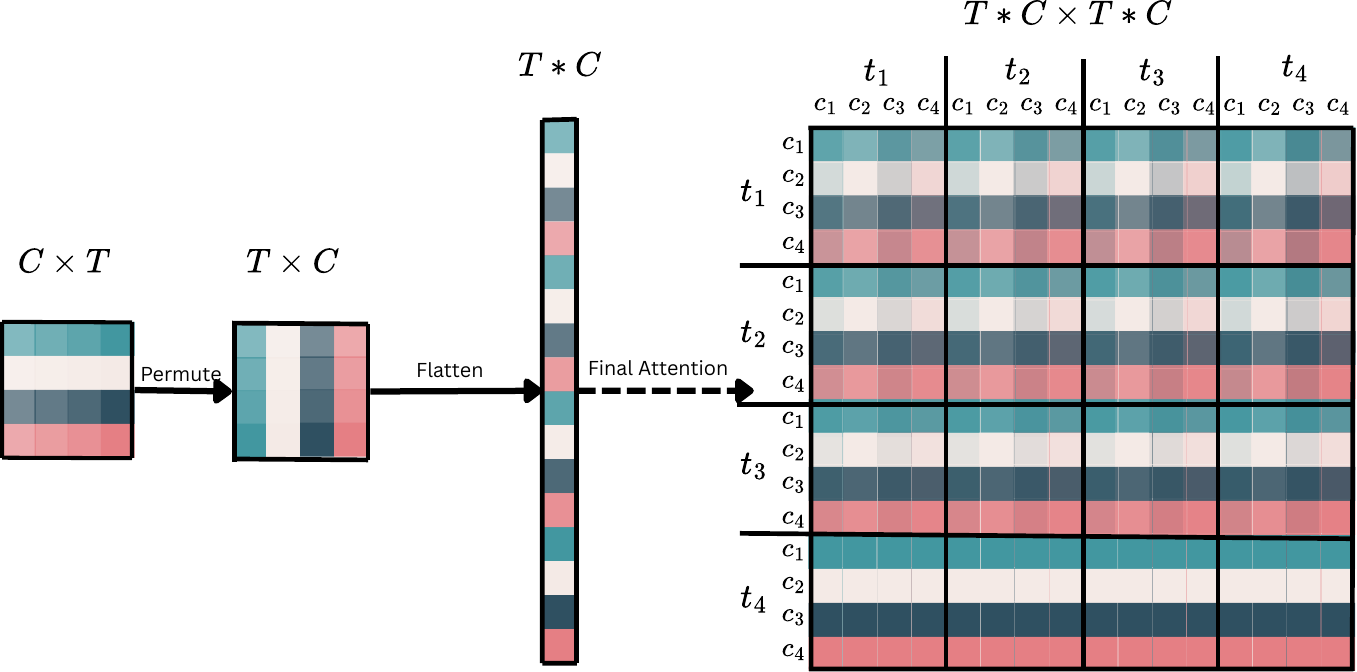}
  \vspace{-5mm}
\caption{Interpretable Learned Mask Structure: The data permutation step places the patch sequence dimension first, creating an attention mask that can be visualized as a grid of squares where each square represents cross-channel relationships between pairs of time-steps. Note: batch and data dimensions are excluded from this diagram for clarity.}
  \label{fig:attention_ordering_visualization}
\end{figure}

Following data processing (Sec. \ref{method:data_processing}), the input data is first permuted so that the patch-sequence dimension is placed before the channel dimension, and then the sequence and channel dimensions are flattened. This creates an attention mask structure that can be visualized as a grid of squares where each square represents cross-channel relationships between pairs of time steps (refer to Figure \ref{fig:attention_ordering_visualization} for visual representation), with the main diagonal squares capturing cross-channel interactions within the same time step and off-diagonal squares revealing temporal cross-channel dependencies.

\paragraph{Analysis of Learnable Masks on ETTm1 Dataset}

\begin{figure}[ht]
\centering
\includegraphics[page=1,width=\textwidth]{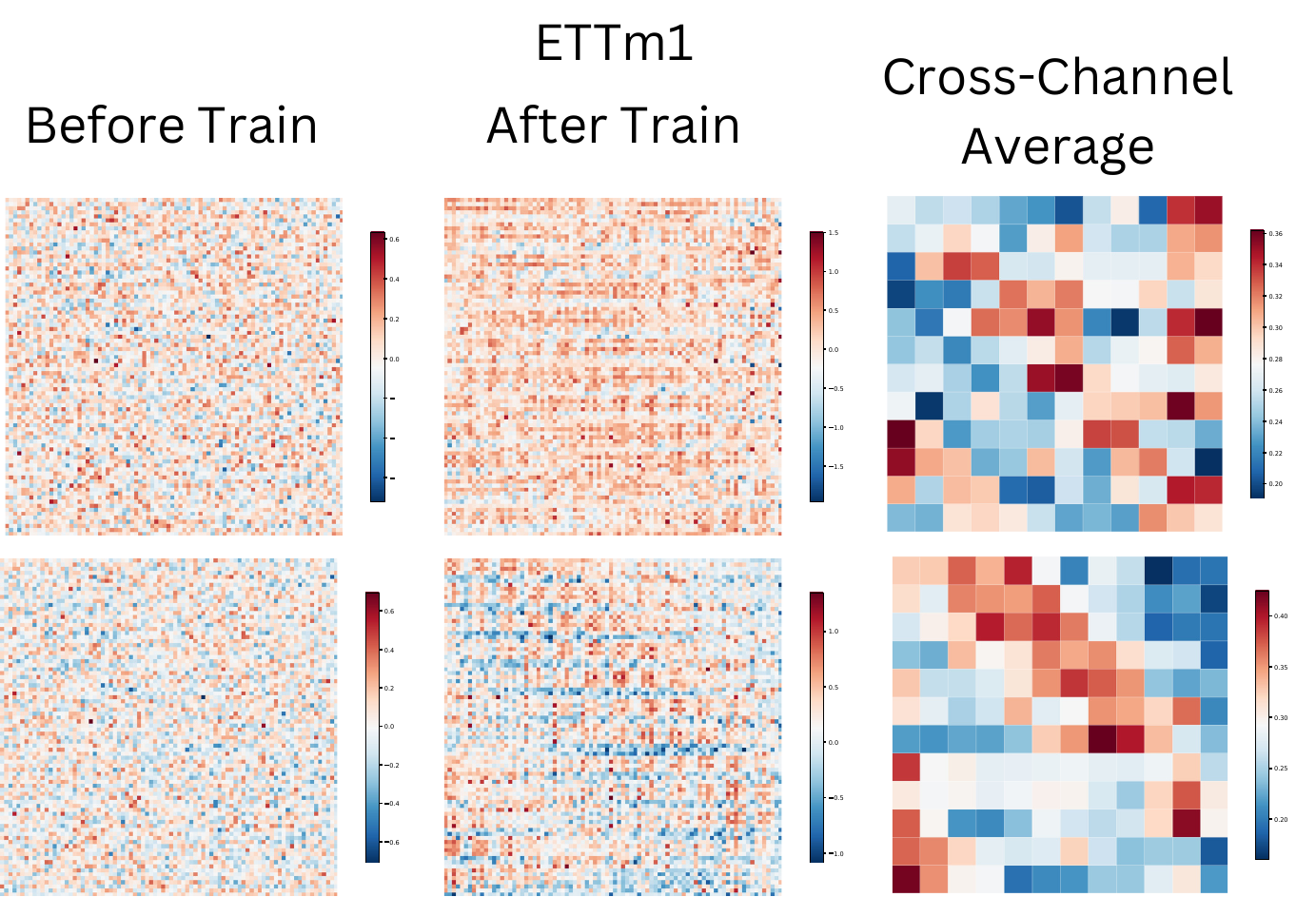}
\vspace{-5mm}
\caption{Analysis of learnable attention masks on ETTm1 dataset. Top row: 96$\rightarrow$96 forecasting; bottom row: 96$\rightarrow$192 forecasting. Left column: initial random masks; middle column: learned structured patterns after training; right column: heatmaps of cross-channel dependency strength derived from trained masks. The heatmaps visualize the strength of cross-channel dependencies between time points, with darker red regions indicating stronger relationships.}
\label{fig:ettm1_mask_analysis}
\end{figure}

Figure \ref{fig:ettm1_mask_analysis} analyzes learnable attention masks trained on the ETTm1 dataset across two forecasting scenarios: 96$\rightarrow$96 (top row) and 96$\rightarrow$192 (bottom row). Each row displays three visualizations: initial random masks initiated from a normal distribution (left), learned patterns after training (middle), and corresponding heatmaps quantifying cross-channel dependency strength (right).

\textbf{Data Processing and Architecture.} The ETTm1 dataset was processed using patches of length 16 with a stride of 8, generating 12 patches across ETTm1's 7 channels. This configuration produces an $84 \times 84$ attention matrix ($12 \times 7 = 84$ dimensions) that captures both temporal and cross-channel relationships.

\textbf{Heatmap Interpretation.} The dependency strength heatmaps are derived from the trained masks by averaging the absolute values within each cross-channel grid. Since masks are applied to attention weights, higher absolute mask values correspond to more dominant dependencies, with darker red regions in the heatmap indicating stronger cross-channel relationships between specific time steps.

\textbf{Key Findings.} The trained masks exhibit several notable patterns. First, they develop structured grid formations that align precisely with the 12-patch architecture, suggesting the model learns systematic cross-channel dependencies. By examining the heatmaps from both configurations, we observe a high density of dominant dependencies along the main diagonal. This diagonal concentration indicates that the model learns strong self-attention patterns, in which each time step primarily attends to itself and its immediate temporal neighbors. Such patterns suggest that the most informative relationships for forecasting are local temporal dependencies, in which recent observations carry the greatest predictive power for future values. This finding aligns with the intuitive understanding that in time-series analysis, temporally proximate data points are typically more relevant than distant historical information.

\subsection{DeCoP compression size analysis} \label{appendix:decop_k_analysis} 

As recalled, DeCoP compresses token-to-token interactions into a low-dimensional representation; the choice of $k$ directly controls the expressive capacity of this bottleneck and thus can affect both accuracy and efficiency. In particular, larger $k$ increases the dimensionality of the compressed attention embedding, which summarizes each token’s pairwise interactions with all other tokens, enabling richer dependency modeling. However, this comes at a higher compute and memory cost. Therefore, analyzing sensitivity to $k$ is important for providing practical guidance on choosing $k$ that accounts for resource constraints.

\begin{figure}
    \centering
    \includegraphics[width=1\linewidth]{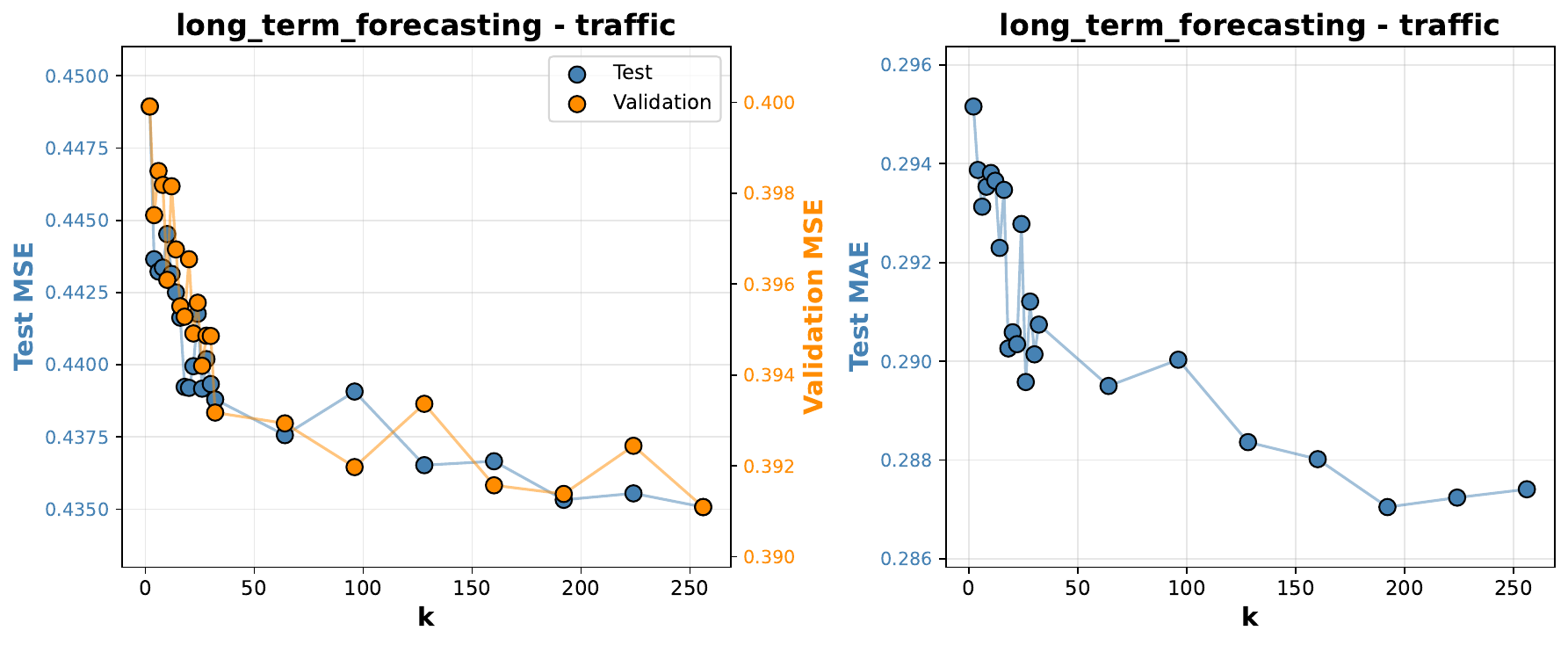}
    \caption{\textbf{DeCoP $k$ sensitivity on Traffic (forecasting).} Average MAE/MSE across horizons $\{96,192,336,720\}$ for different compressed representation sizes $k$.}
    \label{fig:decop_k_forecasting_traffic}
\end{figure}

\begin{figure}
    \centering
    \includegraphics[width=1\linewidth]{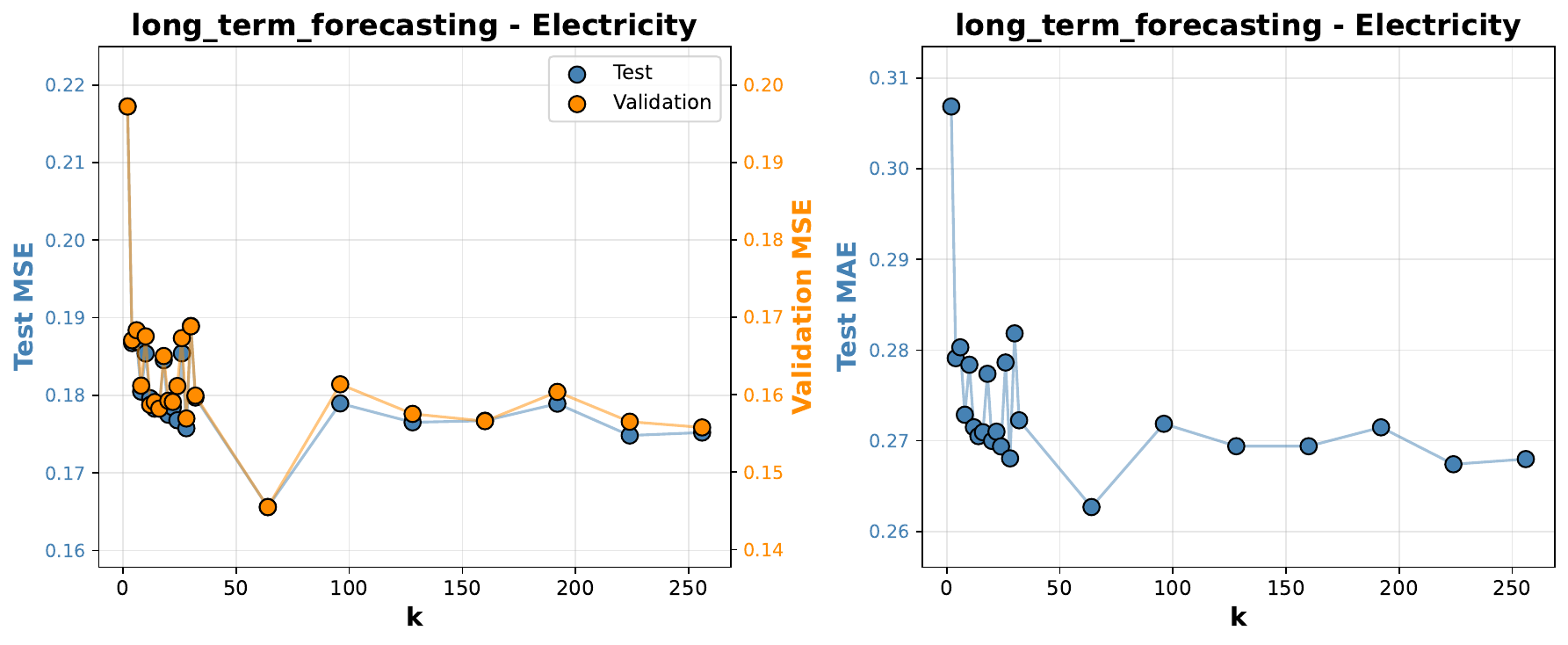}
    \caption{\textbf{DeCoP $k$ sensitivity on ECL (forecasting).} Average MAE/MSE across horizons $\{96,192,336,720\}$ for different compressed representation sizes $k$.}
    \label{fig:decop_k_forecasting_electricity}
\end{figure}

\begin{figure}
    \centering
    \includegraphics[width=1\linewidth]{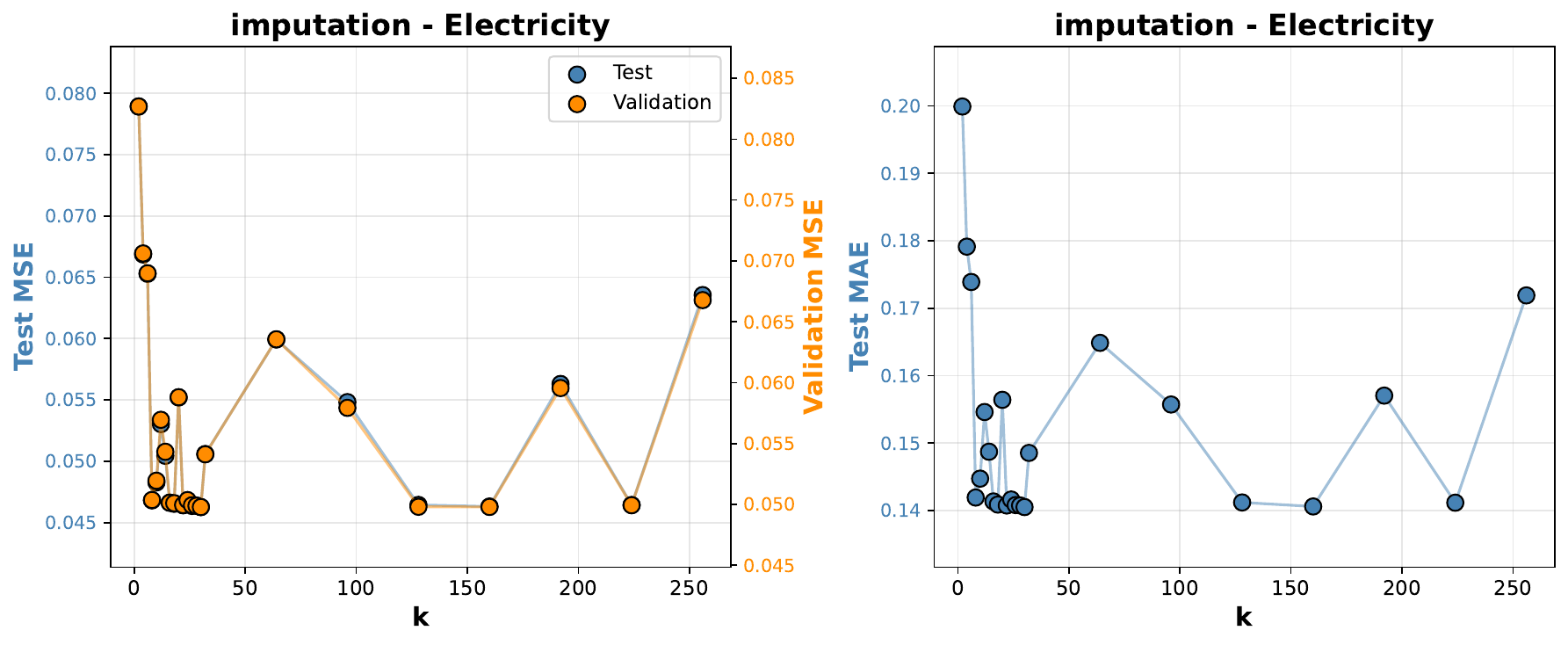}
    \caption{\textbf{DeCoP $k$ sensitivity on ECL (imputation).} Average MAE/MSE across mask ratios $\{0.125,0.25,0.375,0.5\}$ for different compressed representation sizes $k$.}
    \label{fig:decop_k_imputation_electricity}
\end{figure}

Concretely, we tested $k \in [2, 32]$ (step 2) and $k \in [64, 256]$ (step 32). For each $k$, we report the average MAE/MSE across forecasting horizons ${96,192,336,720}$ and the average MAE/MSE across imputation mask ratios ${0.125,0.25,0.375,0.5}$. Overall, we observe dataset-dependent behavior. On \textbf{Traffic} (forecasting analysis Table \ref{fig:decop_k_forecasting_traffic}), performance tends to improve as $k$ increases, indicating that a higher-capacity compressed representation better captures the complex multivariate dependencies in this dataset. In contrast, on \textbf{ECL} (forecasting analysis Table \ref{fig:decop_k_forecasting_electricity}, imputation analysis Table \ref{fig:decop_k_imputation_electricity}), smaller or mid-range $k$ values are often competitive and occasionally slightly better, suggesting that stronger compression can provide a useful regularizing effect and that further increasing $k$ yields limited additional benefit. Based on these results, we recommend treating $k$ as a \emph{dataset-specific} hyperparameter and tuning it to balance accuracy with compute and memory costs, as larger $k$ values often yield only marginal gains.

\subsection{Patch length and stride analysis.} \label{appendix:patch_len_stride_analysis}

Patching determines how the input sequence is divided into fixed-length windows, where patch\_len sets the window size, and each window is projected into a token embedding. The stride controls the overlap between consecutive windows and thus the density with which the sequence is covered. Together, they affect which information is represented in each token and how many tokens the transformer processes. With a larger \texttt{patch\_len} or stride, the model uses fewer tokens, but each token must represent a longer window, which can make it harder to preserve meaningful temporal information. With a smaller \texttt{patch\_len} or stride, tokens represent more local information and overlap increases, but the longer token sequence raises the computational cost of attention. Therefore, \texttt{patch\_len} and stride define a fundamental accuracy efficiency trade-off. This sensitivity analysis evaluates how robust our method is to this configuration and if 
dataset-specific tuning may be required.

Following PatchTST’s \citep{nie2023time} best practice, we kept the patching configuration fixed in most of our main experiments. For forecasting and anomaly detection, we used \texttt{patch\_len}$=16$ and \texttt{stride}$=8$. For imputation, where the lookback was substantially larger ($L{=}1024$ versus 96 in forecasting and 100 in anomaly detection), we tested a small set of \texttt{patch\_len} values ${16,64,128}$ with the corresponding \texttt{stride}$=\texttt{patch\_len}/2$. Based on validation error during hyperparameter search, we selected \texttt{patch\_len}$=64$ and \texttt{stride}$=32$ for some imputation datasets.

\begin{figure}
    \centering
    \includegraphics[width=1\linewidth]{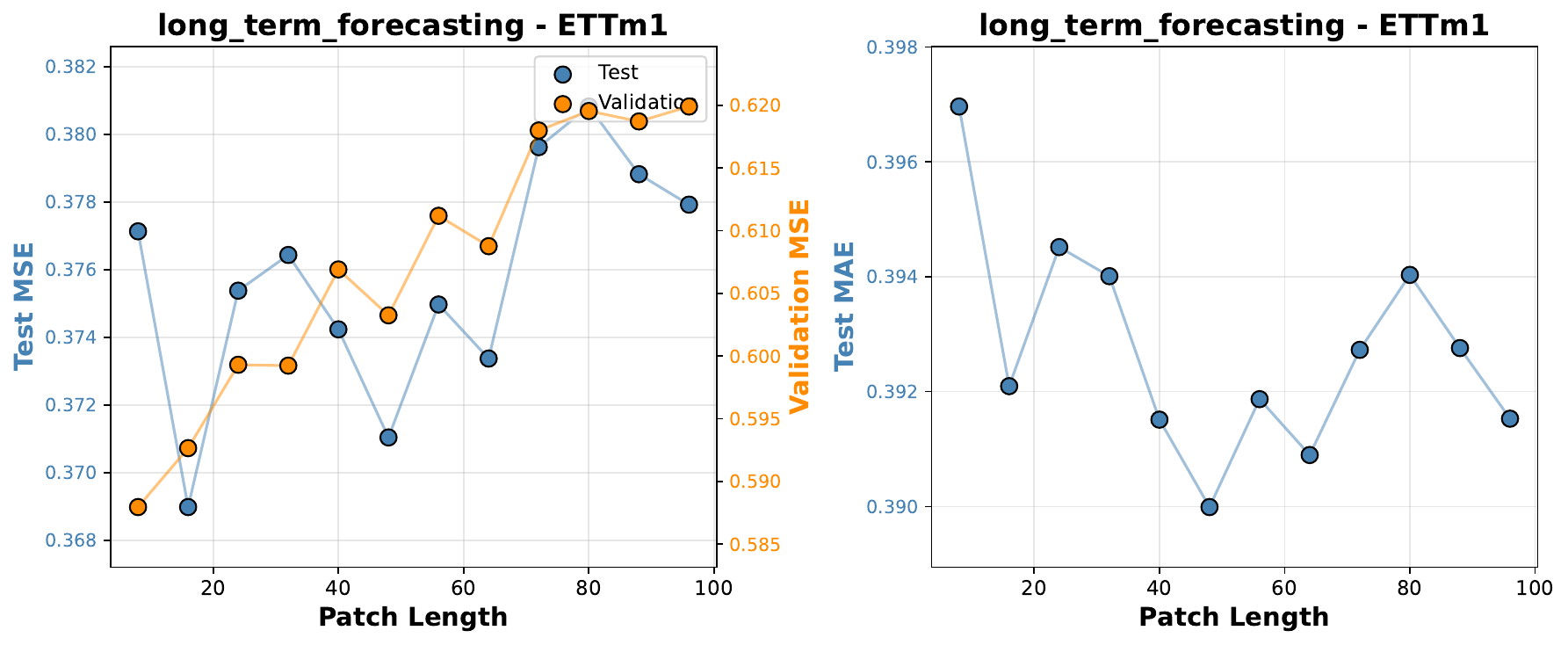}
    \caption{\textbf{Patch length sensitivity on ETTm1 (forecasting).} Average validation and test losses across horizons $\{96,192,336,720\}$ for different \texttt{patch\_len} values with \texttt{stride} $=\texttt{patch\_len}/2$.}
    \label{fig:patch_len_forecasting_ettm1}
\end{figure}

\begin{figure}
    \centering
    \includegraphics[width=1\linewidth]{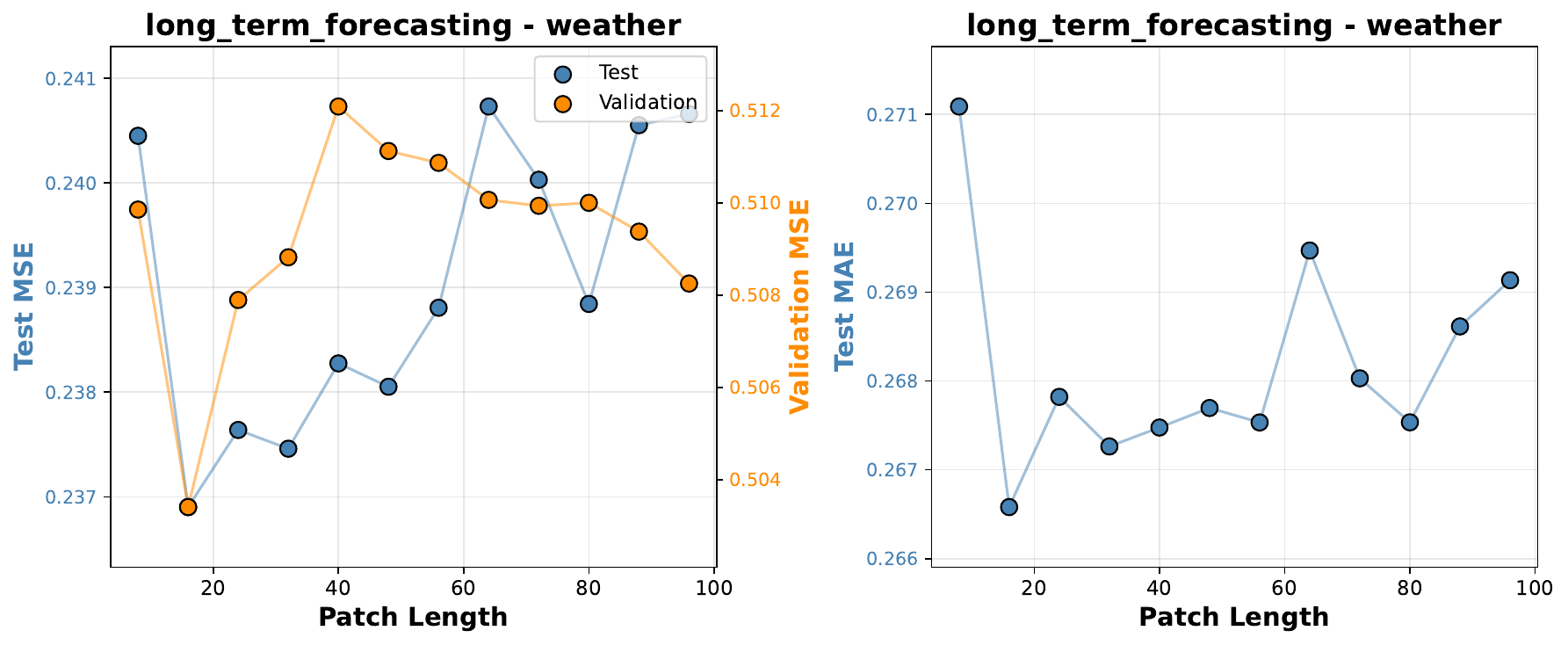}
    \caption{\textbf{Patch length sensitivity on Weather (forecasting).} Average validation and test losses across horizons $\{96,192,336,720\}$ for different \texttt{patch\_len} values with \texttt{stride} $=\texttt{patch\_len}/2$.}
    \label{fig:patch_len_forecasting_weather}
\end{figure}

In this subsection, we conducted a patching sensitivity experiment for forecasting by fixing $\texttt{stride}=\texttt{patch\_len}/2$ and testing different values of $\texttt{patch\_len}$ from  8 to 96 in increments of 8. The plots report the average validation and test losses on the ETTm1 (Figure~\ref{fig:patch_len_forecasting_ettm1}) and Weather (Figure~\ref{fig:patch_len_forecasting_weather}) datasets, where the average is computed across horizons $\{96,192,336,720\}$. Overall, performance varies only slightly across different setups, indicating that our method is robust to this hyperparameter. While the effect is small on \textbf{ETTm1}, it is more noticeable on \textbf{Weather}, yet the differences remain limited to roughly 2-4\%, suggesting that the preferred patching configuration can be dataset-dependent without requiring highly precise tuning.

\subsection{Signed-attention mask analysis.} \label{appendix:signed_attention_mask_analysis}

\paragraph{Distribution of negative weights.}

\begin{figure}[t]
    \centering
    \includegraphics[width=1\linewidth]{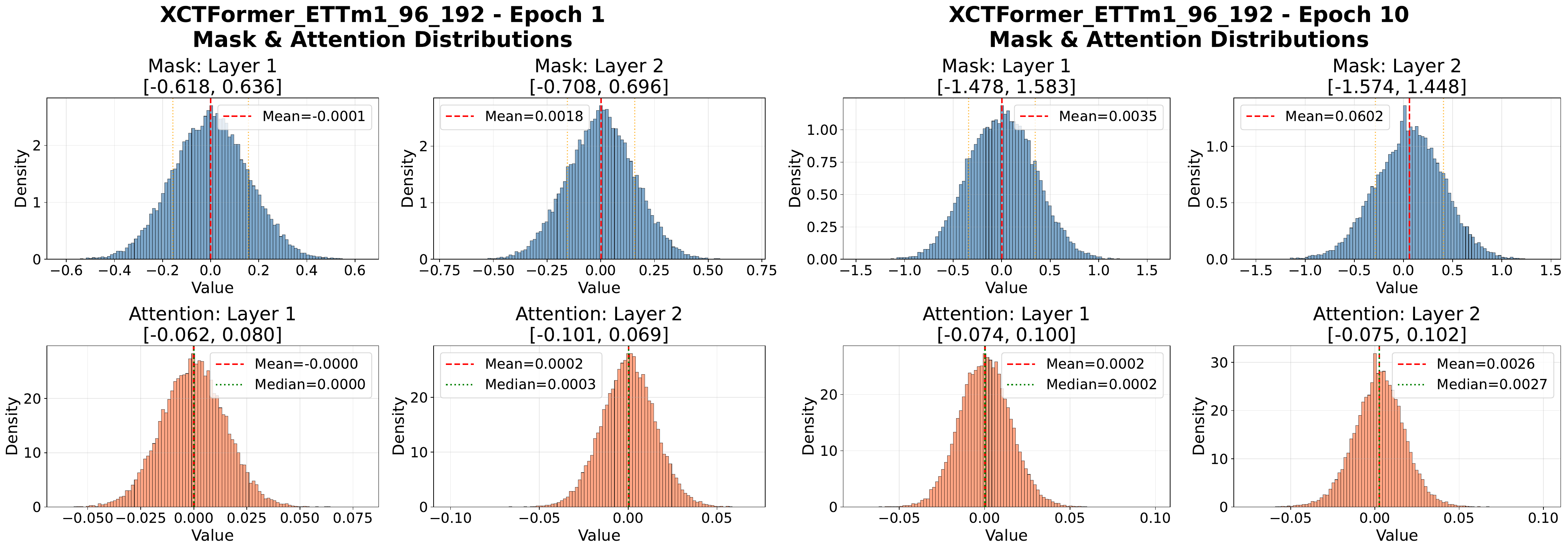}
    \caption{\textbf{Distributions of mask and signed-attention weights.} Histograms of the learnable mask values $M$ and the resulting activated attention weights. Results are shown for the forecasting task upon the ETTm1 dataset with lookback $L{=}96$ and horizon $H{=}192$. The left panel shows the distributions after the first training epoch, and the right panel after the final (10th) epoch.  The distributions remain approximately Gaussian over training, indicating that negative weights persist.}
    \label{fig:mask_attn_histogram}
\end{figure}

As recalled from the Method (Sec.~\ref{method:CRAB}), we do not use the vanilla attention scores directly. Instead, we first apply a positive transformation and then adjust the scores with a learnable, non-Boolean attention mask, which controls both the sign and the magnitude of the resulting attention weights. Concretely, given an attention score matrix $A\in\mathbb{R}^{N\times N}$, we remove sign information via a global shift $A_{+}=A-\min(A)$, which ensures $A_{+}\ge 0$ element-wise, and then form the signed attention as $A = M \circ A_{+}$, where $M\in\mathbb{R}^{N\times N}$ is a learnable real-valued mask applied element-wise. Since $A_{+}$ is non-negative, the sign of each entry in $A$ is determined entirely by the corresponding mask value: $M_{ij}<0$ produces a negative attention weight, while $M_{ij}>0$ produces a positive one. We initialize $M$ from a zero-mean Gaussian distribution, so roughly half of its entries are negative at initialization, and consequently, about half of the resulting attention weights are negative as well. Although $M$ is fine-tuned during training, its distribution remains close to normal, so negative weights persist and can contribute throughout optimization. Take a look at Figure \ref{fig:mask_attn_histogram}, which shows the distributions of the mask values and the activated attention weights at the end of the first and last training epochs.

\paragraph{Ablation: clipping negative weights.} 

We further isolate the role of negative weights with a targeted ablation. We keep our signed-attention activation unchanged, and only add a final ReLU that clips all negative values to zero. This variant is trained with the exact same hyperparameters and experimental settings, covering all forecasting horizons $\{96,192,336,720\}$ and all imputation mask ratios $\{0.125,0.25,0.375,0.5\}$. We then report performance averaged across these settings (see Table~\ref{tab:ablation_study_xctformer_negative}). Overall, the original model that permits negative weights improves performance by about 1.3\% over the clipped variant, suggesting that negative weights provide a consistent but modest performance gain.

\begin{table}[htp]
    \caption{Ablation study results across different tasks, evaluated with different XCTFormer with clipped values.}
    \label{tab:ablation_study_xctformer_negative}
    \centering
    \resizebox{\columnwidth}{!}{
    \begin{tabular}{lcccccccc}
        \toprule
            & \multicolumn{2}{c}{Long-term Forecasting} & \multicolumn{2}{c}{Imputation} & \multicolumn{3}{c}{Anomaly Detection} & \begin{tabular}[c]{@{}c@{}}XCTFormer\\vs Others\end{tabular} \\
            \cmidrule(lr){2-3} \cmidrule(lr){4-5} \cmidrule(lr){6-8} \cmidrule(lr){9-9}
            & MSE & MAE & MSE & MAE & Precision & Recall & F-Score & (\%) \\
        \midrule
           \rowcolor{tabhighlight}
            XCTFormer (Original) & \textbf{0.328} & \textbf{0.337} & \textbf{0.044} & \textbf{0.124} & \textbf{92.1} & \textbf{83.7} & \textbf{87.6} & - \\
            \hspace{2em} AbsAct + ReLU & \underline{0.331} & \underline{0.339} & \underline{0.044} & \underline{0.124} & \underline{92.1} & \underline{79.5} & \underline{85.2} & 1.3\% \\
        \bottomrule
    \end{tabular}
   }
\end{table}

\subsection{Empirical Scalability Analysis with Respect to Input Dimension}
\label{appendix:scalability_input_dim}

As noted in the Limitations, XCTFormer is sensitive to input size because it explicitly models all pairwise time and channel relationships, which induces quadratic scaling. We design DeCoP to mitigate this cost by compressing per-token dependencies and thereby achieving approximately linear scaling. To verify these theoretical expectations, we conduct an empirical scalability analysis. Because the effective input equals the product of the sequence tokens (after patching) and the number of channels, we vary each factor independently and measure how the parameter amount changes. Table \ref{tab:scalability_analysis_res_data_dim} represents the hyperparameter used.

\begin{table}[htbp]
\centering
\caption{Hyperparameter settings for \textbf{XCTFormer} scalability analysis}
\label{tab:scalability_analysis_res_data_dim}
\resizebox{\textwidth}{!}{%
\begin{tabular}{|l|c|c|c|c|c|c|c|c|c|c|c|}
\hline
\multicolumn{1}{|l|}{} &
\multicolumn{2}{c|}{\textbf{Data Processing}} &
\multicolumn{6}{c|}{\textbf{Transformer}} &
\multicolumn{2}{c|}{\textbf{XCTFormer}} &
\multicolumn{1}{c|}{\textbf{Variable}} \\
\hline
\textbf{Experiment} &
\textbf{patch\_len} &
\textbf{stride} &
\textbf{e\_layers} &
\textbf{n\_heads} &
\textbf{d\_model} &
\textbf{d\_ff} &
\textbf{dropout} &
\textbf{fc\_dropout} &
\textbf{attn\_dropout} &
\textbf{k} &
\textbf{Range} \\
\hline
\multicolumn{12}{|c|}{\textbf{Dimension Scaling Analysis}} \\
\hline
XCTFormer            & 16 & 8 & 2 & 4 & 128 & 256 & 0.1 & 0.05 & 0.0 & -  & n\_features: 1--100 (step 2) \\
XCTFormer (DeCoP)    & 16 & 8 & 2 & 4 & 128 & 256 & 0.1 & 0.05 & 0.0 & 64 & n\_features: 1--100 (step 2), 100--800 (step 16) \\
\hline
\multicolumn{12}{|c|}{\textbf{Sequence Length Scaling Analysis}} \\
\hline
XCTFormer            & 16 & 8 & 2 & 4 & 128 & 256 & 0.1 & 0.05 & 0.0 & -  & seq\_len: 64--1024 (step 32) \\
XCTFormer (DeCoP)    & 16 & 8 & 2 & 4 & 128 & 256 & 0.1 & 0.05 & 0.0 & 64 & seq\_len: 64--1024 (step 32), 1024--4096 (step 128) \\
\hline
\end{tabular}%
}
\vspace{0.2cm}

\end{table}

\paragraph{Channel scaling.}

\begin{figure}[t]
    \centering
    \includegraphics[width=\linewidth]{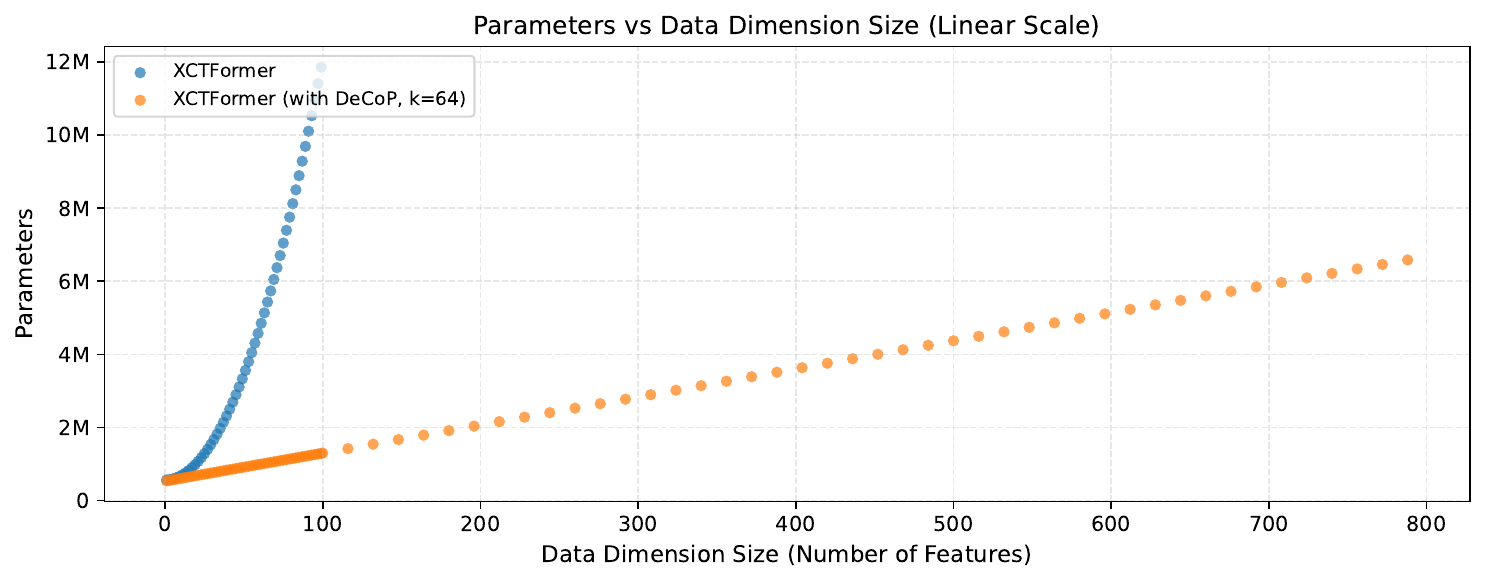}
    \caption{Scalability with respect to channel dimensionality ($n\_features$). XCTFormer's parameter count grows quadratically, while the DeCoP variant scales approximately linearly.}
    \label{fig:scalability_channels}
\end{figure}

We fix the lookback sequence length to 96 and vary the number of channels to match the dataset's dimensionality. Specifically, we tested $n\_features$ from 1 to 100 in steps of 2, and further extended from 100 to 800 in steps of 16 for the DeCoP variant. Because utilizing XCTFormer without DeCoP on larger datasets is impractical, the original XCTFormer is evaluated only over the smaller range. As shown in Fig.~\ref{fig:scalability_channels}, XCTFormer exhibits quadratic parameter growth with respect to the number of channels, whereas adding DeCoP yields approximately linear growth.

\begin{figure}[t]
    \centering
    \includegraphics[width=\linewidth]{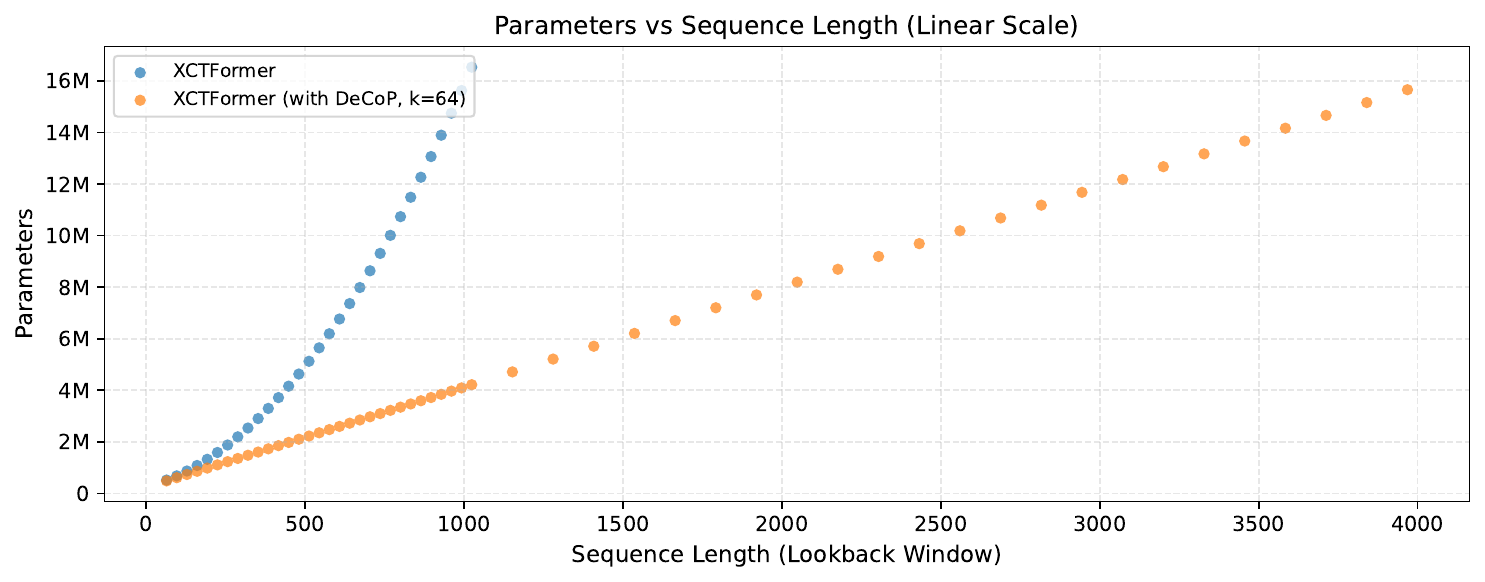}
    \caption{Scalability with respect to sequence length ($\texttt{seq\_len}$). XCTFormer exhibits quadratic parameter growth, while the DeCoP variant scales approximately linearly.}
    \label{fig:scalability_sequence}
\end{figure}

\paragraph{Sequence scaling.}
We fix the data dimensionality to 10 and vary the sequence length, which determines the number of tokens after patching. We evaluate sequences from 64 to 1024 in steps of 32, and extend from 1024 to 4096 in steps of 128 for the DeCoP variant. Again, XCTFormer is evaluated only on the shorter range. Fig.~\ref{fig:scalability_sequence} shows the same trend: XCTFormer grows quadratically with sequence length, whereas XCTFormer with DeCoP scales approximately linearly.

\section{Appendix: Synthetic Dataset}\label{appendix:synthetic_dataset}

\subsection{Overview \& Motivation}

While existing synthetic datasets are available~\citep{nochumsohn2024beyond, nochumsohn2025data, oreshkin2026zero} and generative approaches enable sampling from empirical data~\citep{naiman2024generative, naiman2024utilizing, gonen2025time, fadlon025diffusion}, they do not explicitly target the interplay between temporal and cross-variate dependencies. We propose a new synthetic dataset, evaluated on the forecasting task, to emphasize settings in which forecastability depends jointly on cross-variate and lagged cross-time relationships. The dataset consists of seven channels and is evaluated under the multivariate-to-single (MS) setting: all channels serve as input, but training and evaluation loss is computed only on the 7th channel (the synthetic target). The MS setting is adopted to specifically evaluate each model's ability to forecast data that depends solely on cross-channel and cross-time relationships. This target channel is constructed entirely from lagged signals in the first four channels, making it difficult to predict from its own history alone. Channels 5 and 6 (var\_5, var\_6) are independent random walks that act as distractors, meant to probe whether a model latches onto spurious correlations, i.e., learns to rely on channels that share no causal relationship with the target, which would degrade generalization.

\subsection{Dataset Construction}

The dataset comprises 7 channels and 10{,}000 time steps. At a high level, the channels fall into three groups:
\begin{itemize}
  \item \textbf{Informative sources} (channels 1--4): four sine waves with distinct amplitudes and frequencies that carry the information needed to predict the target.
  \item \textbf{Distractors} (channels 5--6): two random walks that share no causal relationship with the target. They test whether a model can ignore spurious correlations.
  \item \textbf{Target} (channel 7): constructed entirely from \emph{lagged patches} of the four source channels. At each point in time the target blends patches from different sources at different look-back distances, and the blend weight itself changes over time. Because the target is derived solely from other channels' past values, it is difficult to predict from its own history alone. Small Gaussian noise is added for realism.
\end{itemize}

The remainder of this section specifies each construction step in detail.

\subsubsection{Step 1: Source Signals}

Table~\ref{tab:synthetic_variates} lists all seven channels. The four informative sources are sine waves with varying amplitudes ($1.0$--$5.0$), frequencies ($0.002$--$0.03$), and phase offsets, ensuring the signals are diverse and not trivially correlated. The two distractor channels are random walks, which are non-stationary and can exhibit spurious short-range correlations with any signal. A model that naively attends to all variates equally would be misled by these distractors.

\begin{table}[htbp]
  \caption{Synthetic dataset variate specifications.}\label{tab:synthetic_variates}
  \vskip 0.05in
  \centering
  \begin{small}
  \begin{tabular}{c l l l}
    \toprule
    Index & Name & Type & Role \\
    \midrule
    1 & var\_1 & Sine wave ($A{=}1.0$, $f{=}0.02$) & Informative source \\
    2 & var\_2 & Sine wave ($A{=}3.0$, $f{=}0.03$) & Informative source \\
    3 & var\_3 & Sine wave ($A{=}2.0$, $f{=}0.01$) & Informative source \\
    4 & var\_4 & Sine wave ($A{=}5.0$, $f{=}0.002$) & Informative source \\
    5 & var\_5 & Random walk ($\sigma{=}0.1$) & Distractor \\
    6 & var\_6 & Random walk ($\sigma{=}0.15$) & Distractor \\
    7 & target & Patch-dependent blend & Target (predicted) \\
    \bottomrule
  \end{tabular}
  \end{small}
\end{table}

\subsubsection{Step 2: Patching}

The source signals are divided into non-overlapping patches of length 16 with a stride of 8 (50\% overlap). All subsequent operations are defined at the patch level.

\subsubsection{Step 3: Target Construction}

The target channel is built by blending patches from two pairs of source variates, each read at a specific lag. The two pairs are:
\begin{itemize}
  \item \textbf{Pair 1}: var\_1 (lag\,=\,1) and var\_2 (lag\,=\,2), contributing weight 0.5.
  \item \textbf{Pair 2}: var\_3 (lag\,=\,2) and var\_4 (lag\,=\,3), contributing weight 0.5.
\end{itemize}

The different lags ensure that the model cannot simply align patches at a single fixed offset; it must learn to look back different distances for different variates.

Within each pair, the relative contribution of the two sources is controlled by a blend weight $w(k)$ that varies over time. Specifically, $w(k)$ follows a triangle wave cycling between 0 and 1 with a period of 20 patches (160 time steps), creating a non-stationary dependency structure:
\begin{equation}
w(k) = \begin{cases}
  (k \bmod 20) / 10, & \text{if } (k \bmod 20) \le 10 \\
  2 - (k \bmod 20) / 10, & \text{otherwise}
\end{cases}
\end{equation}
That is, $w(k)$ cycles through the values $0, 0.1, 0.2, \ldots, 0.9, 1.0, 0.9, \ldots, 0.1, 0, 0.1, \ldots$ as $k$ increases.

Putting it all together, the target patch at index $k$ is:
\begin{equation}
\text{target}[k] = 0.5 \cdot \bigl[w(k) \cdot \text{var\_1}[k{-}1] + (1{-}w(k)) \cdot \text{var\_2}[k{-}2]\bigr]
  + 0.5 \cdot \bigl[w(k) \cdot \text{var\_3}[k{-}2] + (1{-}w(k)) \cdot \text{var\_4}[k{-}3]\bigr]
\end{equation}
where $\text{var\_}i[k{-}\ell]$ denotes the 16-element patch extracted from variate $i$ at patch position $k{-}\ell$. Overlapping patch regions are averaged.

\subsubsection{Step 4: Noise}

Finally, Gaussian noise ($\sigma{=}0.02$) is added to the target channel to prevent trivial exact recovery and simulate real-world measurement noise.

\subsection{Dataset Splits \& Preprocessing}

Table~\ref{tab:synthetic_splits} summarizes the dataset splits and evaluation configuration.

\begin{table}[htbp]
  \caption{Synthetic dataset splits and evaluation configuration.}\label{tab:synthetic_splits}
  \vskip 0.05in
  \centering
  \begin{small}
  \begin{tabular}{l c c}
    \toprule
    Property & Parameter & Value \\
    \midrule
    Train / Val / Test split & Proportion & 70\% / 10\% / 20\% \\
    Total points & $n$ & 10{,}000 \\
    Normalization & StandardScaler & Fitted on train only \\
    Lookback window & \texttt{seq\_len} & 96 \\
    Decoder label length & \texttt{label\_len} & 48 \\
    Prediction horizon & \texttt{pred\_len} & \{96, 192, 336, 720\} \\
    Training epochs & \texttt{train\_epochs} & 10 \\
    \bottomrule
  \end{tabular}
  \end{small}
\end{table}

\subsection{Results}

Table~\ref{tab:synthetic_results} reports per-horizon forecasting results on the synthetic dataset, together with the trainable parameter count and FLOPs for each model on this dataset.

\begin{table}[htbp]
  \caption{Forecasting results on the synthetic dataset (MS setting) across prediction horizons, with computational cost on this dataset. All models use ETTm1 hyperparameters without dataset-specific tuning.}\label{tab:synthetic_results}
  \vskip 0.05in
  \centering
  \resizebox{1.0\columnwidth}{!}{
  \begin{small}
  \setlength{\tabcolsep}{4pt}
  \begin{tabular}{l|cc|cc|cc|cc|cc|r|r}
    \toprule
    \multirow{2}{*}{Model} &
    \multicolumn{2}{c|}{96} &
    \multicolumn{2}{c|}{192} &
    \multicolumn{2}{c|}{336} &
    \multicolumn{2}{c|}{720} &
    \multicolumn{2}{c|}{Avg} &
    \multirow{2}{*}{Params} &
    \multirow{2}{*}{FLOPs} \\
    & MSE & MAE & MSE & MAE & MSE & MAE & MSE & MAE & MSE & MAE & & \\
    \midrule
    \textbf{Crossformer}   & \boldres{0.018} & \boldres{0.106} & \boldres{0.026} & \boldres{0.126} & \boldres{0.027} & \boldres{0.129} & \boldres{0.031} & \boldres{0.137} & \boldres{0.025} & \boldres{0.124} & 1.1E+07 & 1.2E+09 \\
    \textbf{XCTFormer} (Ours) & 0.038 & 0.151 & \secondres{0.043} & \secondres{0.161} & \secondres{0.045} & \secondres{0.165} & \secondres{0.035} & \secondres{0.144} & \secondres{0.040} & \secondres{0.155} & 2.0E+05 & 3.3E+06 \\
    TimeMixer       & \secondres{0.025} & \secondres{0.123} & 0.070 & 0.204 & 0.053 & 0.178 & 0.133 & 0.281 & 0.070 & 0.197 & 1.2E+05 & 1.6E+07 \\
    Leddam          & 0.043 & 0.159 & 0.094 & 0.235 & 0.095 & 0.234 & 0.082 & 0.216 & 0.079 & 0.211 & 2.9E+06 & 3.5E+07 \\
    SCINet          & 0.130 & 0.277 & 0.092 & 0.235 & 0.085 & 0.225 & 0.079 & 0.218 & 0.097 & 0.239 & 2.8E+05 & 3.4E+06 \\
    iTransformer    & 0.217 & 0.362 & 0.236 & 0.375 & 0.282 & 0.411 & 0.282 & 0.413 & 0.254 & 0.390 & 2.6E+05 & 2.9E+06 \\
    PatchTST        & 0.222 & 0.365 & 0.319 & 0.442 & 0.319 & 0.441 & 0.249 & 0.384 & 0.277 & 0.408 & 9.2E+05 & 3.8E+07 \\
    TimesNet        & 0.554 & 0.597 & 0.471 & 0.538 & 0.404 & 0.499 & 0.350 & 0.459 & 0.445 & 0.523 & 2.7E+06 & 3.8E+09 \\
    DLinear         & 0.516 & 0.577 & 0.496 & 0.572 & 0.482 & 0.565 & 0.428 & 0.523 & 0.480 & 0.559 & 6.5E+04 & 4.5E+05 \\
    MTLinear        & 0.670 & 0.640 & 1.189 & 0.881 & 1.388 & 0.975 & 1.149 & 0.845 & 1.099 & 0.835 & 2.0E+05 & 3.4E+05 \\
    TiDE            & 0.836 & 0.746 & 1.290 & 0.934 & 1.476 & 1.016 & 1.182 & 0.875 & 1.196 & 0.893 & 9.6E+06 & 9.2E+07 \\
    Autoformer      & 0.573 & 0.605 & 1.753 & 1.094 & 1.743 & 1.079 & 1.606 & 1.019 & 1.419 & 0.949 & 1.2E+05 & 2.4E+07 \\
    FEDformer       & 1.537 & 1.006 & 2.422 & 1.282 & 4.065 & 1.663 & 1.879 & 1.110 & 2.476 & 1.265 & 1.6E+07 & 1.8E+09 \\
    \bottomrule
  \end{tabular}
  \end{small}
  }
\end{table}

The two models that explicitly capture both cross-variate and cross-time patch interactions, namely Crossformer and XCTFormer, achieve the lowest errors, consistent with the dataset's construction in which the target depends on lagged patches from other channels. Crossformer ranks first (Avg MSE\,=\,0.025) and XCTFormer ranks second (Avg MSE\,=\,0.040). However, Crossformer uses approximately $55{\times}$ more trainable parameters (11M vs.\ 200K) and $364{\times}$ more FLOPs (1.2E+09 vs.\ 3.3E+06) than XCTFormer on this dataset, highlighting a substantial efficiency gap.

We note that no dataset-specific hyperparameter tuning was performed: all models were trained with ETTm1 hyperparameters. Consequently, the reported results may not reflect each model's optimal performance on this dataset.

Channel-independent models (e.g., PatchTST, DLinear) tend to show higher error, which is expected given that the target is difficult to predict from its own history alone. The distractor channels (random walks) do not appear to noticeably affect the performance of cross-variate models, suggesting these architectures can, to some extent, distinguish informative from non-informative channels.

\begin{figure}[htbp]
  \centering
  \includegraphics[width=\linewidth]{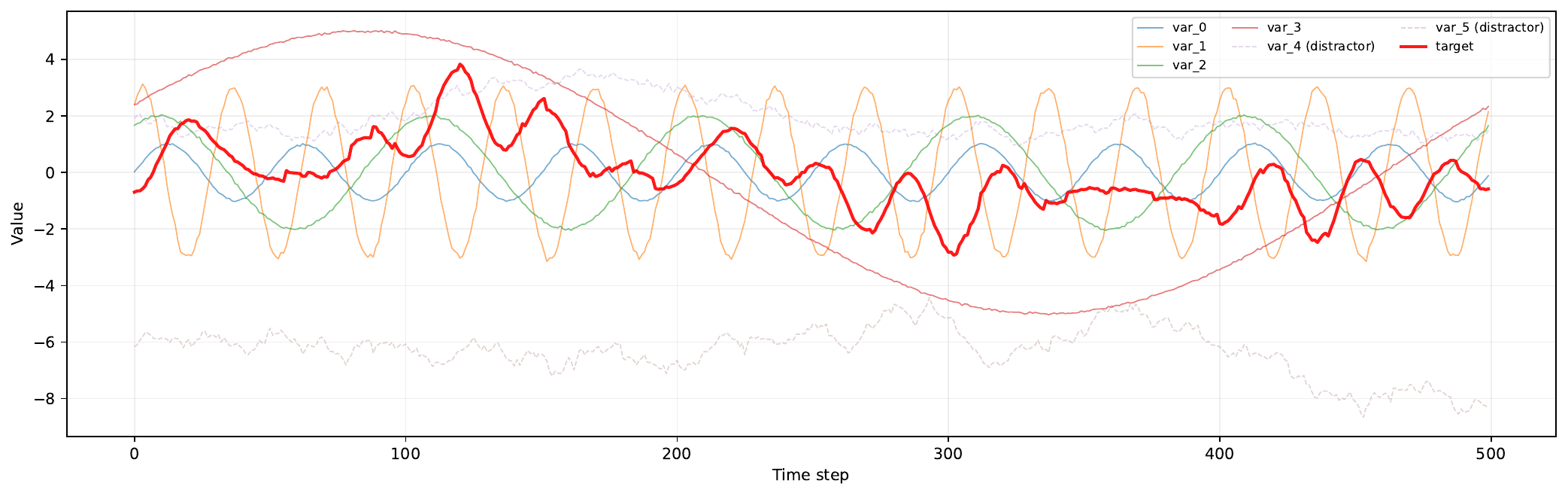}
  \caption{Synthetic dataset visualization: source signals (var\_1 through var\_6) and the constructed target channel.}
  \label{fig:synthetic_data_viz}
\end{figure}

\begin{figure}[htbp]
  \centering
  \includegraphics[width=\linewidth]{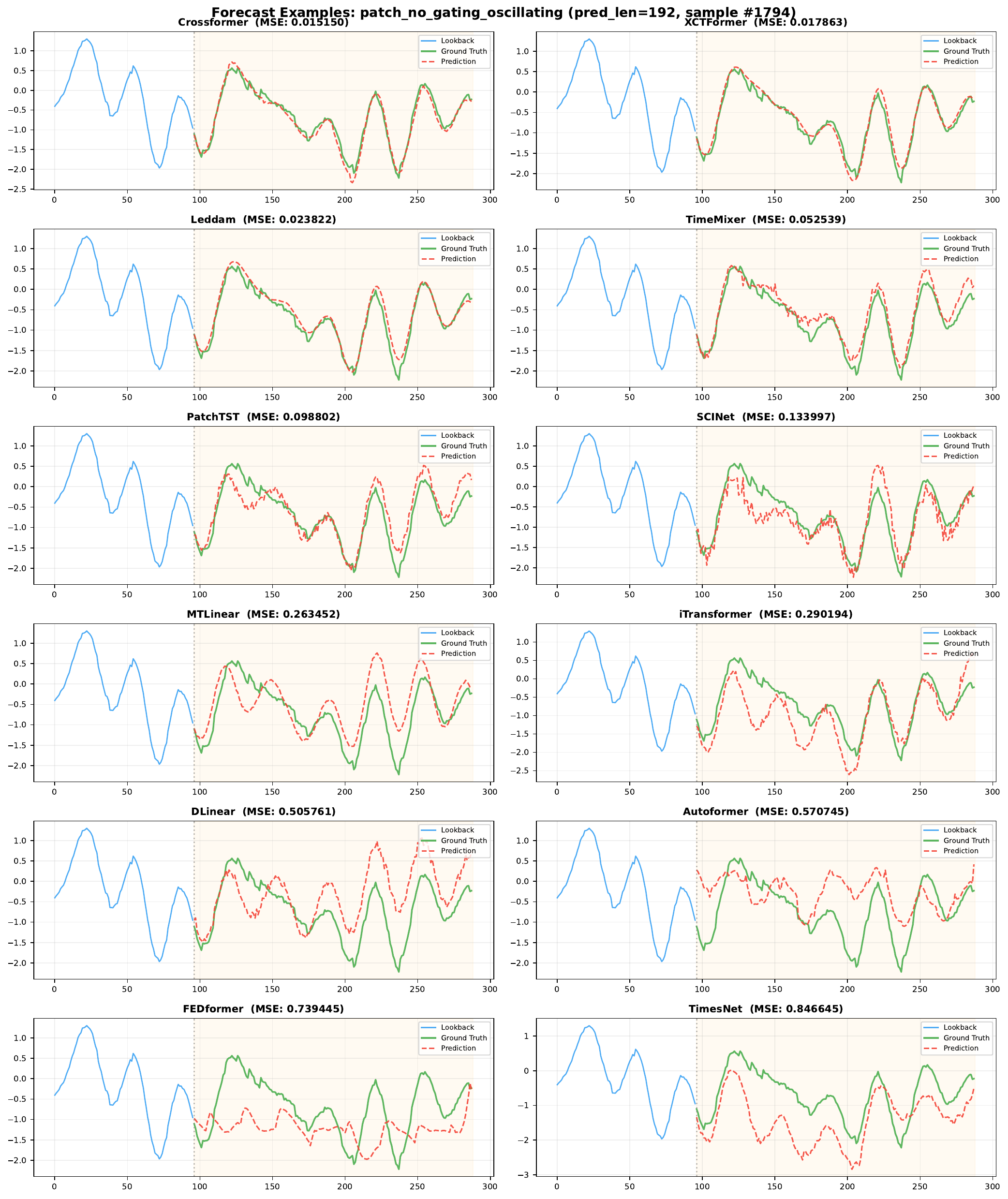}
  \caption{Prediction examples on the synthetic dataset: ground-truth target vs.\ model forecasts for selected models.}
  \label{fig:synthetic_forecast_examples}
\end{figure}

A visualization of the synthetic dataset is shown in Figure~\ref{fig:synthetic_data_viz}, and qualitative forecasting examples on this dataset are provided in Figure~\ref{fig:synthetic_forecast_examples}.

\newpage
\section{Appendix: Full Results}\label{appendix:full_search}

\subsection{Statistical significance tests}
\label{app:significance_tests}

To assess whether the observed performance differences are robust to random initialization, we conducted statistical significance tests for the \textbf{forecasting} task against two competitive baselines, \emph{LeDDAM} \citep{leddam} and \emph{iTransformer} \cite{liu2023itransformer}, under the same evaluation protocol used throughout the paper. The results are reported in Tables \ref{tab:ttest_comparison_leddam} and \ref{tab:ttest_comparison_itransformer} respectively.

\paragraph{Experimental setup.}
For each dataset, we evaluated all models on the four standard prediction horizons $\{96, 192, 336, 720\}$.
For every method (ours and each baseline), we trained the model \textbf{five} times using different random seeds $\{2021, 2022, 2023, 2024, 2025\}$.
For our model, we used the hyperparameters reported in Appendix~\ref{appendix:hyperparameters}.
For the baselines, we used the best hyperparameters provided in their official GitHub repositories.
All runs followed the same training procedure as the main experiments, including model selection based on validation loss and reporting test-set errors. We applied a two-sided paired t-test with $\alpha=0.05$. We mark a result as significant when the mean difference favors our model and $p \le 0.05$; otherwise, we treat it as inconclusive.

\begin{table}[H]
\caption{Statistical comparison of XCTFormer vs LeDDAM on forecasting datasets. Results averaged over prediction lengths $\{96, 192, 336, 720\}$ across five seeds (2021-2025). Confidence level derived from Welch's t-test (99.9\%: $p<0.001$, 99\%: $p<0.01$, 95\%: $p<0.05$). \textbf{Bold} indicates statistically significant better performance.}
  \label{tab:ttest_comparison_leddam}
  \vskip 0.05in
  \centering
  \begin{threeparttable}
  \begin{small}
  \renewcommand{\multirowsetup}{\centering}
  \setlength{\tabcolsep}{4pt}
  \begin{tabular}{l|cc|cc|c}
    \toprule
    & \multicolumn{2}{c|}{XCTFormer (Ours)} & \multicolumn{2}{c|}{LeDDAM} & Confidence \\
    \cmidrule(lr){2-3}\cmidrule(lr){4-5}\cmidrule(lr){6-6}
    Dataset & MSE & MAE & MSE & MAE & Level\\
    \midrule
    ETTh1 & $0.449 \pm 0.002$ & $0.436 \pm 0.001$ & $\mathbf{0.436} \pm 0.0073$ & $\mathbf{0.432} \pm 0.0034$ & 95\% \\
    ETTh2 & $0.374 \pm 0.007$ & $0.399 \pm 0.004$ & $0.374 \pm 0.0019$ & $0.398 \pm 0.0007$ & n.s. \\
    ETTm1 & $\mathbf{0.371} \pm 0.003$ & $\mathbf{0.393} \pm 0.002$ & $0.388 \pm 0.0034$ & $0.398 \pm 0.0023$ & 99\% \\
    ETTm2 & $\mathbf{0.271} \pm 0.001$ & $\mathbf{0.319} \pm 0.001$ & $0.282 \pm 0.0019$ & $0.326 \pm 0.0009$ & 99.9\% \\
    ECL$^p$ & $0.176 \pm 0.007$ & $0.270 \pm 0.007$ & $0.171 \pm 0.0042$ & $0.264 \pm 0.0027$ & n.s. \\
    Traffic$^p$ & $\mathbf{0.435} \pm 0.001$ & $\mathbf{0.287} \pm 0.001$ & $0.468 \pm 0.0075$ & $0.294 \pm 0.0052$ & 95\% \\
    Weather & $\mathbf{0.237} \pm 0.001$ & $\mathbf{0.267} \pm 0.001$ & $0.244 \pm 0.0013$ & $0.273 \pm 0.0012$ & 99.9\% \\
    \bottomrule
  \end{tabular}
  \end{small}
  \begin{tablenotes}
    \small
    \item[n.s.] Not statistically significant ($p \geq 0.10$).
    \item[$^p$] DeCoP was enabled for XCTFormer on this dataset.
  \end{tablenotes}
  \end{threeparttable}
\end{table}

\begin{table}[H]
\caption{Statistical comparison of XCTFormer vs iTransformer on forecasting datasets. Results averaged over prediction lengths $\{96, 192, 336, 720\}$ across five seeds (2021-2025). Confidence level derived from Welch's t-test (99.9\%: $p<0.001$, 99\%: $p<0.01$, 95\%: $p<0.05$). \textbf{Bold} indicates statistically significant better performance.}
  \label{tab:ttest_comparison_itransformer}
  \vskip 0.05in
  \centering
  \begin{threeparttable}
  \begin{small}
  \renewcommand{\multirowsetup}{\centering}
  \setlength{\tabcolsep}{4pt}
  \begin{tabular}{l|cc|cc|c}
    \toprule
    & \multicolumn{2}{c|}{XCTFormer (Ours)} & \multicolumn{2}{c|}{iTransformer} & Confidence \\
    \cmidrule(lr){2-3}\cmidrule(lr){4-5}\cmidrule(lr){6-6}
    Dataset & MSE & MAE & MSE & MAE & Level\\
    \midrule
    ETTh1 & $\mathbf{0.449} \pm 0.002$ & $\mathbf{0.436} \pm 0.001$ & $0.457 \pm 0.0014$ & $0.449 \pm 0.0013$ & 99.9\% \\
    ETTh2 & $\mathbf{0.374} \pm 0.007$ & $\mathbf{0.399} \pm 0.004$ & $0.383 \pm 0.0022$ & $0.407 \pm 0.0011$ & 95\% \\
    ETTm1 & $\mathbf{0.371} \pm 0.003$ & $\mathbf{0.393} \pm 0.002$ & $0.408 \pm 0.0022$ & $0.412 \pm 0.0013$ & 99.9\% \\
    ETTm2 & $\mathbf{0.271} \pm 0.001$ & $\mathbf{0.319} \pm 0.001$ & $0.291 \pm 0.0010$ & $0.335 \pm 0.0011$ & 99.9\% \\
    ECL$^p$ & $0.176 \pm 0.007$ & $0.270 \pm 0.007$ & $0.176 \pm 0.0037$ & $0.267 \pm 0.0026$ & n.s. \\
    Traffic$^p$ & $0.435 \pm 0.001$ & $0.287 \pm 0.001$ & $\mathbf{0.430} \pm 0.0010$ & $\mathbf{0.283} \pm 0.0010$ & 99.9\% \\
    Weather & $\mathbf{0.237} \pm 0.001$ & $\mathbf{0.267} \pm 0.001$ & $0.260 \pm 0.0014$ & $0.281 \pm 0.0016$ & 99.9\% \\
    \bottomrule
  \end{tabular}
  \end{small}
  \begin{tablenotes}
    \small
    \item[n.s.] Not statistically significant ($p \geq 0.10$).
    \item[$^p$] DeCoP was enabled for XCTFormer on this dataset.
  \end{tablenotes}
  \end{threeparttable}
\end{table}

\paragraph{Results.}
Overall, our model achieves strong performance and shows statistically significant improvements across most datasets relative to both baselines.
Compared to \emph{LeDDAM}, the results are inconclusive on ECL and ETTh2, while ETTh1 favors LeDDAM.
Compared to \emph{iTransformer}, the result is inconclusive on ECL, and Traffic favors iTransformer.
These outcomes suggest that gains are often consistent across seeds, but in some datasets, the differences are not statistically significant.

\subsection{Ablation Study: Complete Analysis} \label{appendix:ablation_details}

We conduct a systematic ablation study with six configurations to isolate the contribution of each architectural component in \textbf{XCTFormer}. All variants maintain identical data processing (patch length/stride), training procedures, and model parameters, except for the specific component being modified.

\paragraph{Configuration Details}

\begin{enumerate}
    \item \textbf{Full XCTFormer (Baseline)}: Complete architecture including CRAB module  with learnable non-boolean mask (for datasets with $\leq$60 channels) or DeCoP (for datasets with $>$60 channels), cross-time and cross-channel attention, and our proposed attention activation function.

    \item \textbf{W/o Learnable Mask}: Removes the learnable mask component. Attention masks are not converted to positive values and no element-wise multiplication is applied. The CRAB module remains unchanged otherwise.

    \item \textbf{Standard Softmax Activation}: Replaces our proposed activation function with standard Transformer softmax while preserving CRAB and the learnable mask. Note that our \texttt{attention\_dropout} rate parameter is replaced with the standard \texttt{dropout} argument commonly used in related work for fair comparison.

    \item \textbf{Vanilla Transformer}: Substitutes CRAB (and DeCoP) with standard attention blocks following \citet{vaswani2017attention}.

    \item \textbf{Sequence Modeling Only}: Retains only temporal self-attention within each channel, disabling cross-channel modeling (channel-independent processing). This configuration tests the necessity of modeling cross-channel relationships, mirroring approaches like PatchTST \citep{nie2023time}.

    \item \textbf{Channel Modeling Only}: Preserves only cross-channel attention at each time step while removing temporal self-attention. This configuration tests the necessity of modeling temporal relationships, similar to designs that emphasize cross-variable mixing like iTransformer \citep{liu2023itransformer}.
\end{enumerate}

\paragraph{Full Ablation Study Results}

Complete ablation study results for each time-series task are presented in Tables \ref{tab:ablation_study_long_term_forecasting_xctformer}, \ref{tab:ablation_study_imputation_xctformer}, and \ref{tab:ablation_study_anomaly_detection_xctformer}. For long-term forecasting and imputation tasks, the results shown for each dataset represent averages across all prediction horizons and mask ratios, respectively.

\begin{table}[htp]
    \caption{Ablation study results for Long-term Forecasting across different datasets, evaluated with different XCTFormer variations.}
    \label{tab:ablation_study_long_term_forecasting_xctformer}
    \centering
    \resizebox{\columnwidth}{!}{
    \begin{tabular}{lccccccccccccccc}
        \toprule
            & \multicolumn{2}{c}{ETTh1} & \multicolumn{2}{c}{ETTh2} & \multicolumn{2}{c}{ETTm1} & \multicolumn{2}{c}{ETTm2} & \multicolumn{2}{c}{ECL$^p$} & \multicolumn{2}{c}{Traffic$^p$} & \multicolumn{2}{c}{Weather} & \begin{tabular}[c]{@{}c@{}}XCTFormer\\vs Others\end{tabular} \\
            \cmidrule(lr){2-3} \cmidrule(lr){4-5} \cmidrule(lr){6-7} \cmidrule(lr){8-9} \cmidrule(lr){10-11} \cmidrule(lr){12-13} \cmidrule(lr){14-15}
            & MSE & MAE & MSE & MAE & MSE & MAE & MSE & MAE & MSE & MAE & MSE & MAE & MSE & MAE & (\%) \\
        \midrule
           \rowcolor{tabhighlight}
            XCTFormer (Original) & 0.450 & \underline{0.436} & \textbf{0.369} & \textbf{0.396} & \textbf{0.369} & \textbf{0.392} & \textbf{0.270} & \textbf{0.319} & \textbf{0.166} & \textbf{0.263} & \textbf{0.435} & \underline{0.287} & \textbf{0.237} & \textbf{0.267} & - \\
            \hspace{2em} W/o mask$^\dagger$ & 0.453 & 0.440 & 0.383 & 0.405 & 0.379 & 0.394 & 0.282 & 0.326 &  &  &  &  & 0.256 & 0.280 &  +2.7\% \\
            \hspace{2em} Original softmax activation & \textbf{0.443} & \textbf{0.435} & 0.390 & 0.409 & 0.410 & 0.413 & 0.280 & 0.326 & 0.224 & 0.317 & 0.520 & 0.368 & 0.247 & 0.278 & +8.0\% \\
            \hspace{2em} Vanilla transformer & 0.452 & 0.439 & 0.397 & 0.412 & 0.385 & 0.398 & 0.285 & 0.330 & 0.224 & 0.317 & 0.520 & 0.368 & 0.263 & 0.284 & +8.3\% \\
            \hspace{2em} Sequence modeling & \underline{0.449} & 0.436 & \underline{0.380} & \underline{0.404} & \underline{0.375} & \underline{0.393} & 0.281 & 0.327 & 0.195 & 0.280 & 0.450 & \textbf{0.284} & 0.256 & 0.279 & +2.8\% \\
            \hspace{2em} Channel modeling & 0.461 & 0.450 & 0.388 & 0.409 & 0.378 & 0.393 & \underline{0.278} & \underline{0.321} & \textbf{0.166} & \textbf{0.263} & 0.476 & 0.329 & \underline{0.239} & \underline{0.269} & +3.4\% \\
        \bottomrule
    \end{tabular}
   }
   \vspace{0.3em}
   \footnotesize{$^p$ DeCoP was enabled for XCTFormer on this dataset.} \footnotesize{$^\dagger$ W/o mask results exclude DeCoP datasets where the learnable mask is not applicable; \% is recomputed over remaining datasets.}
\end{table}

\begin{table}[htp]
    \caption{Ablation study results for Imputation across different datasets, evaluated with different XCTFormer variations.}
    \label{tab:ablation_study_imputation_xctformer}
    \centering
    \resizebox{\columnwidth}{!}{
    \begin{tabular}{lccccccccccccc}
        \toprule
            & \multicolumn{2}{c}{ETTm1} & \multicolumn{2}{c}{ETTm2} & \multicolumn{2}{c}{ETTh1} & \multicolumn{2}{c}{ETTh2} & \multicolumn{2}{c}{Weather} & \multicolumn{2}{c}{ECL$^p$} & \begin{tabular}[c]{@{}c@{}}XCTFormer\\vs Others\end{tabular} \\
            \cmidrule(lr){2-3} \cmidrule(lr){4-5} \cmidrule(lr){6-7} \cmidrule(lr){8-9} \cmidrule(lr){10-11} \cmidrule(lr){12-13}
            & MSE & MAE & MSE & MAE & MSE & MAE & MSE & MAE & MSE & MAE & MSE & MAE & (\%) \\
        \midrule
        \rowcolor{tabhighlight}
        XCTFormer (Original) & \textbf{0.029} & \textbf{0.113} & \textbf{0.024} & \textbf{0.092} & \underline{0.087} & 0.201 & \textbf{0.046} & \textbf{0.144} & \textbf{0.031} & 0.050 & \textbf{0.046} & \underline{0.141} & - \\
        \hspace{2em} W/o mask$^\dagger$ & 0.041 & 0.132 & 0.029 & 0.100 & 0.092 & 0.206 & 0.064 & 0.169 & \underline{0.031} & \underline{0.047} &  &  &  +13.9\% \\
        \hspace{2em} Original softmax activation & \underline{0.032} & \underline{0.117} & \underline{0.027} & \underline{0.099} & \textbf{0.078} & \textbf{0.192} & 0.065 & 0.178 & 0.040 & 0.075 & 0.077 & 0.196 & +14.7\% \\
        \hspace{2em} Vanilla transformer & 0.043 & 0.137 & 0.033 & 0.113 & 0.094 & 0.208 & 0.079 & 0.193 & 0.031 & 0.048 & 0.077 & 0.196 & +19.9\% \\
        \hspace{2em} Sequence modeling & 0.040 & 0.130 & 0.029 & 0.101 & 0.089 & \underline{0.201} & 0.081 & \underline{0.166} & \textbf{0.031} & \textbf{0.047} & \textbf{0.046} & \textbf{0.139} & +8.7\% \\
        \hspace{2em} Channel modeling & 0.065 & 0.174 & 0.044 & 0.136 & 0.203 & 0.305 & \underline{0.060} & 0.167 & 0.041 & 0.072 & \underline{0.075} & 0.191 & +34.6\% \\
        \bottomrule
    \end{tabular}
   }
   \vspace{0.3em}
   \footnotesize{$^p$ DeCoP was enabled for XCTFormer on this dataset.} \footnotesize{$^\dagger$ W/o mask results exclude DeCoP datasets where the learnable mask is not applicable; \% is recomputed over remaining datasets.}
\end{table}

\begin{table}[htp]
    \caption{Ablation study results for Anomaly Detection across different datasets, evaluated with different XCTFormer variations.}
    \label{tab:ablation_study_anomaly_detection_xctformer}
    \centering
    \resizebox{\columnwidth}{!}{
    \begin{tabular}{lcccccc}
        \toprule
            & PSM & SWaT & MSL & SMAP & SMD & \begin{tabular}[c]{@{}c@{}}XCTFormer\\vs Others\end{tabular} \\
            & F-Score & F-Score & F-Score & F-Score & F-Score & (\%) \\
        \midrule
           \rowcolor{tabhighlight}
            XCTFormer (Original) & \textbf{95.3} & 92.6 & \underline{79.0} & \textbf{86.7} & \underline{84.2} & - \\
            \hspace{2em} W/o mask & 95.3 & 88.2 & 72.3 & 66.7 & 83.0 & +8.0\% \\
            \hspace{2em} Original softmax activation & 95.3 & 90.2 & 69.0 & \underline{68.6} & \textbf{84.3} & +7.5\% \\
            \hspace{2em} Vanilla transformer & \underline{95.3} & \underline{93.1} & 71.8 & 66.9 & 83.2 & +6.7\% \\
            \hspace{2em} Sequence modeling & 95.3 & \textbf{93.5} & \textbf{79.4} & 67.5 & 83.9 & +4.4\% \\
            \hspace{2em} Channel modeling & 92.9 & 92.5 & 76.9 & 66.5 & 82.8 & +6.4\% \\
        \bottomrule
    \end{tabular}
   }
\end{table}

\subsection{Robustness Across Random Seeds: Complete Analysis} \label{appendix:full_seed_analysis}

\paragraph{Coefficient of Variation (CV) for a Single Metric}
We quantify run-to-run stability using the coefficient of variation, a unitless measure of dispersion relative to the mean \citep{Reed2002CV}. For a metric with mean $\mu$ and standard deviation $\sigma$ across five seeds (2021 to 2025), we compute:
\[
\mathrm{CV}(\%) = 100 \cdot \frac{\sigma}{|\mu|}.
\]
The coefficient of variation tells us how much results vary around their mean \emph{relative} to the mean itself. Since CV is unitless, it enables comparison across datasets and metrics: smaller values indicate greater stability, while larger values indicate greater variability.

\noindent\textbf{Confidence score mapping}
For intuitive interpretation, we report a complementary confidence score:
\[
\mathrm{Conf}(\%) = 100 - \mathrm{CV}(\%).
\]
This confidence score inverts the scale so that lower variability corresponds to higher confidence. For example, if $\mathrm{CV}=3.2\%$, then $\mathrm{Conf}=96.8\%$, indicating that repeated runs with identical setups produce very similar results.

\paragraph{Full Seed Analysis Results}
To enhance readability, we include only the averaged analysis table for all time-series tasks in the main paper, while the complete results are provided in Tables \ref{tab:forecasting_results_seed}, \ref{tab:imputation_results_seed} and \ref{tab:anomaly_results_seed}. The confidence score presented for each dataset represents the average confidence score across all of its metrics.

\begin{table}[H]
\caption{Standard deviation for XCTFormer on forecasting datasets, evaluated across five seeds (2021-2025). Results averaged over the four prediction lengths $\{96, 192, 336, 720\}$.}
  \label{tab:forecasting_results_seed}
  \vskip 0.05in
  \centering
  \begin{threeparttable}
  \begin{small}
  \renewcommand{\multirowsetup}{\centering}
  \setlength{\tabcolsep}{5pt}
  \begin{tabular}{l|cc|c}
    \toprule
    Model & \multicolumn{2}{c|}{XCTFormer (Ours)} & Confidence Score  \\
    \cmidrule(lr){0-1}\cmidrule(lr){2-3}\cmidrule(lr){4-4}
    Dataset & MSE & MAE & Score \%\\
    \midrule
    ETTh1 & $0.449 \pm 0.002$ & $0.436 \pm 0.001$ & 99.7\% \\
    ETTh2 & $0.374 \pm 0.007$ & $0.399 \pm 0.004$ & 98.5\% \\
    ETTm1 & $0.371 \pm 0.003$ & $0.393 \pm 0.002$ & 99.3\% \\
    ETTm2 & $0.271 \pm 0.001$ & $0.319 \pm 0.001$ & 99.6\% \\
    ECL$^p$ & $0.176 \pm 0.007$ & $0.270 \pm 0.007$ & 96.6\% \\
    Traffic$^p$ & $0.435 \pm 0.001$ & $0.287 \pm 0.001$ & 99.7\% \\
    Weather & $0.237 \pm 0.001$ & $0.267 \pm 9.81e-04$ & 99.5\% \\
    \bottomrule
  \end{tabular}
    \end{small}
    \begin{tablenotes}
      \footnotesize
      \item[$^p$] DeCoP was enabled for XCTFormer on this dataset.
    \end{tablenotes}
  \end{threeparttable}
\end{table}

\begin{table}[H]
  \caption{Results of the imputation task across datasets, evaluated across five seeds (2021-2025). We randomly mask $\{12.5\%, 25\%, 37.5\%, 50\%\}$ of the time points; the final results are averaged across these four masking ratios.}
  \label{tab:imputation_results_seed}
  \vskip 0.05in
  \centering
  \begin{threeparttable}
  \begin{small}
  \renewcommand{\multirowsetup}{\centering}
  \setlength{\tabcolsep}{5pt}
  \begin{tabular}{l|cc|c}
    \toprule
    Model & \multicolumn{2}{c|}{XCTFormer (Ours)} & Confidence Score  \\
    \cmidrule(lr){0-1}\cmidrule(lr){2-3}\cmidrule(lr){4-4}
    Dataset & MSE & MAE & Score \%\\
    \midrule
    ETTh1 & $0.090 \pm 0.002$ & $0.204 \pm 0.003$ &  98.0 \% \\
    ETTh2 & $0.052 \pm 0.013$ & $0.153 \pm 0.020$ &  80.9 \% \\
    ETTm1 & $0.031 \pm 0.004$ & $0.116 \pm 0.007$ &  90.2 \% \\
    ETTm2 & $0.026 \pm 0.003$ & $0.097 \pm 0.007$ &  90.8 \% \\
    ETT(Avg) & $0.049 \pm 0.006$ & $0.143 \pm 0.009$ &  91.1 \% \\
    ECL$^p$ & $0.051 \pm 0.008$ & $0.149 \pm 0.014$ &  87.6 \% \\
    Weather & $0.031 \pm 5.26e-04$ & $0.049 \pm 0.002$ &  96.9 \% \\
    \bottomrule
  \end{tabular}
    \end{small}
    \begin{tablenotes}
      \footnotesize
      \item[$^p$] DeCoP was enabled for XCTFormer on this dataset.
    \end{tablenotes}
  \end{threeparttable}
\end{table}

\begin{table}[H]
\caption{Results for the anomaly detection task (P, R, and F1 are precision, recall, and F1-score in \%), evaluated across five seeds (2021-2025).}
  \label{tab:anomaly_results_seed}
  \vskip 0.05in
  \centering
  \begin{threeparttable}
  \begin{small}
  \renewcommand{\multirowsetup}{\centering}
  \setlength{\tabcolsep}{5pt}
  \begin{tabular}{l|ccc|c}
    \toprule
    Model & \multicolumn{3}{c|}{XCTFormer (Ours)} & Confidence Score  \\
    \cmidrule(lr){0-1}\cmidrule(lr){2-4}\cmidrule(lr){5-5}
    Dataset & Precision  & Recall & F1 & Score \% \\
    \midrule
    MSL & $87.84 \pm 1.73$ & $66.66 \pm 3.92$ & $75.77 \pm 3.18$ &  96.0\% \\
    PSM & $98.31 \pm 0.09$ & $93.05 \pm 0.57$ & $95.61 \pm 0.34$ &  99.6\% \\
    SMAP & $91.81 \pm 1.78$ & $64.59 \pm 14.53$ & $75.22 \pm 10.43$ &  87.2\% \\
    SMD & $87.01 \pm 0.25$ & $81.99 \pm 1.15$ & $84.42 \pm 0.69$ &  99.2\% \\
    SWaT & $91.94 \pm 0.49$ & $92.01 \pm 1.32$ & $91.98 \pm 0.86$ &  99.0\% \\
    \bottomrule
  \end{tabular}
    \end{small}
  \end{threeparttable}
\end{table}

\subsection{Long-Term Forecasting Results}
To improve readability, we present only the averaged table for long-term forecasting in the main paper and provide the full results here.

\begin{table}[htbp]
  \caption{Long-term forecasting results comparison across multiple datasets and horizons. Synthetic$^\dagger$ is evaluated under multivariate-to-single (MS) setting while all other datasets use multivariate (M). We compare extensive competitive models under different prediction lengths. \emph{Avg} is averaged from all four prediction lengths, that $\{96, 192, 336, 720\} $.}\label{tab:long_term_forecasting_results}
  \vskip 0.05in
  \centering
  \resizebox{1.0\columnwidth}{!}{
  \begin{threeparttable}
  \begin{small}
  \renewcommand{\multirowsetup}{\centering}
  \setlength{\tabcolsep}{1pt}
  \begin{tabular}{c|c|cc|cc|cc|cc|cc|cc|cc|cc|cc|cc|cc|cc|cc|cc|}
    \toprule
    \multicolumn{2}{c}{\multirow{2}{*}{Models}} &
    \multicolumn{2}{c}{\rotatebox{0}{\scalebox{0.8}{\textbf{XCTFormer}}}} &
    \multicolumn{2}{c}{\rotatebox{0}{\scalebox{0.8}{TimeMixer++}}} &
    \multicolumn{2}{c}{\rotatebox{0}{\scalebox{0.8}{MTLinear}}} &
    \multicolumn{2}{c}{\rotatebox{0}{\scalebox{0.8}{Leddam}}} &
    \multicolumn{2}{c}{\rotatebox{0}{\scalebox{0.8}{TimeMixer}}} &
    \multicolumn{2}{c}{\rotatebox{0}{\scalebox{0.8}{iTransformer}}} &
    \multicolumn{2}{c}{\rotatebox{0}{\scalebox{0.8}{PatchTST}}} &
    \multicolumn{2}{c}{\rotatebox{0}{\scalebox{0.8}{Crossformer}}} &
    \multicolumn{2}{c}{\rotatebox{0}{\scalebox{0.8}{TiDE}}} &
    \multicolumn{2}{c}{\rotatebox{0}{\scalebox{0.8}{TimesNet}}} &
    \multicolumn{2}{c}{\rotatebox{0}{\scalebox{0.8}{DLinear}}} &
    \multicolumn{2}{c}{\rotatebox{0}{\scalebox{0.8}{SCINet}}} &
    \multicolumn{2}{c}{\rotatebox{0}{\scalebox{0.8}{FEDformer}}} &
    \multicolumn{2}{c}{\rotatebox{0}{\scalebox{0.8}{Autoformer}}} \\
    \multicolumn{2}{c}{} &
    \multicolumn{2}{c}{\scalebox{0.8}{(\textbf{Ours})}} &
    \multicolumn{2}{c}{\scalebox{0.8}{(ICLR 2025)}} &
    \multicolumn{2}{c}{\scalebox{0.8}{(AISTATS 2025)}} &
    \multicolumn{2}{c}{\scalebox{0.8}{(ICML 2024)}} &
    \multicolumn{2}{c}{\scalebox{0.8}{(ICLR 2024)}} &
    \multicolumn{2}{c}{\scalebox{0.8}{(ICLR 2024)}} &
    \multicolumn{2}{c}{\scalebox{0.8}{(ICLR 2023)}} &
    \multicolumn{2}{c}{\scalebox{0.8}{(ICLR 2023)}} &
    \multicolumn{2}{c}{\scalebox{0.8}{(TMLR 2023)}} &
    \multicolumn{2}{c}{\scalebox{0.8}{(ICLR 2023)}} &
    \multicolumn{2}{c}{\scalebox{0.8}{(AAAI 2023)}} &
    \multicolumn{2}{c}{\scalebox{0.8}{(NeurIPS 2022)}} &
    \multicolumn{2}{c}{\scalebox{0.8}{(ICML 2022)}} &
    \multicolumn{2}{c}{\scalebox{0.8}{(NeurIPS 2021)}} \\
    
    \cmidrule(lr){3-4} \cmidrule(lr){5-6} \cmidrule(lr){7-8} \cmidrule(lr){9-10} \cmidrule(lr){11-12} \cmidrule(lr){13-14} \cmidrule(lr){15-16} \cmidrule(lr){17-18} \cmidrule(lr){19-20} \cmidrule(lr){21-22} \cmidrule(lr){23-24} \cmidrule(lr){25-26} \cmidrule(lr){27-28} \cmidrule(lr){29-30}
    \multicolumn{2}{c}{Metric} & \scalebox{0.68}{MSE} & \scalebox{0.68}{MAE} & \scalebox{0.68}{MSE} & \scalebox{0.68}{MAE} & \scalebox{0.68}{MSE} & \scalebox{0.68}{MAE} & \scalebox{0.68}{MSE} & \scalebox{0.68}{MAE} & \scalebox{0.68}{MSE} & \scalebox{0.68}{MAE} & \scalebox{0.68}{MSE} & \scalebox{0.68}{MAE} & \scalebox{0.68}{MSE} & \scalebox{0.68}{MAE} & \scalebox{0.68}{MSE} & \scalebox{0.68}{MAE} & \scalebox{0.68}{MSE} & \scalebox{0.68}{MAE} & \scalebox{0.68}{MSE} & \scalebox{0.68}{MAE} & \scalebox{0.68}{MSE} & \scalebox{0.68}{MAE} & \scalebox{0.68}{MSE} & \scalebox{0.68}{MAE} & \scalebox{0.68}{MSE} & \scalebox{0.68}{MAE} & \scalebox{0.68}{MSE} & \scalebox{0.68}{MAE}\\
    \toprule
    \multirow{5}{*}{\rotatebox{90}{\scalebox{0.95}{ETTm1}}} & \scalebox{0.85}{96} & \boldres{\scalebox{0.85}{0.302}} & \secondres{\scalebox{0.85}{0.350}} & \secondres{\scalebox{0.85}{0.310}} & \boldres{\scalebox{0.85}{0.334}} & \scalebox{0.85}{0.337} & \scalebox{0.85}{0.363} & \scalebox{0.85}{0.319} & \scalebox{0.85}{0.359} & \scalebox{0.85}{0.320} & \scalebox{0.85}{0.357} & \scalebox{0.85}{0.334} & \scalebox{0.85}{0.368} & \scalebox{0.85}{0.352} & \scalebox{0.85}{0.374} & \scalebox{0.85}{0.404} & \scalebox{0.85}{0.426} & \scalebox{0.85}{0.364} & \scalebox{0.85}{0.387} & \scalebox{0.85}{0.338} & \scalebox{0.85}{0.375} & \scalebox{0.85}{0.346} & \scalebox{0.85}{0.374} & \scalebox{0.85}{0.418} & \scalebox{0.85}{0.438} & \scalebox{0.85}{0.379} & \scalebox{0.85}{0.419} & \scalebox{0.85}{0.505} & \scalebox{0.85}{0.475} \\
     & \scalebox{0.85}{192} & \secondres{\scalebox{0.85}{0.354}} & \scalebox{0.85}{0.382} & \boldres{\scalebox{0.85}{0.348}} & \boldres{\scalebox{0.85}{0.362}} & \scalebox{0.85}{0.379} & \scalebox{0.85}{0.387} & \scalebox{0.85}{0.369} & \scalebox{0.85}{0.383} & \scalebox{0.85}{0.361} & \secondres{\scalebox{0.85}{0.381}} & \scalebox{0.85}{0.390} & \scalebox{0.85}{0.393} & \scalebox{0.85}{0.374} & \scalebox{0.85}{0.387} & \scalebox{0.85}{0.450} & \scalebox{0.85}{0.451} & \scalebox{0.85}{0.398} & \scalebox{0.85}{0.404} & \scalebox{0.85}{0.374} & \scalebox{0.85}{0.387} & \scalebox{0.85}{0.382} & \scalebox{0.85}{0.391} & \scalebox{0.85}{0.439} & \scalebox{0.85}{0.450} & \scalebox{0.85}{0.426} & \scalebox{0.85}{0.441} & \scalebox{0.85}{0.553} & \scalebox{0.85}{0.496} \\
     & \scalebox{0.85}{336} & \secondres{\scalebox{0.85}{0.385}} & \scalebox{0.85}{0.404} & \boldres{\scalebox{0.85}{0.376}} & \boldres{\scalebox{0.85}{0.391}} & \scalebox{0.85}{0.412} & \scalebox{0.85}{0.409} & \scalebox{0.85}{0.394} & \secondres{\scalebox{0.85}{0.402}} & \scalebox{0.85}{0.390} & \scalebox{0.85}{0.404} & \scalebox{0.85}{0.426} & \scalebox{0.85}{0.420} & \scalebox{0.85}{0.421} & \scalebox{0.85}{0.414} & \scalebox{0.85}{0.532} & \scalebox{0.85}{0.515} & \scalebox{0.85}{0.428} & \scalebox{0.85}{0.425} & \scalebox{0.85}{0.410} & \scalebox{0.85}{0.411} & \scalebox{0.85}{0.415} & \scalebox{0.85}{0.415} & \scalebox{0.85}{0.490} & \scalebox{0.85}{0.485} & \scalebox{0.85}{0.445} & \scalebox{0.85}{0.459} & \scalebox{0.85}{0.621} & \scalebox{0.85}{0.537} \\
     & \scalebox{0.85}{720} & \boldres{\scalebox{0.85}{0.435}} & \secondres{\scalebox{0.85}{0.433}} & \secondres{\scalebox{0.85}{0.440}} & \boldres{\scalebox{0.85}{0.423}} & \scalebox{0.85}{0.468} & \scalebox{0.85}{0.443} & \scalebox{0.85}{0.460} & \scalebox{0.85}{0.442} & \scalebox{0.85}{0.454} & \scalebox{0.85}{0.441} & \scalebox{0.85}{0.491} & \scalebox{0.85}{0.459} & \scalebox{0.85}{0.462} & \scalebox{0.85}{0.449} & \scalebox{0.85}{0.666} & \scalebox{0.85}{0.589} & \scalebox{0.85}{0.487} & \scalebox{0.85}{0.461} & \scalebox{0.85}{0.478} & \scalebox{0.85}{0.450} & \scalebox{0.85}{0.473} & \scalebox{0.85}{0.451} & \scalebox{0.85}{0.595} & \scalebox{0.85}{0.550} & \scalebox{0.85}{0.543} & \scalebox{0.85}{0.490} & \scalebox{0.85}{0.671} & \scalebox{0.85}{0.561} \\
    \midrule
     & \scalebox{0.85}{Avg} & \secondres{\scalebox{0.85}{0.369}} & \secondres{\scalebox{0.85}{0.392}} & \boldres{\scalebox{0.85}{0.368}} & \boldres{\scalebox{0.85}{0.378}} & \scalebox{0.85}{0.399} & \scalebox{0.85}{0.401} & \scalebox{0.85}{0.385} & \scalebox{0.85}{0.397} & \scalebox{0.85}{0.381} & \scalebox{0.85}{0.396} & \scalebox{0.85}{0.410} & \scalebox{0.85}{0.410} & \scalebox{0.85}{0.402} & \scalebox{0.85}{0.406} & \scalebox{0.85}{0.513} & \scalebox{0.85}{0.495} & \scalebox{0.85}{0.419} & \scalebox{0.85}{0.419} & \scalebox{0.85}{0.400} & \scalebox{0.85}{0.406} & \scalebox{0.85}{0.404} & \scalebox{0.85}{0.408} & \scalebox{0.85}{0.485} & \scalebox{0.85}{0.481} & \scalebox{0.85}{0.448} & \scalebox{0.85}{0.452} & \scalebox{0.85}{0.588} & \scalebox{0.85}{0.517} \\
    \midrule
    \multirow{5}{*}{\rotatebox{90}{\scalebox{0.95}{ETTm2}}} & \scalebox{0.85}{96} & \boldres{\scalebox{0.85}{0.168}} & \secondres{\scalebox{0.85}{0.252}} & \secondres{\scalebox{0.85}{0.170}} & \boldres{\scalebox{0.85}{0.245}} & \scalebox{0.85}{0.175} & \scalebox{0.85}{0.254} & \scalebox{0.85}{0.176} & \scalebox{0.85}{0.257} & \scalebox{0.85}{0.175} & \scalebox{0.85}{0.258} & \scalebox{0.85}{0.180} & \scalebox{0.85}{0.264} & \scalebox{0.85}{0.183} & \scalebox{0.85}{0.270} & \scalebox{0.85}{0.287} & \scalebox{0.85}{0.366} & \scalebox{0.85}{0.207} & \scalebox{0.85}{0.305} & \scalebox{0.85}{0.187} & \scalebox{0.85}{0.267} & \scalebox{0.85}{0.193} & \scalebox{0.85}{0.293} & \scalebox{0.85}{0.286} & \scalebox{0.85}{0.377} & \scalebox{0.85}{0.203} & \scalebox{0.85}{0.287} & \scalebox{0.85}{0.255} & \scalebox{0.85}{0.339} \\
     & \scalebox{0.85}{192} & \secondres{\scalebox{0.85}{0.232}} & \secondres{\scalebox{0.85}{0.295}} & \boldres{\scalebox{0.85}{0.229}} & \boldres{\scalebox{0.85}{0.291}} & \scalebox{0.85}{0.240} & \scalebox{0.85}{0.296} & \scalebox{0.85}{0.243} & \scalebox{0.85}{0.303} & \scalebox{0.85}{0.237} & \scalebox{0.85}{0.299} & \scalebox{0.85}{0.250} & \scalebox{0.85}{0.309} & \scalebox{0.85}{0.255} & \scalebox{0.85}{0.314} & \scalebox{0.85}{0.414} & \scalebox{0.85}{0.492} & \scalebox{0.85}{0.290} & \scalebox{0.85}{0.364} & \scalebox{0.85}{0.249} & \scalebox{0.85}{0.309} & \scalebox{0.85}{0.284} & \scalebox{0.85}{0.361} & \scalebox{0.85}{0.399} & \scalebox{0.85}{0.445} & \scalebox{0.85}{0.269} & \scalebox{0.85}{0.328} & \scalebox{0.85}{0.281} & \scalebox{0.85}{0.340} \\
     & \scalebox{0.85}{336} & \boldres{\scalebox{0.85}{0.289}} & \boldres{\scalebox{0.85}{0.332}} & \scalebox{0.85}{0.303} & \scalebox{0.85}{0.343} & \scalebox{0.85}{0.301} & \secondres{\scalebox{0.85}{0.335}} & \scalebox{0.85}{0.303} & \scalebox{0.85}{0.341} & \secondres{\scalebox{0.85}{0.298}} & \scalebox{0.85}{0.340} & \scalebox{0.85}{0.311} & \scalebox{0.85}{0.348} & \scalebox{0.85}{0.309} & \scalebox{0.85}{0.347} & \scalebox{0.85}{0.597} & \scalebox{0.85}{0.542} & \scalebox{0.85}{0.377} & \scalebox{0.85}{0.422} & \scalebox{0.85}{0.321} & \scalebox{0.85}{0.351} & \scalebox{0.85}{0.382} & \scalebox{0.85}{0.429} & \scalebox{0.85}{0.637} & \scalebox{0.85}{0.591} & \scalebox{0.85}{0.325} & \scalebox{0.85}{0.366} & \scalebox{0.85}{0.339} & \scalebox{0.85}{0.372} \\
     & \scalebox{0.85}{720} & \secondres{\scalebox{0.85}{0.391}} & \secondres{\scalebox{0.85}{0.395}} & \boldres{\scalebox{0.85}{0.373}} & \scalebox{0.85}{0.399} & \scalebox{0.85}{0.402} & \boldres{\scalebox{0.85}{0.393}} & \scalebox{0.85}{0.400} & \scalebox{0.85}{0.398} & \secondres{\scalebox{0.85}{0.391}} & \scalebox{0.85}{0.396} & \scalebox{0.85}{0.412} & \scalebox{0.85}{0.407} & \scalebox{0.85}{0.412} & \scalebox{0.85}{0.404} & \scalebox{0.85}{1.730} & \scalebox{0.85}{1.042} & \scalebox{0.85}{0.558} & \scalebox{0.85}{0.524} & \scalebox{0.85}{0.408} & \scalebox{0.85}{0.403} & \scalebox{0.85}{0.558} & \scalebox{0.85}{0.525} & \scalebox{0.85}{0.960} & \scalebox{0.85}{0.735} & \scalebox{0.85}{0.421} & \scalebox{0.85}{0.415} & \scalebox{0.85}{0.433} & \scalebox{0.85}{0.432} \\
    \midrule
     & \scalebox{0.85}{Avg} & \secondres{\scalebox{0.85}{0.270}} & \boldres{\scalebox{0.85}{0.319}} & \boldres{\scalebox{0.85}{0.269}} & \secondres{\scalebox{0.85}{0.320}} & \scalebox{0.85}{0.279} & \secondres{\scalebox{0.85}{0.320}} & \scalebox{0.85}{0.280} & \scalebox{0.85}{0.325} & \scalebox{0.85}{0.275} & \scalebox{0.85}{0.323} & \scalebox{0.85}{0.288} & \scalebox{0.85}{0.332} & \scalebox{0.85}{0.290} & \scalebox{0.85}{0.334} & \scalebox{0.85}{0.757} & \scalebox{0.85}{0.611} & \scalebox{0.85}{0.358} & \scalebox{0.85}{0.404} & \scalebox{0.85}{0.291} & \scalebox{0.85}{0.333} & \scalebox{0.85}{0.354} & \scalebox{0.85}{0.402} & \scalebox{0.85}{0.571} & \scalebox{0.85}{0.537} & \scalebox{0.85}{0.304} & \scalebox{0.85}{0.349} & \scalebox{0.85}{0.327} & \scalebox{0.85}{0.371} \\
    \midrule
    \multirow{5}{*}{\rotatebox{90}{\scalebox{0.95}{ETTh1}}} & \scalebox{0.85}{96} & \scalebox{0.85}{0.389} & \scalebox{0.85}{0.400} & \boldres{\scalebox{0.85}{0.361}} & \scalebox{0.85}{0.403} & \scalebox{0.85}{0.386} & \boldres{\scalebox{0.85}{0.393}} & \scalebox{0.85}{0.377} & \secondres{\scalebox{0.85}{0.394}} & \secondres{\scalebox{0.85}{0.375}} & \scalebox{0.85}{0.400} & \scalebox{0.85}{0.386} & \scalebox{0.85}{0.405} & \scalebox{0.85}{0.460} & \scalebox{0.85}{0.447} & \scalebox{0.85}{0.423} & \scalebox{0.85}{0.448} & \scalebox{0.85}{0.479} & \scalebox{0.85}{0.464} & \scalebox{0.85}{0.384} & \scalebox{0.85}{0.402} & \scalebox{0.85}{0.397} & \scalebox{0.85}{0.412} & \scalebox{0.85}{0.654} & \scalebox{0.85}{0.599} & \scalebox{0.85}{0.395} & \scalebox{0.85}{0.424} & \scalebox{0.85}{0.449} & \scalebox{0.85}{0.459} \\
     & \scalebox{0.85}{192} & \scalebox{0.85}{0.440} & \scalebox{0.85}{0.429} & \boldres{\scalebox{0.85}{0.416}} & \scalebox{0.85}{0.441} & \scalebox{0.85}{0.439} & \boldres{\scalebox{0.85}{0.421}} & \secondres{\scalebox{0.85}{0.424}} & \scalebox{0.85}{0.422} & \scalebox{0.85}{0.429} & \boldres{\scalebox{0.85}{0.421}} & \scalebox{0.85}{0.441} & \scalebox{0.85}{0.512} & \scalebox{0.85}{0.477} & \scalebox{0.85}{0.429} & \scalebox{0.85}{0.471} & \scalebox{0.85}{0.474} & \scalebox{0.85}{0.525} & \scalebox{0.85}{0.492} & \scalebox{0.85}{0.436} & \scalebox{0.85}{0.429} & \scalebox{0.85}{0.446} & \scalebox{0.85}{0.441} & \scalebox{0.85}{0.719} & \scalebox{0.85}{0.631} & \scalebox{0.85}{0.469} & \scalebox{0.85}{0.470} & \scalebox{0.85}{0.500} & \scalebox{0.85}{0.482} \\
     & \scalebox{0.85}{336} & \scalebox{0.85}{0.479} & \scalebox{0.85}{0.447} & \boldres{\scalebox{0.85}{0.430}} & \boldres{\scalebox{0.85}{0.434}} & \scalebox{0.85}{0.476} & \secondres{\scalebox{0.85}{0.441}} & \secondres{\scalebox{0.85}{0.459}} & \scalebox{0.85}{0.442} & \scalebox{0.85}{0.484} & \scalebox{0.85}{0.458} & \scalebox{0.85}{0.487} & \scalebox{0.85}{0.458} & \scalebox{0.85}{0.546} & \scalebox{0.85}{0.496} & \scalebox{0.85}{0.570} & \scalebox{0.85}{0.546} & \scalebox{0.85}{0.565} & \scalebox{0.85}{0.515} & \scalebox{0.85}{0.491} & \scalebox{0.85}{0.469} & \scalebox{0.85}{0.489} & \scalebox{0.85}{0.467} & \scalebox{0.85}{0.778} & \scalebox{0.85}{0.659} & \scalebox{0.85}{0.530} & \scalebox{0.85}{0.499} & \scalebox{0.85}{0.521} & \scalebox{0.85}{0.496} \\
     & \scalebox{0.85}{720} & \scalebox{0.85}{0.490} & \scalebox{0.85}{0.468} & \secondres{\scalebox{0.85}{0.467}} & \boldres{\scalebox{0.85}{0.451}} & \scalebox{0.85}{0.472} & \scalebox{0.85}{0.460} & \boldres{\scalebox{0.85}{0.463}} & \secondres{\scalebox{0.85}{0.459}} & \scalebox{0.85}{0.498} & \scalebox{0.85}{0.482} & \scalebox{0.85}{0.503} & \scalebox{0.85}{0.491} & \scalebox{0.85}{0.544} & \scalebox{0.85}{0.517} & \scalebox{0.85}{0.653} & \scalebox{0.85}{0.621} & \scalebox{0.85}{0.594} & \scalebox{0.85}{0.558} & \scalebox{0.85}{0.521} & \scalebox{0.85}{0.500} & \scalebox{0.85}{0.513} & \scalebox{0.85}{0.510} & \scalebox{0.85}{0.836} & \scalebox{0.85}{0.699} & \scalebox{0.85}{0.598} & \scalebox{0.85}{0.544} & \scalebox{0.85}{0.514} & \scalebox{0.85}{0.512} \\
    \midrule
     & \scalebox{0.85}{Avg} & \scalebox{0.85}{0.450} & \scalebox{0.85}{0.436} & \boldres{\scalebox{0.85}{0.418}} & \scalebox{0.85}{0.432} & \scalebox{0.85}{0.443} & \boldres{\scalebox{0.85}{0.429}} & \secondres{\scalebox{0.85}{0.431}} & \boldres{\scalebox{0.85}{0.429}} & \scalebox{0.85}{0.447} & \scalebox{0.85}{0.440} & \scalebox{0.85}{0.454} & \scalebox{0.85}{0.467} & \scalebox{0.85}{0.507} & \scalebox{0.85}{0.472} & \scalebox{0.85}{0.529} & \scalebox{0.85}{0.522} & \scalebox{0.85}{0.541} & \scalebox{0.85}{0.507} & \scalebox{0.85}{0.458} & \scalebox{0.85}{0.450} & \scalebox{0.85}{0.461} & \scalebox{0.85}{0.458} & \scalebox{0.85}{0.747} & \scalebox{0.85}{0.647} & \scalebox{0.85}{0.498} & \scalebox{0.85}{0.484} & \scalebox{0.85}{0.496} & \scalebox{0.85}{0.487} \\
    \midrule
    \multirow{5}{*}{\rotatebox{90}{\scalebox{0.95}{ETTh2}}} & \scalebox{0.85}{96} & \scalebox{0.85}{0.295} & \scalebox{0.85}{0.342} & \boldres{\scalebox{0.85}{0.276}} & \boldres{\scalebox{0.85}{0.328}} & \secondres{\scalebox{0.85}{0.288}} & \secondres{\scalebox{0.85}{0.336}} & \scalebox{0.85}{0.292} & \scalebox{0.85}{0.343} & \scalebox{0.85}{0.289} & \scalebox{0.85}{0.341} & \scalebox{0.85}{0.297} & \scalebox{0.85}{0.349} & \scalebox{0.85}{0.308} & \scalebox{0.85}{0.355} & \scalebox{0.85}{0.745} & \scalebox{0.85}{0.584} & \scalebox{0.85}{0.400} & \scalebox{0.85}{0.440} & \scalebox{0.85}{0.340} & \scalebox{0.85}{0.374} & \scalebox{0.85}{0.340} & \scalebox{0.85}{0.394} & \scalebox{0.85}{0.707} & \scalebox{0.85}{0.621} & \scalebox{0.85}{0.358} & \scalebox{0.85}{0.397} & \scalebox{0.85}{0.346} & \scalebox{0.85}{0.388} \\
     & \scalebox{0.85}{192} & \scalebox{0.85}{0.370} & \scalebox{0.85}{0.393} & \boldres{\scalebox{0.85}{0.342}} & \boldres{\scalebox{0.85}{0.379}} & \scalebox{0.85}{0.375} & \secondres{\scalebox{0.85}{0.388}} & \secondres{\scalebox{0.85}{0.367}} & \scalebox{0.85}{0.389} & \scalebox{0.85}{0.372} & \scalebox{0.85}{0.392} & \scalebox{0.85}{0.380} & \scalebox{0.85}{0.400} & \scalebox{0.85}{0.393} & \scalebox{0.85}{0.405} & \scalebox{0.85}{0.877} & \scalebox{0.85}{0.656} & \scalebox{0.85}{0.528} & \scalebox{0.85}{0.509} & \scalebox{0.85}{0.402} & \scalebox{0.85}{0.414} & \scalebox{0.85}{0.482} & \scalebox{0.85}{0.479} & \scalebox{0.85}{0.860} & \scalebox{0.85}{0.689} & \scalebox{0.85}{0.429} & \scalebox{0.85}{0.439} & \scalebox{0.85}{0.456} & \scalebox{0.85}{0.452} \\
     & \scalebox{0.85}{336} & \scalebox{0.85}{0.402} & \scalebox{0.85}{0.417} & \boldres{\scalebox{0.85}{0.346}} & \boldres{\scalebox{0.85}{0.398}} & \scalebox{0.85}{0.412} & \scalebox{0.85}{0.423} & \scalebox{0.85}{0.412} & \scalebox{0.85}{0.424} & \secondres{\scalebox{0.85}{0.386}} & \secondres{\scalebox{0.85}{0.414}} & \scalebox{0.85}{0.428} & \scalebox{0.85}{0.432} & \scalebox{0.85}{0.427} & \scalebox{0.85}{0.436} & \scalebox{0.85}{1.043} & \scalebox{0.85}{0.731} & \scalebox{0.85}{0.643} & \scalebox{0.85}{0.571} & \scalebox{0.85}{0.452} & \scalebox{0.85}{0.452} & \scalebox{0.85}{0.591} & \scalebox{0.85}{0.541} & \scalebox{0.85}{1.000} & \scalebox{0.85}{0.744} & \scalebox{0.85}{0.496} & \scalebox{0.85}{0.487} & \scalebox{0.85}{0.482} & \scalebox{0.85}{0.486} \\
     & \scalebox{0.85}{720} & \secondres{\scalebox{0.85}{0.411}} & \secondres{\scalebox{0.85}{0.433}} & \boldres{\scalebox{0.85}{0.392}} & \boldres{\scalebox{0.85}{0.415}} & \scalebox{0.85}{0.418} & \scalebox{0.85}{0.440} & \scalebox{0.85}{0.419} & \scalebox{0.85}{0.438} & \scalebox{0.85}{0.412} & \scalebox{0.85}{0.434} & \scalebox{0.85}{0.427} & \scalebox{0.85}{0.445} & \scalebox{0.85}{0.436} & \scalebox{0.85}{0.450} & \scalebox{0.85}{1.104} & \scalebox{0.85}{0.763} & \scalebox{0.85}{0.874} & \scalebox{0.85}{0.679} & \scalebox{0.85}{0.462} & \scalebox{0.85}{0.468} & \scalebox{0.85}{0.839} & \scalebox{0.85}{0.661} & \scalebox{0.85}{1.249} & \scalebox{0.85}{0.838} & \scalebox{0.85}{0.463} & \scalebox{0.85}{0.474} & \scalebox{0.85}{0.515} & \scalebox{0.85}{0.511} \\
    \midrule
     & \scalebox{0.85}{Avg} & \scalebox{0.85}{0.369} & \scalebox{0.85}{0.396} & \boldres{\scalebox{0.85}{0.339}} & \boldres{\scalebox{0.85}{0.380}} & \scalebox{0.85}{0.373} & \scalebox{0.85}{0.397} & \scalebox{0.85}{0.372} & \scalebox{0.85}{0.398} & \secondres{\scalebox{0.85}{0.365}} & \secondres{\scalebox{0.85}{0.395}} & \scalebox{0.85}{0.383} & \scalebox{0.85}{0.407} & \scalebox{0.85}{0.391} & \scalebox{0.85}{0.411} & \scalebox{0.85}{0.942} & \scalebox{0.85}{0.683} & \scalebox{0.85}{0.611} & \scalebox{0.85}{0.550} & \scalebox{0.85}{0.414} & \scalebox{0.85}{0.427} & \scalebox{0.85}{0.563} & \scalebox{0.85}{0.519} & \scalebox{0.85}{0.954} & \scalebox{0.85}{0.723} & \scalebox{0.85}{0.436} & \scalebox{0.85}{0.449} & \scalebox{0.85}{0.450} & \scalebox{0.85}{0.459} \\
    \midrule
    \multirow{5}{*}{\rotatebox{90}{\scalebox{0.95}{Weather}}} & \scalebox{0.85}{96} & \boldres{\scalebox{0.85}{0.153}} & \boldres{\scalebox{0.85}{0.199}} & \secondres{\scalebox{0.85}{0.155}} & \scalebox{0.85}{0.205} & \scalebox{0.85}{0.159} & \scalebox{0.85}{0.211} & \scalebox{0.85}{0.156} & \secondres{\scalebox{0.85}{0.202}} & \scalebox{0.85}{0.163} & \scalebox{0.85}{0.209} & \scalebox{0.85}{0.174} & \scalebox{0.85}{0.214} & \scalebox{0.85}{0.186} & \scalebox{0.85}{0.227} & \scalebox{0.85}{0.195} & \scalebox{0.85}{0.271} & \scalebox{0.85}{0.202} & \scalebox{0.85}{0.261} & \scalebox{0.85}{0.172} & \scalebox{0.85}{0.220} & \scalebox{0.85}{0.195} & \scalebox{0.85}{0.252} & \scalebox{0.85}{0.221} & \scalebox{0.85}{0.306} & \scalebox{0.85}{0.217} & \scalebox{0.85}{0.296} & \scalebox{0.85}{0.266} & \scalebox{0.85}{0.336} \\
     & \scalebox{0.85}{192} & \boldres{\scalebox{0.85}{0.199}} & \boldres{\scalebox{0.85}{0.242}} & \secondres{\scalebox{0.85}{0.201}} & \secondres{\scalebox{0.85}{0.245}} & \scalebox{0.85}{0.202} & \scalebox{0.85}{0.252} & \scalebox{0.85}{0.207} & \scalebox{0.85}{0.250} & \scalebox{0.85}{0.208} & \scalebox{0.85}{0.250} & \scalebox{0.85}{0.221} & \scalebox{0.85}{0.254} & \scalebox{0.85}{0.234} & \scalebox{0.85}{0.265} & \scalebox{0.85}{0.209} & \scalebox{0.85}{0.277} & \scalebox{0.85}{0.242} & \scalebox{0.85}{0.298} & \scalebox{0.85}{0.219} & \scalebox{0.85}{0.261} & \scalebox{0.85}{0.237} & \scalebox{0.85}{0.295} & \scalebox{0.85}{0.261} & \scalebox{0.85}{0.340} & \scalebox{0.85}{0.276} & \scalebox{0.85}{0.336} & \scalebox{0.85}{0.307} & \scalebox{0.85}{0.367} \\
     & \scalebox{0.85}{336} & \scalebox{0.85}{0.257} & \secondres{\scalebox{0.85}{0.286}} & \boldres{\scalebox{0.85}{0.237}} & \boldres{\scalebox{0.85}{0.265}} & \scalebox{0.85}{0.259} & \scalebox{0.85}{0.294} & \scalebox{0.85}{0.262} & \scalebox{0.85}{0.291} & \secondres{\scalebox{0.85}{0.251}} & \scalebox{0.85}{0.287} & \scalebox{0.85}{0.278} & \scalebox{0.85}{0.296} & \scalebox{0.85}{0.284} & \scalebox{0.85}{0.301} & \scalebox{0.85}{0.273} & \scalebox{0.85}{0.332} & \scalebox{0.85}{0.287} & \scalebox{0.85}{0.335} & \scalebox{0.85}{0.280} & \scalebox{0.85}{0.306} & \scalebox{0.85}{0.282} & \scalebox{0.85}{0.331} & \scalebox{0.85}{0.309} & \scalebox{0.85}{0.378} & \scalebox{0.85}{0.339} & \scalebox{0.85}{0.380} & \scalebox{0.85}{0.359} & \scalebox{0.85}{0.395} \\
     & \scalebox{0.85}{720} & \scalebox{0.85}{0.339} & \secondres{\scalebox{0.85}{0.340}} & \boldres{\scalebox{0.85}{0.312}} & \boldres{\scalebox{0.85}{0.334}} & \secondres{\scalebox{0.85}{0.332}} & \scalebox{0.85}{0.346} & \scalebox{0.85}{0.343} & \scalebox{0.85}{0.343} & \scalebox{0.85}{0.339} & \scalebox{0.85}{0.341} & \scalebox{0.85}{0.358} & \scalebox{0.85}{0.347} & \scalebox{0.85}{0.356} & \scalebox{0.85}{0.349} & \scalebox{0.85}{0.379} & \scalebox{0.85}{0.401} & \scalebox{0.85}{0.351} & \scalebox{0.85}{0.386} & \scalebox{0.85}{0.365} & \scalebox{0.85}{0.359} & \scalebox{0.85}{0.345} & \scalebox{0.85}{0.382} & \scalebox{0.85}{0.377} & \scalebox{0.85}{0.427} & \scalebox{0.85}{0.403} & \scalebox{0.85}{0.428} & \scalebox{0.85}{0.419} & \scalebox{0.85}{0.428} \\
    \midrule
     & \scalebox{0.85}{Avg} & \secondres{\scalebox{0.85}{0.237}} & \secondres{\scalebox{0.85}{0.267}} & \boldres{\scalebox{0.85}{0.226}} & \boldres{\scalebox{0.85}{0.262}} & \scalebox{0.85}{0.238} & \scalebox{0.85}{0.276} & \scalebox{0.85}{0.242} & \scalebox{0.85}{0.272} & \scalebox{0.85}{0.240} & \scalebox{0.85}{0.272} & \scalebox{0.85}{0.258} & \scalebox{0.85}{0.278} & \scalebox{0.85}{0.265} & \scalebox{0.85}{0.285} & \scalebox{0.85}{0.264} & \scalebox{0.85}{0.320} & \scalebox{0.85}{0.270} & \scalebox{0.85}{0.320} & \scalebox{0.85}{0.259} & \scalebox{0.85}{0.286} & \scalebox{0.85}{0.265} & \scalebox{0.85}{0.315} & \scalebox{0.85}{0.292} & \scalebox{0.85}{0.363} & \scalebox{0.85}{0.309} & \scalebox{0.85}{0.360} & \scalebox{0.85}{0.338} & \scalebox{0.85}{0.382} \\
    \midrule
    \multirow{5}{*}{\rotatebox{90}{\scalebox{0.95}{ECL\tnote{p}}}} & \scalebox{0.85}{96} & \secondres{\scalebox{0.85}{0.138}} & \scalebox{0.85}{0.237} & \boldres{\scalebox{0.85}{0.135}} & \boldres{\scalebox{0.85}{0.222}} & \scalebox{0.85}{0.183} & \scalebox{0.85}{0.265} & \scalebox{0.85}{0.141} & \secondres{\scalebox{0.85}{0.235}} & \scalebox{0.85}{0.153} & \scalebox{0.85}{0.247} & \scalebox{0.85}{0.148} & \scalebox{0.85}{0.240} & \scalebox{0.85}{0.190} & \scalebox{0.85}{0.296} & \scalebox{0.85}{0.219} & \scalebox{0.85}{0.314} & \scalebox{0.85}{0.237} & \scalebox{0.85}{0.329} & \scalebox{0.85}{0.168} & \scalebox{0.85}{0.272} & \scalebox{0.85}{0.210} & \scalebox{0.85}{0.302} & \scalebox{0.85}{0.247} & \scalebox{0.85}{0.345} & \scalebox{0.85}{0.193} & \scalebox{0.85}{0.308} & \scalebox{0.85}{0.201} & \scalebox{0.85}{0.317} \\
     & \scalebox{0.85}{192} & \scalebox{0.85}{0.164} & \scalebox{0.85}{0.261} & \boldres{\scalebox{0.85}{0.147}} & \boldres{\scalebox{0.85}{0.235}} & \scalebox{0.85}{0.183} & \scalebox{0.85}{0.268} & \secondres{\scalebox{0.85}{0.159}} & \secondres{\scalebox{0.85}{0.252}} & \scalebox{0.85}{0.166} & \scalebox{0.85}{0.256} & \scalebox{0.85}{0.162} & \scalebox{0.85}{0.253} & \scalebox{0.85}{0.199} & \scalebox{0.85}{0.304} & \scalebox{0.85}{0.231} & \scalebox{0.85}{0.322} & \scalebox{0.85}{0.236} & \scalebox{0.85}{0.330} & \scalebox{0.85}{0.184} & \scalebox{0.85}{0.322} & \scalebox{0.85}{0.210} & \scalebox{0.85}{0.305} & \scalebox{0.85}{0.257} & \scalebox{0.85}{0.355} & \scalebox{0.85}{0.201} & \scalebox{0.85}{0.315} & \scalebox{0.85}{0.222} & \scalebox{0.85}{0.334} \\
     & \scalebox{0.85}{336} & \secondres{\scalebox{0.85}{0.170}} & \secondres{\scalebox{0.85}{0.266}} & \boldres{\scalebox{0.85}{0.164}} & \boldres{\scalebox{0.85}{0.245}} & \scalebox{0.85}{0.196} & \scalebox{0.85}{0.283} & \scalebox{0.85}{0.173} & \scalebox{0.85}{0.268} & \scalebox{0.85}{0.185} & \scalebox{0.85}{0.277} & \scalebox{0.85}{0.178} & \scalebox{0.85}{0.269} & \scalebox{0.85}{0.217} & \scalebox{0.85}{0.319} & \scalebox{0.85}{0.246} & \scalebox{0.85}{0.337} & \scalebox{0.85}{0.249} & \scalebox{0.85}{0.344} & \scalebox{0.85}{0.198} & \scalebox{0.85}{0.300} & \scalebox{0.85}{0.223} & \scalebox{0.85}{0.319} & \scalebox{0.85}{0.269} & \scalebox{0.85}{0.369} & \scalebox{0.85}{0.214} & \scalebox{0.85}{0.329} & \scalebox{0.85}{0.231} & \scalebox{0.85}{0.443} \\
     & \scalebox{0.85}{720} & \boldres{\scalebox{0.85}{0.190}} & \boldres{\scalebox{0.85}{0.286}} & \scalebox{0.85}{0.212} & \scalebox{0.85}{0.310} & \scalebox{0.85}{0.231} & \scalebox{0.85}{0.317} & \secondres{\scalebox{0.85}{0.201}} & \secondres{\scalebox{0.85}{0.295}} & \scalebox{0.85}{0.225} & \scalebox{0.85}{0.310} & \scalebox{0.85}{0.225} & \scalebox{0.85}{0.317} & \scalebox{0.85}{0.258} & \scalebox{0.85}{0.352} & \scalebox{0.85}{0.280} & \scalebox{0.85}{0.363} & \scalebox{0.85}{0.284} & \scalebox{0.85}{0.373} & \scalebox{0.85}{0.220} & \scalebox{0.85}{0.320} & \scalebox{0.85}{0.258} & \scalebox{0.85}{0.350} & \scalebox{0.85}{0.299} & \scalebox{0.85}{0.390} & \scalebox{0.85}{0.246} & \scalebox{0.85}{0.355} & \scalebox{0.85}{0.254} & \scalebox{0.85}{0.361} \\
    \midrule
     & \scalebox{0.85}{Avg} & \secondres{\scalebox{0.85}{0.166}} & \secondres{\scalebox{0.85}{0.263}} & \boldres{\scalebox{0.85}{0.165}} & \boldres{\scalebox{0.85}{0.253}} & \scalebox{0.85}{0.198} & \scalebox{0.85}{0.283} & \scalebox{0.85}{0.168} & \secondres{\scalebox{0.85}{0.263}} & \scalebox{0.85}{0.182} & \scalebox{0.85}{0.273} & \scalebox{0.85}{0.178} & \scalebox{0.85}{0.270} & \scalebox{0.85}{0.216} & \scalebox{0.85}{0.318} & \scalebox{0.85}{0.244} & \scalebox{0.85}{0.334} & \scalebox{0.85}{0.252} & \scalebox{0.85}{0.344} & \scalebox{0.85}{0.193} & \scalebox{0.85}{0.304} & \scalebox{0.85}{0.225} & \scalebox{0.85}{0.319} & \scalebox{0.85}{0.268} & \scalebox{0.85}{0.365} & \scalebox{0.85}{0.213} & \scalebox{0.85}{0.327} & \scalebox{0.85}{0.227} & \scalebox{0.85}{0.364} \\
    \midrule
    \multirow{5}{*}{\rotatebox{90}{\scalebox{0.95}{Traffic\tnote{p}}}} & \scalebox{0.85}{96} & \scalebox{0.85}{0.402} & \scalebox{0.85}{0.269} & \boldres{\scalebox{0.85}{0.392}} & \boldres{\scalebox{0.85}{0.253}} & \scalebox{0.85}{0.647} & \scalebox{0.85}{0.383} & \scalebox{0.85}{0.426} & \scalebox{0.85}{0.276} & \scalebox{0.85}{0.462} & \scalebox{0.85}{0.285} & \secondres{\scalebox{0.85}{0.395}} & \secondres{\scalebox{0.85}{0.268}} & \scalebox{0.85}{0.526} & \scalebox{0.85}{0.347} & \scalebox{0.85}{0.644} & \scalebox{0.85}{0.429} & \scalebox{0.85}{0.805} & \scalebox{0.85}{0.493} & \scalebox{0.85}{0.593} & \scalebox{0.85}{0.321} & \scalebox{0.85}{0.650} & \scalebox{0.85}{0.396} & \scalebox{0.85}{0.788} & \scalebox{0.85}{0.499} & \scalebox{0.85}{0.587} & \scalebox{0.85}{0.366} & \scalebox{0.85}{0.613} & \scalebox{0.85}{0.388} \\
     & \scalebox{0.85}{192} & \scalebox{0.85}{0.424} & \scalebox{0.85}{0.281} & \boldres{\scalebox{0.85}{0.402}} & \boldres{\scalebox{0.85}{0.258}} & \scalebox{0.85}{0.594} & \scalebox{0.85}{0.359} & \scalebox{0.85}{0.458} & \scalebox{0.85}{0.289} & \scalebox{0.85}{0.473} & \scalebox{0.85}{0.296} & \secondres{\scalebox{0.85}{0.417}} & \secondres{\scalebox{0.85}{0.276}} & \scalebox{0.85}{0.522} & \scalebox{0.85}{0.332} & \scalebox{0.85}{0.665} & \scalebox{0.85}{0.431} & \scalebox{0.85}{0.756} & \scalebox{0.85}{0.474} & \scalebox{0.85}{0.617} & \scalebox{0.85}{0.336} & \scalebox{0.85}{0.598} & \scalebox{0.85}{0.370} & \scalebox{0.85}{0.789} & \scalebox{0.85}{0.505} & \scalebox{0.85}{0.604} & \scalebox{0.85}{0.373} & \scalebox{0.85}{0.616} & \scalebox{0.85}{0.382} \\
     & \scalebox{0.85}{336} & \scalebox{0.85}{0.444} & \scalebox{0.85}{0.291} & \boldres{\scalebox{0.85}{0.428}} & \boldres{\scalebox{0.85}{0.263}} & \scalebox{0.85}{0.601} & \scalebox{0.85}{0.362} & \scalebox{0.85}{0.486} & \scalebox{0.85}{0.297} & \scalebox{0.85}{0.498} & \scalebox{0.85}{0.296} & \secondres{\scalebox{0.85}{0.433}} & \secondres{\scalebox{0.85}{0.283}} & \scalebox{0.85}{0.517} & \scalebox{0.85}{0.334} & \scalebox{0.85}{0.674} & \scalebox{0.85}{0.420} & \scalebox{0.85}{0.762} & \scalebox{0.85}{0.477} & \scalebox{0.85}{0.629} & \scalebox{0.85}{0.336} & \scalebox{0.85}{0.605} & \scalebox{0.85}{0.373} & \scalebox{0.85}{0.797} & \scalebox{0.85}{0.508} & \scalebox{0.85}{0.621} & \scalebox{0.85}{0.383} & \scalebox{0.85}{0.622} & \scalebox{0.85}{0.337} \\
     & \scalebox{0.85}{720} & \scalebox{0.85}{0.472} & \scalebox{0.85}{0.307} & \boldres{\scalebox{0.85}{0.441}} & \boldres{\scalebox{0.85}{0.282}} & \scalebox{0.85}{0.640} & \scalebox{0.85}{0.382} & \scalebox{0.85}{0.498} & \scalebox{0.85}{0.313} & \scalebox{0.85}{0.506} & \scalebox{0.85}{0.313} & \secondres{\scalebox{0.85}{0.467}} & \secondres{\scalebox{0.85}{0.302}} & \scalebox{0.85}{0.552} & \scalebox{0.85}{0.352} & \scalebox{0.85}{0.683} & \scalebox{0.85}{0.424} & \scalebox{0.85}{0.719} & \scalebox{0.85}{0.449} & \scalebox{0.85}{0.640} & \scalebox{0.85}{0.350} & \scalebox{0.85}{0.645} & \scalebox{0.85}{0.394} & \scalebox{0.85}{0.841} & \scalebox{0.85}{0.523} & \scalebox{0.85}{0.626} & \scalebox{0.85}{0.382} & \scalebox{0.85}{0.660} & \scalebox{0.85}{0.408} \\
    \midrule
     & \scalebox{0.85}{Avg} & \scalebox{0.85}{0.435} & \scalebox{0.85}{0.287} & \boldres{\scalebox{0.85}{0.416}} & \boldres{\scalebox{0.85}{0.264}} & \scalebox{0.85}{0.621} & \scalebox{0.85}{0.372} & \scalebox{0.85}{0.467} & \scalebox{0.85}{0.294} & \scalebox{0.85}{0.485} & \scalebox{0.85}{0.297} & \secondres{\scalebox{0.85}{0.428}} & \secondres{\scalebox{0.85}{0.282}} & \scalebox{0.85}{0.529} & \scalebox{0.85}{0.341} & \scalebox{0.85}{0.667} & \scalebox{0.85}{0.426} & \scalebox{0.85}{0.760} & \scalebox{0.85}{0.473} & \scalebox{0.85}{0.620} & \scalebox{0.85}{0.336} & \scalebox{0.85}{0.625} & \scalebox{0.85}{0.383} & \scalebox{0.85}{0.804} & \scalebox{0.85}{0.509} & \scalebox{0.85}{0.609} & \scalebox{0.85}{0.376} & \scalebox{0.85}{0.628} & \scalebox{0.85}{0.379} \\
    \midrule
    \multirow{5}{*}{\rotatebox{90}{\scalebox{0.95}{Synthetic$^\dagger$}}} & \scalebox{0.85}{96} & \scalebox{0.85}{0.038} & \scalebox{0.85}{0.151} & \scalebox{0.85}{--} & \scalebox{0.85}{--} & \scalebox{0.85}{0.670} & \scalebox{0.85}{0.640} & \scalebox{0.85}{0.043} & \scalebox{0.85}{0.159} & \secondres{\scalebox{0.85}{0.025}} & \secondres{\scalebox{0.85}{0.123}} & \scalebox{0.85}{0.217} & \scalebox{0.85}{0.362} & \scalebox{0.85}{0.222} & \scalebox{0.85}{0.365} & \boldres{\scalebox{0.85}{0.018}} & \boldres{\scalebox{0.85}{0.106}} & \scalebox{0.85}{0.836} & \scalebox{0.85}{0.746} & \scalebox{0.85}{0.554} & \scalebox{0.85}{0.597} & \scalebox{0.85}{0.516} & \scalebox{0.85}{0.577} & \scalebox{0.85}{0.130} & \scalebox{0.85}{0.277} & \scalebox{0.85}{1.537} & \scalebox{0.85}{1.006} & \scalebox{0.85}{0.573} & \scalebox{0.85}{0.605} \\
     & \scalebox{0.85}{192} & \secondres{\scalebox{0.85}{0.043}} & \secondres{\scalebox{0.85}{0.161}} & \scalebox{0.85}{--} & \scalebox{0.85}{--} & \scalebox{0.85}{1.189} & \scalebox{0.85}{0.881} & \scalebox{0.85}{0.094} & \scalebox{0.85}{0.235} & \scalebox{0.85}{0.070} & \scalebox{0.85}{0.204} & \scalebox{0.85}{0.236} & \scalebox{0.85}{0.375} & \scalebox{0.85}{0.319} & \scalebox{0.85}{0.442} & \boldres{\scalebox{0.85}{0.026}} & \boldres{\scalebox{0.85}{0.126}} & \scalebox{0.85}{1.290} & \scalebox{0.85}{0.934} & \scalebox{0.85}{0.471} & \scalebox{0.85}{0.538} & \scalebox{0.85}{0.496} & \scalebox{0.85}{0.572} & \scalebox{0.85}{0.092} & \scalebox{0.85}{0.235} & \scalebox{0.85}{2.422} & \scalebox{0.85}{1.282} & \scalebox{0.85}{1.753} & \scalebox{0.85}{1.094} \\
     & \scalebox{0.85}{336} & \secondres{\scalebox{0.85}{0.045}} & \secondres{\scalebox{0.85}{0.165}} & \scalebox{0.85}{--} & \scalebox{0.85}{--} & \scalebox{0.85}{1.388} & \scalebox{0.85}{0.975} & \scalebox{0.85}{0.095} & \scalebox{0.85}{0.234} & \scalebox{0.85}{0.053} & \scalebox{0.85}{0.178} & \scalebox{0.85}{0.282} & \scalebox{0.85}{0.411} & \scalebox{0.85}{0.319} & \scalebox{0.85}{0.441} & \boldres{\scalebox{0.85}{0.027}} & \boldres{\scalebox{0.85}{0.129}} & \scalebox{0.85}{1.476} & \scalebox{0.85}{1.016} & \scalebox{0.85}{0.404} & \scalebox{0.85}{0.499} & \scalebox{0.85}{0.482} & \scalebox{0.85}{0.565} & \scalebox{0.85}{0.085} & \scalebox{0.85}{0.225} & \scalebox{0.85}{4.065} & \scalebox{0.85}{1.663} & \scalebox{0.85}{1.743} & \scalebox{0.85}{1.079} \\
     & \scalebox{0.85}{720} & \secondres{\scalebox{0.85}{0.035}} & \secondres{\scalebox{0.85}{0.144}} & \scalebox{0.85}{--} & \scalebox{0.85}{--} & \scalebox{0.85}{1.149} & \scalebox{0.85}{0.845} & \scalebox{0.85}{0.082} & \scalebox{0.85}{0.216} & \scalebox{0.85}{0.133} & \scalebox{0.85}{0.281} & \scalebox{0.85}{0.282} & \scalebox{0.85}{0.413} & \scalebox{0.85}{0.249} & \scalebox{0.85}{0.384} & \boldres{\scalebox{0.85}{0.031}} & \boldres{\scalebox{0.85}{0.137}} & \scalebox{0.85}{1.182} & \scalebox{0.85}{0.875} & \scalebox{0.85}{0.350} & \scalebox{0.85}{0.459} & \scalebox{0.85}{0.428} & \scalebox{0.85}{0.523} & \scalebox{0.85}{0.079} & \scalebox{0.85}{0.218} & \scalebox{0.85}{1.879} & \scalebox{0.85}{1.110} & \scalebox{0.85}{1.606} & \scalebox{0.85}{1.019} \\
    \midrule
     & \scalebox{0.85}{Avg} & \secondres{\scalebox{0.85}{0.040}} & \secondres{\scalebox{0.85}{0.155}} & \scalebox{0.85}{--} & \scalebox{0.85}{--} & \scalebox{0.85}{1.099} & \scalebox{0.85}{0.835} & \scalebox{0.85}{0.079} & \scalebox{0.85}{0.211} & \scalebox{0.85}{0.070} & \scalebox{0.85}{0.197} & \scalebox{0.85}{0.254} & \scalebox{0.85}{0.390} & \scalebox{0.85}{0.277} & \scalebox{0.85}{0.408} & \boldres{\scalebox{0.85}{0.025}} & \boldres{\scalebox{0.85}{0.124}} & \scalebox{0.85}{1.196} & \scalebox{0.85}{0.893} & \scalebox{0.85}{0.445} & \scalebox{0.85}{0.523} & \scalebox{0.85}{0.480} & \scalebox{0.85}{0.559} & \scalebox{0.85}{0.097} & \scalebox{0.85}{0.239} & \scalebox{0.85}{2.476} & \scalebox{0.85}{1.265} & \scalebox{0.85}{1.419} & \scalebox{0.85}{0.949} \\
    \cmidrule(lr){2-30}
    \midrule
     & \scalebox{0.85}{$1^{st}$ Count} & \scalebox{0.85}{7} & \scalebox{0.85}{4} & \scalebox{0.85}{20} & \scalebox{0.85}{21} & \scalebox{0.85}{0} & \scalebox{0.85}{3} & \scalebox{0.85}{1} & \scalebox{0.85}{0} & \scalebox{0.85}{0} & \scalebox{0.85}{1} & \scalebox{0.85}{0} & \scalebox{0.85}{0} & \scalebox{0.85}{0} & \scalebox{0.85}{0} & \scalebox{0.85}{4} & \scalebox{0.85}{4} & \scalebox{0.85}{0} & \scalebox{0.85}{0} & \scalebox{0.85}{0} & \scalebox{0.85}{0} & \scalebox{0.85}{0} & \scalebox{0.85}{0} & \scalebox{0.85}{0} & \scalebox{0.85}{0} & \scalebox{0.85}{0} & \scalebox{0.85}{0} & \scalebox{0.85}{0} & \scalebox{0.85}{0} \\
    \midrule
    \midrule
    \multicolumn{2}{c|}{\scalebox{0.85}{Avg FLOPs}} & \multicolumn{2}{c|}{\scalebox{0.85}{3.33E+09}} & \multicolumn{2}{c|}{\scalebox{0.85}{--}} & \multicolumn{2}{c|}{\scalebox{0.85}{7.59E+06}} & \multicolumn{2}{c|}{\scalebox{0.85}{8.64E+08}} & \multicolumn{2}{c|}{\scalebox{0.85}{8.72E+08}} & \multicolumn{2}{c|}{\scalebox{0.85}{1.34E+09}} & \multicolumn{2}{c|}{\scalebox{0.85}{8.28E+08}} & \multicolumn{2}{c|}{\scalebox{0.85}{2.31E+09}} & \multicolumn{2}{c|}{\scalebox{0.85}{1.31E+09}} & \multicolumn{2}{c|}{\scalebox{0.85}{1.23E+11}} & \multicolumn{2}{c|}{\scalebox{0.85}{1.01E+07}} & \multicolumn{2}{c|}{\scalebox{0.85}{1.41E+09}} & \multicolumn{2}{c|}{\scalebox{0.85}{2.01E+09}} & \multicolumn{2}{c|}{\scalebox{0.85}{5.26E+07}} \\
    \midrule
    \multicolumn{2}{c|}{\scalebox{0.85}{Avg Params}} & \multicolumn{2}{c|}{\scalebox{0.85}{5.08E+06}} & \multicolumn{2}{c|}{\scalebox{0.85}{--}} & \multicolumn{2}{c|}{\scalebox{0.85}{7.71E+05}} & \multicolumn{2}{c|}{\scalebox{0.85}{2.85E+06}} & \multicolumn{2}{c|}{\scalebox{0.85}{1.34E+05}} & \multicolumn{2}{c|}{\scalebox{0.85}{2.30E+06}} & \multicolumn{2}{c|}{\scalebox{0.85}{7.10E+05}} & \multicolumn{2}{c|}{\scalebox{0.85}{7.87E+06}} & \multicolumn{2}{c|}{\scalebox{0.85}{8.53E+06}} & \multicolumn{2}{c|}{\scalebox{0.85}{5.74E+07}} & \multicolumn{2}{c|}{\scalebox{0.85}{6.50E+04}} & \multicolumn{2}{c|}{\scalebox{0.85}{7.14E+07}} & \multicolumn{2}{c|}{\scalebox{0.85}{1.69E+07}} & \multicolumn{2}{c|}{\scalebox{0.85}{2.15E+05}} \\
    \bottomrule
  \end{tabular}
    \end{small}
  \begin{tablenotes}[flushleft]
    \footnotesize
    \item[$\dagger$] Evaluated under multivariate-to-single (MS) setting; hyperparameters chosen identical to ETTm1 for all models.
    \item[1] Reported MTLinear results reflect the per-dataset best of MTNLinear and MTDLinear.
    \item[p] DeCoP was enabled for XCTFormer on this dataset.
  \end{tablenotes}
  \end{threeparttable}
  }
\end{table}

\subsection{Imputation Full Results} \label{appendix:imputation_full_results}
To improve readability, we present only the averaged table for imputation in the main paper and provide the full per-mask-ratio results for XCTFormer in Table~\ref{tab:imputation_full_results}.
\begin{table}[t]
  \caption{Full imputation results for XCTFormer across all datasets and mask ratios. We randomly mask $\{12.5\%, 25\%, 37.5\%, 50\%\}$ of the time points in a time series of length 1024.}\label{tab:imputation_full_results}
  \centering
  \begin{threeparttable}
  \begin{small}
  \renewcommand\arraystretch{0.9}
  \setlength{\tabcolsep}{3.5pt}
  \begin{tabular}{l|cc|cc|cc|cc|cc}
    \toprule
    \multicolumn{1}{c}{\multirow{2}{*}{Dataset}} &
    \multicolumn{2}{c}{Mask 12.5\%} &
    \multicolumn{2}{c}{Mask 25\%} &
    \multicolumn{2}{c}{Mask 37.5\%} &
    \multicolumn{2}{c}{Mask 50\%} &
    \multicolumn{2}{c}{Avg} \\
    \cmidrule(lr){2-3} \cmidrule(lr){4-5} \cmidrule(lr){6-7} \cmidrule(lr){8-9} \cmidrule(lr){10-11}
    & MSE & MAE & MSE & MAE & MSE & MAE & MSE & MAE & MSE & MAE \\
    \midrule
    ETTh1       & 0.069 & 0.179 & 0.080 & 0.192 & 0.093 & 0.208 & 0.105 & 0.221 & 0.087 & 0.200 \\
    ETTh2       & 0.045 & 0.143 & 0.065 & 0.168 & 0.068 & 0.181 & 0.060 & 0.165 & 0.059 & 0.164 \\
    ETTm1       & 0.019 & 0.092 & 0.026 & 0.107 & 0.034 & 0.122 & 0.035 & 0.124 & 0.028 & 0.111 \\
    ETTm2       & 0.020 & 0.083 & 0.022 & 0.089 & 0.025 & 0.095 & 0.028 & 0.101 & 0.024 & 0.092 \\
    \midrule
    ETT(Avg)    & 0.038 & 0.124 & 0.048 & 0.139 & 0.055 & 0.152 & 0.057 & 0.153 & 0.050 & 0.141 \\
    \midrule
    ECL\tnote{p}& 0.037 & 0.124 & 0.043 & 0.135 & 0.049 & 0.146 & 0.056 & 0.157 & 0.046 & 0.140 \\
    Weather     & 0.029 & 0.047 & 0.031 & 0.048 & 0.033 & 0.049 & 0.036 & 0.052 & 0.032 & 0.049 \\
    \bottomrule
  \end{tabular}
  \end{small}
  \begin{tablenotes}[flushleft]
    \footnotesize
    \item[p] DeCoP was enabled for XCTFormer on this dataset.
  \end{tablenotes}
  \end{threeparttable}
\end{table}

\subsection{Anomaly Detection Full Results} \label{appendix:anomaly_detection_full_results}

To improve readability, we present only the averaged plot for anomaly detection in the main paper and
provide the full results in Table~\ref{tab:full_anomaly_results}.

\begin{table}[t]
  \caption{Full results for the anomaly detection task. The P, R, and F1 represent the precision, recall, and F1-score, (\%) respectively. F1-score is the harmonic mean of precision and recall. A higher value of P, R and F1 indicates a better performance. \textcolor{red}{Red} indicates highest F1 score, \textcolor{blue}{blue} indicates second highest F1 score.}
  \label{tab:full_anomaly_results}
  \vskip 0.05in
  \centering
  \begin{threeparttable}
  \begin{small}
  \renewcommand{\multirowsetup}{\centering}
  \setlength{\tabcolsep}{1.4pt}
  \begin{tabular}{lc|ccc|ccc|ccc|ccc|ccc|c}
    \toprule
    \multicolumn{2}{c}{\scalebox{0.8}{Datasets}} &
    \multicolumn{3}{c}{\scalebox{0.8}{\rotatebox{0}{SMD}}} &
    \multicolumn{3}{c}{\scalebox{0.8}{\rotatebox{0}{MSL}}} &
    \multicolumn{3}{c}{\scalebox{0.8}{\rotatebox{0}{SMAP}}} &
    \multicolumn{3}{c}{\scalebox{0.8}{\rotatebox{0}{SWaT}}} &
    \multicolumn{3}{c}{\scalebox{0.8}{\rotatebox{0}{PSM}}} & \scalebox{0.8}{Avg F1} \\
    \cmidrule(lr){3-5} \cmidrule(lr){6-8}\cmidrule(lr){9-11} \cmidrule(lr){12-14}\cmidrule(lr){15-17}
    \multicolumn{2}{c}{\scalebox{0.8}{Metrics}} & \scalebox{0.8}{P} & \scalebox{0.8}{R} & \scalebox{0.8}{F1} & \scalebox{0.8}{P} & \scalebox{0.8}{R} & \scalebox{0.8}{F1} & \scalebox{0.8}{P} & \scalebox{0.8}{R} & \scalebox{0.8}{F1} & \scalebox{0.8}{P} & \scalebox{0.8}{R} & \scalebox{0.8}{F1} & \scalebox{0.8}{P} & \scalebox{0.8}{R} & \scalebox{0.8}{F1} & \scalebox{0.8}{(\%)}\\
    \toprule
        \scalebox{0.85}{LSTM} &
        \scalebox{0.85}{\citeyearpar{hochreiter1997long}}
        & \scalebox{0.85}{78.52} & \scalebox{0.85}{65.47} & \scalebox{0.85}{71.41}
        & \scalebox{0.85}{78.04} & \scalebox{0.85}{86.22} & \scalebox{0.85}{81.93}
        & \scalebox{0.85}{91.06} & \scalebox{0.85}{57.49} & \scalebox{0.85}{70.48}
        & \scalebox{0.85}{78.06} & \scalebox{0.85}{91.72} & \scalebox{0.85}{84.34}
        & \scalebox{0.85}{69.24} & \scalebox{0.85}{99.53} & \scalebox{0.85}{81.67}
        & \scalebox{0.85}{77.97} \\
        \scalebox{0.85}{Transformer} &
        \scalebox{0.85}{\citeyearpar{vaswani2017attention}}
        & \scalebox{0.85}{83.58} & \scalebox{0.85}{76.13} & \scalebox{0.85}{79.56}
        & \scalebox{0.85}{71.57} & \scalebox{0.85}{87.37} & \scalebox{0.85}{78.68}
        & \scalebox{0.85}{89.37} & \scalebox{0.85}{57.12} & \scalebox{0.85}{69.70}
        & \scalebox{0.85}{68.84} & \scalebox{0.85}{96.53} & \scalebox{0.85}{80.37}
        & \scalebox{0.85}{62.75} & \scalebox{0.85}{96.56} & \scalebox{0.85}{76.07}
        & \scalebox{0.85}{76.88} \\
        \scalebox{0.85}{LogTrans} &
        \scalebox{0.85}{\citeyearpar{2019Enhancing}}
        & \scalebox{0.85}{83.46} & \scalebox{0.85}{70.13} & \scalebox{0.85}{76.21}
        & \scalebox{0.85}{73.05} & \scalebox{0.85}{87.37} & \scalebox{0.85}{79.57}
        & \scalebox{0.85}{89.15} & \scalebox{0.85}{57.59} & \scalebox{0.85}{69.97}
        & \scalebox{0.85}{68.67} & \scalebox{0.85}{97.32} & \scalebox{0.85}{80.52}
        & \scalebox{0.85}{63.06} & \scalebox{0.85}{98.00} & \scalebox{0.85}{76.74}
        & \scalebox{0.85}{76.60} \\
        \scalebox{0.85}{TCN} &
        \scalebox{0.85}{\citeyearpar{Franceschi2018TCN}}
        & \scalebox{0.85}{84.06} & \scalebox{0.85}{79.07} & \scalebox{0.85}{81.49}
        & \scalebox{0.85}{75.11} & \scalebox{0.85}{82.44} & \scalebox{0.85}{78.60}
        & \scalebox{0.85}{86.90} & \scalebox{0.85}{59.23} & \scalebox{0.85}{70.45}
        & \scalebox{0.85}{76.59} & \scalebox{0.85}{95.71} & \scalebox{0.85}{85.09}
        & \scalebox{0.85}{54.59} & \scalebox{0.85}{99.77} & \scalebox{0.85}{70.57}
        & \scalebox{0.85}{77.24} \\
        \scalebox{0.85}{Reformer} &
        \scalebox{0.85}{\citeyearpar{kitaev2020reformer}}
        & \scalebox{0.85}{82.58} & \scalebox{0.85}{69.24} & \scalebox{0.85}{75.32}
        & \scalebox{0.85}{85.51} & \scalebox{0.85}{83.31} & \scalebox{0.85}{84.40}
        & \scalebox{0.85}{90.91} & \scalebox{0.85}{57.44} & \scalebox{0.85}{70.40}
        & \scalebox{0.85}{72.50} & \scalebox{0.85}{96.53} & \scalebox{0.85}{82.80}
        & \scalebox{0.85}{59.93} & \scalebox{0.85}{95.38} & \scalebox{0.85}{73.61}
        & \scalebox{0.85}{77.31} \\
        \scalebox{0.85}{Informer} &
        \scalebox{0.85}{\citeyearpar{zhou2021informer}}
        & \scalebox{0.85}{86.60} & \scalebox{0.85}{77.23} & \scalebox{0.85}{81.65}
        & \scalebox{0.85}{81.77} & \scalebox{0.85}{86.48} & \scalebox{0.85}{84.06}
        & \scalebox{0.85}{90.11} & \scalebox{0.85}{57.13} & \scalebox{0.85}{69.92}
        & \scalebox{0.85}{70.29} & \scalebox{0.85}{96.75} & \scalebox{0.85}{81.43}
        & \scalebox{0.85}{64.27} & \scalebox{0.85}{96.33} & \scalebox{0.85}{77.10}
        & \scalebox{0.85}{78.83} \\
        \scalebox{0.85}{Anomaly$^\ast$} &
        \scalebox{0.85}{\citeyearpar{xu2021anomaly}}
        & \scalebox{0.85}{88.91} & \scalebox{0.85}{82.23} & \scalebox{0.85}{85.49}
        & \scalebox{0.85}{79.61} & \scalebox{0.85}{87.37} & \scalebox{0.85}{83.31}
        & \scalebox{0.85}{91.85} & \scalebox{0.85}{58.11} & \scalebox{0.85}{71.18}
        & \scalebox{0.85}{72.51} & \scalebox{0.85}{97.32} & \scalebox{0.85}{83.10}
        & \scalebox{0.85}{68.35} & \scalebox{0.85}{94.72} & \scalebox{0.85}{79.40}
        & \scalebox{0.85}{80.50} \\
        \scalebox{0.85}{Pyraformer} &
        \scalebox{0.85}{\citeyearpar{liu2021pyraformer}}
        & \scalebox{0.85}{85.61} & \scalebox{0.85}{80.61} & \scalebox{0.85}{83.04}
        & \scalebox{0.85}{83.81} & \scalebox{0.85}{85.93} & \scalebox{0.85}{84.86}
        & \scalebox{0.85}{92.54} & \scalebox{0.85}{57.71} & \scalebox{0.85}{71.09}
        & \scalebox{0.85}{87.92} & \scalebox{0.85}{96.00} & \scalebox{0.85}{91.78}
        & \scalebox{0.85}{71.67} & \scalebox{0.85}{96.02} & \scalebox{0.85}{82.08}
        & \scalebox{0.85}{82.57} \\
        \scalebox{0.85}{Autoformer} &
        \scalebox{0.85}{\citeyearpar{wu2021autoformer}}
        & \scalebox{0.85}{88.06} & \scalebox{0.85}{82.35} & \scalebox{0.85}{85.11}
        & \scalebox{0.85}{77.27} & \scalebox{0.85}{80.92} & \scalebox{0.85}{79.05}
        & \scalebox{0.85}{90.40} & \scalebox{0.85}{58.62} & \scalebox{0.85}{71.12}
        & \scalebox{0.85}{89.85} & \scalebox{0.85}{95.81} & \scalebox{0.85}{92.74}
        & \scalebox{0.85}{99.08} & \scalebox{0.85}{88.15} & \scalebox{0.85}{93.29}
        & \scalebox{0.85}{84.26} \\
        \scalebox{0.85}{LSSL} &
        \scalebox{0.85}{\citeyearpar{gu2022efficiently}}
        & \scalebox{0.85}{78.51} & \scalebox{0.85}{65.32} & \scalebox{0.85}{71.31}
        & \scalebox{0.85}{77.55} & \scalebox{0.85}{88.18} & \scalebox{0.85}{82.53}
        & \scalebox{0.85}{89.43} & \scalebox{0.85}{53.43} & \scalebox{0.85}{66.90}
        & \scalebox{0.85}{79.05} & \scalebox{0.85}{93.72} & \scalebox{0.85}{85.76}
        & \scalebox{0.85}{66.02} & \scalebox{0.85}{92.93} & \scalebox{0.85}{77.20}
        & \scalebox{0.85}{76.74} \\
        \scalebox{0.85}{Stationary} &
        \scalebox{0.85}{\citeyearpar{liu2022non}}
        & \scalebox{0.85}{88.33} & \scalebox{0.85}{81.21} & \scalebox{0.85}{84.62}
        & \scalebox{0.85}{68.55} & \scalebox{0.85}{89.14} & \scalebox{0.85}{77.50}
        & \scalebox{0.85}{89.37} & \scalebox{0.85}{59.02} & \scalebox{0.85}{71.09}
        & \scalebox{0.85}{68.03} & \scalebox{0.85}{96.75} & \scalebox{0.85}{79.88}
        & \scalebox{0.85}{97.82} & \scalebox{0.85}{96.76} & \scalebox{0.85}{97.29}
        & \scalebox{0.85}{82.08} \\
        \scalebox{0.85}{DLinear} &
        \scalebox{0.85}{\citeyearpar{zeng2023transformers}}
        & \scalebox{0.85}{83.62} & \scalebox{0.85}{71.52} & \scalebox{0.85}{77.10}
        & \scalebox{0.85}{84.34} & \scalebox{0.85}{85.42} & \scalebox{0.85}{84.88}
        & \scalebox{0.85}{92.32} & \scalebox{0.85}{55.41} & \scalebox{0.85}{69.26}
        & \scalebox{0.85}{80.91} & \scalebox{0.85}{95.30} & \scalebox{0.85}{87.52}
        & \scalebox{0.85}{98.28} & \scalebox{0.85}{89.26} & \scalebox{0.85}{93.55}
        & \scalebox{0.85}{82.46} \\
        \scalebox{0.85}{ETSformer} &
        \scalebox{0.85}{\citeyearpar{woo2022etsformer}}
        & \scalebox{0.85}{87.44} & \scalebox{0.85}{79.23} & \scalebox{0.85}{83.13}
        & \scalebox{0.85}{85.13} & \scalebox{0.85}{84.93} & \scalebox{0.85}{85.03}
        & \scalebox{0.85}{92.25} & \scalebox{0.85}{55.75} & \scalebox{0.85}{69.50}
        & \scalebox{0.85}{90.02} & \scalebox{0.85}{80.36} & \scalebox{0.85}{84.91}
        & \scalebox{0.85}{99.31} & \scalebox{0.85}{85.28} & \scalebox{0.85}{91.76}
        & \scalebox{0.85}{82.87} \\
        \scalebox{0.85}{LightTS} &
        \scalebox{0.85}{\citeyearpar{lightts}}
        & \scalebox{0.85}{87.10} & \scalebox{0.85}{78.42} & \scalebox{0.85}{82.53}
        & \scalebox{0.85}{82.40} & \scalebox{0.85}{75.78} & \scalebox{0.85}{78.95}
        & \scalebox{0.85}{92.58} & \scalebox{0.85}{55.27} & \scalebox{0.85}{69.21}
        & \scalebox{0.85}{91.98} & \scalebox{0.85}{94.72} & \scalebox{0.85}{\textcolor{blue}{93.33}}
        & \scalebox{0.85}{98.37} & \scalebox{0.85}{95.97} & \scalebox{0.85}{97.15}
        & \scalebox{0.85}{84.23} \\
        \scalebox{0.85}{FEDformer} &
        \scalebox{0.85}{\citeyearpar{zhou2022fedformer}}
        & \scalebox{0.85}{87.95} & \scalebox{0.85}{82.39} & \scalebox{0.85}{85.08}
        & \scalebox{0.85}{77.14} & \scalebox{0.85}{80.07} & \scalebox{0.85}{78.57}
        & \scalebox{0.85}{90.47} & \scalebox{0.85}{58.10} & \scalebox{0.85}{70.76}
        & \scalebox{0.85}{90.17} & \scalebox{0.85}{96.42} & \scalebox{0.85}{93.19}
        & \scalebox{0.85}{97.31} & \scalebox{0.85}{97.16} & \scalebox{0.85}{97.23}
        & \scalebox{0.85}{84.97} \\
        \scalebox{0.85}{TimesNet} &
        \scalebox{0.85}{\citeyearpar{wu2022timesnet}}
        & \scalebox{0.85}{88.66} & \scalebox{0.85}{83.14} & \scalebox{0.85}{\textcolor{blue}{85.81}}
        & \scalebox{0.85}{83.92} & \scalebox{0.85}{86.42} & \scalebox{0.85}{\textcolor{blue}{85.15}}
        & \scalebox{0.85}{92.52} & \scalebox{0.85}{58.29} & \scalebox{0.85}{71.52}
        & \scalebox{0.85}{86.76} & \scalebox{0.85}{97.32} & \scalebox{0.85}{91.74}
        & \scalebox{0.85}{98.19} & \scalebox{0.85}{96.76} & \scalebox{0.85}{\textcolor{blue}{97.47}}
        & \scalebox{0.85}{86.34} \\
        \scalebox{0.85}{TiDE} &
        \scalebox{0.85}{\citeyearpar{das2023long}}
        & \scalebox{0.85}{76.00} & \scalebox{0.85}{63.00} & \scalebox{0.85}{68.91}
        & \scalebox{0.85}{84.00} & \scalebox{0.85}{60.00} & \scalebox{0.85}{70.18}
        & \scalebox{0.85}{88.00} & \scalebox{0.85}{50.00} & \scalebox{0.85}{64.00}
        & \scalebox{0.85}{98.00} & \scalebox{0.85}{63.00} & \scalebox{0.85}{76.73}
        & \scalebox{0.85}{93.00} & \scalebox{0.85}{92.00} & \scalebox{0.85}{92.50}
        & \scalebox{0.85}{74.46} \\
        \scalebox{0.85}{iTransformer} &
        \scalebox{0.85}{\citeyearpar{liu2023itransformer}}
        & \scalebox{0.85}{78.45} & \scalebox{0.85}{65.10} & \scalebox{0.85}{71.15}
        & \scalebox{0.85}{86.15} & \scalebox{0.85}{62.65} & \scalebox{0.85}{72.54}
        & \scalebox{0.85}{90.67} & \scalebox{0.85}{52.96} & \scalebox{0.85}{66.87}
        & \scalebox{0.85}{99.96} & \scalebox{0.85}{65.55} & \scalebox{0.85}{79.18}
        & \scalebox{0.85}{95.65} & \scalebox{0.85}{94.69} & \scalebox{0.85}{95.17}
        & \scalebox{0.85}{76.98} \\
        \scalebox{0.85}{TimesMixer++} &
        \scalebox{0.85}{\citeyearpar{wang2024timemixer++}}
        & \scalebox{0.85}{88.59} & \scalebox{0.85}{84.50} & \scalebox{0.85}{\textcolor{red}{86.50}}
        & \scalebox{0.85}{89.73} & \scalebox{0.85}{82.23} & \scalebox{0.85}{\textcolor{red}{85.82}}
        & \scalebox{0.85}{93.47} & \scalebox{0.85}{60.02} & \scalebox{0.85}{\textcolor{blue}{73.10}}
        & \scalebox{0.85}{92.96} & \scalebox{0.85}{94.33} & \scalebox{0.85}{\textcolor{red}{94.64}}
        & \scalebox{0.85}{98.33} & \scalebox{0.85}{96.90} & \scalebox{0.85}{\textcolor{red}{97.60}}
        & \scalebox{0.85}{\textcolor{blue}{87.47}} \\
        \scalebox{0.85}{XCTFormer} &
        \scalebox{0.85}{\textbf{(Ours)}}
        & \scalebox{0.85}{86.94} & \scalebox{0.85}{81.64} & \scalebox{0.85}{84.21}
        & \scalebox{0.85}{89.46} & \scalebox{0.85}{70.81} & \scalebox{0.85}{79.05}
        & \scalebox{0.85}{93.79} & \scalebox{0.85}{80.57} & \scalebox{0.85}{\textcolor{red}{86.68}}
        & \scalebox{0.85}{92.25} & \scalebox{0.85}{92.96} & \scalebox{0.85}{92.60}
        & \scalebox{0.85}{98.26} & \scalebox{0.85}{92.52} & \scalebox{0.85}{95.30}
        & \scalebox{0.85}{\textcolor{red}{87.57}} \\
        \bottomrule
    \end{tabular}
    \begin{tablenotes}
        \item $\ast$  The original paper of Anomaly Transformer \citep{xu2021anomaly} adopts the temporal association and reconstruction error as a joint anomaly criterion. For fair comparisons, we only use reconstruction error here.
    \end{tablenotes}
    \end{small}
  \end{threeparttable}
  \vspace{-10pt}
\end{table}

\subsection{Computational Complexity}\label{appendix:flops_params}

We report computational complexity in terms of floating-point operations (FLOPs) and trainable parameter counts. For a detailed description of the measurement methodology, see Section~\ref{appendix:computational_profiling}. Tables~\ref{tab:flops} and~\ref{tab:params} present the full per-dataset results.

\begin{table}[htbp]
  \caption{FLOPs (floating-point operations) per dataset, averaged over prediction lengths $\{96, 192, 336, 720\}$. Model and dataset ordering follows Table~\ref{tab:long_term_forecasting_results}. See Section~\ref{appendix:computational_profiling} for measurement methodology.}\label{tab:flops}
  \vskip 0.05in
  \centering
  \resizebox{1.0\columnwidth}{!}{
  \begin{threeparttable}
  \begin{small}
  \setlength{\tabcolsep}{5pt}
  \begin{tabular}{l|rrrrrrrr|r}
    \toprule
    \textbf{Model} & \scalebox{0.9}{ETTm1} & \scalebox{0.9}{ETTm2} & \scalebox{0.9}{ETTh1} & \scalebox{0.9}{ETTh2} & \scalebox{0.9}{Weather} & \scalebox{0.9}{ECL} & \scalebox{0.9}{Traffic} & \scalebox{0.9}{Synth.$^\dagger$} & \scalebox{0.9}{\textbf{Avg}} \\
    \midrule
    MTLinear        & 3.4E+05 & 3.4E+05 & 3.4E+05 & 3.4E+05 & 1.0E+06 & 1.6E+07 & 4.2E+07 & 3.4E+05 & 7.59E+06 \\
    DLinear         & 4.5E+05 & 4.5E+05 & 4.5E+05 & 4.5E+05 & 1.4E+06 & 2.1E+07 & 5.6E+07 & 4.5E+05 & 1.01E+07 \\
    Autoformer      & 2.4E+07 & 2.4E+07 & 2.4E+07 & 2.4E+07 & 2.7E+07 & 8.4E+07 & 1.9E+08 & 2.4E+07 & 5.26E+07 \\
    PatchTST        & 3.8E+07 & 3.8E+07 & 1.9E+06 & 1.9E+06 & 1.1E+08 & 1.7E+09 & 4.7E+09 & 3.8E+07 & 8.28E+08 \\
    Leddam          & 3.5E+07 & 4.7E+07 & 4.7E+07 & 1.2E+07 & 3.5E+07 & 1.6E+09 & 5.1E+09 & 3.5E+07 & 8.64E+08 \\
    TimeMixer       & 1.6E+07 & 3.2E+07 & 1.6E+07 & 1.6E+07 & 6.0E+07 & 9.2E+08 & 5.9E+09 & 1.6E+07 & 8.72E+08 \\
    TiDE            & 9.2E+07 & 7.7E+07 & 2.5E+07 & 1.1E+08 & 1.2E+08 & 6.0E+09 & 3.9E+09 & 9.2E+07 & 1.31E+09 \\
    iTransformer    & 2.9E+06 & 2.9E+06 & 1.0E+07 & 2.9E+06 & 1.3E+08 & 1.9E+09 & 8.7E+09 & 2.9E+06 & 1.34E+09 \\
    SCINet          & 3.4E+06 & 3.4E+06 & 7.3E+05 & 7.3E+05 & 4.8E+06 & 4.7E+09 & 6.6E+09 & 3.4E+06 & 1.41E+09 \\
    FEDformer       & 1.8E+09 & 1.8E+09 & 1.8E+09 & 1.8E+09 & 1.8E+09 & 2.2E+09 & 3.1E+09 & 1.8E+09 & 2.01E+09 \\
    Crossformer     & 1.2E+09 & N/A     & 9.7E+08 & N/A     & 2.5E+09 & 2.6E+09 & 5.4E+09 & 1.2E+09 & 2.31E+09 \\
    \textbf{XCTFormer} (Ours) & 3.3E+06 & 4.4E+07 & 4.0E+05 & 4.1E+06 & 4.9E+08 & 6.1E+09 & 2.0E+10 & 3.3E+06 & 3.33E+09 \\
    TimesNet        & 3.8E+09 & 1.9E+09 & 1.3E+09 & 2.5E+09 & 2.6E+09 & 3.2E+11 & 6.5E+11 & 3.8E+09 & 1.23E+11 \\
    \bottomrule
  \end{tabular}
  \end{small}
  \begin{tablenotes}[flushleft]
    \footnotesize
    \item[$\dagger$] Synthetic uses the multivariate-to-single (MS) setting with ETTm1 hyperparameters.
    \item Crossformer values on ETTh2 and ETTm2 are N/A due to configuration incompatibility. TimeMixer++ is excluded as its code is not publicly available.
  \end{tablenotes}
  \end{threeparttable}
  }
\end{table}

\begin{table}[htbp]
  \caption{Trainable parameter counts per dataset, averaged over prediction lengths $\{96, 192, 336, 720\}$. For MTLinear, values are the average of MTDLinear and MTNLinear. See Section~\ref{appendix:computational_profiling} for measurement methodology.}\label{tab:params}
  \vskip 0.05in
  \centering
  \resizebox{1.0\columnwidth}{!}{
  \begin{threeparttable}
  \begin{small}
  \setlength{\tabcolsep}{5pt}
  \begin{tabular}{l|rrrrrrrr|r}
    \toprule
    \textbf{Model} & \scalebox{0.9}{ETTm1} & \scalebox{0.9}{ETTm2} & \scalebox{0.9}{ETTh1} & \scalebox{0.9}{ETTh2} & \scalebox{0.9}{Weather} & \scalebox{0.9}{ECL} & \scalebox{0.9}{Traffic} & \scalebox{0.9}{Synth.$^\dagger$} & \scalebox{0.9}{\textbf{Avg}} \\
    \midrule
    DLinear         & 6.5E+04 & 6.5E+04 & 6.5E+04 & 6.5E+04 & 6.5E+04 & 6.5E+04 & 6.5E+04 & 6.5E+04 & 6.50E+04 \\
    TimeMixer       & 1.2E+05 & 1.2E+05 & 1.2E+05 & 1.2E+05 & 1.5E+05 & 1.5E+05 & 1.7E+05 & 1.2E+05 & 1.34E+05 \\
    Autoformer      & 1.2E+05 & 1.2E+05 & 1.2E+05 & 1.2E+05 & 1.3E+05 & 3.2E+05 & 6.7E+05 & 1.2E+05 & 2.15E+05 \\
    PatchTST        & 9.2E+05 & 9.2E+05 & 8.2E+04 & 8.2E+04 & 9.2E+05 & 9.2E+05 & 9.2E+05 & 9.2E+05 & 7.10E+05 \\
    MTLinear        & 2.0E+05 & 2.4E+05 & 2.0E+05 & 2.4E+05 & 5.9E+05 & 2.2E+06 & 2.3E+06 & 2.0E+05 & 7.71E+05 \\
    iTransformer    & 2.6E+05 & 2.6E+05 & 9.0E+05 & 2.6E+05 & 5.0E+06 & 5.0E+06 & 6.5E+06 & 2.6E+05 & 2.30E+06 \\
    Leddam          & 2.9E+06 & 3.8E+06 & 3.8E+06 & 1.3E+06 & 1.3E+06 & 3.4E+06 & 3.4E+06 & 2.9E+06 & 2.85E+06 \\
    \textbf{XCTFormer} (Ours) & 2.0E+05 & 1.3E+06 & 4.1E+04 & 1.7E+05 & 2.9E+06 & 3.8E+06 & 3.2E+07 & 2.0E+05 & 5.08E+06 \\
    Crossformer     & 1.1E+07 & N/A     & 1.1E+07 & N/A     & 1.1E+07 & 1.3E+06 & 1.9E+06 & 1.1E+07 & 7.87E+06 \\
    TiDE            & 9.6E+06 & 8.7E+06 & 2.4E+06 & 1.4E+07 & 4.3E+06 & 1.6E+07 & 3.4E+06 & 9.6E+06 & 8.53E+06 \\
    FEDformer       & 1.6E+07 & 1.6E+07 & 1.6E+07 & 1.6E+07 & 1.6E+07 & 1.8E+07 & 2.1E+07 & 1.6E+07 & 1.69E+07 \\
    TimesNet        & 2.7E+06 & 1.1E+06 & 6.3E+05 & 1.2E+06 & 1.2E+06 & 1.5E+08 & 3.0E+08 & 2.7E+06 & 5.74E+07 \\
    SCINet          & 2.8E+05 & 2.8E+05 & 7.2E+04 & 7.2E+04 & 3.1E+05 & 1.2E+08 & 4.5E+08 & 2.8E+05 & 7.14E+07 \\
    \bottomrule
  \end{tabular}
  \end{small}
  \begin{tablenotes}[flushleft]
    \footnotesize
    \item[$\dagger$] Synthetic uses the multivariate-to-single (MS) setting with ETTm1 hyperparameters.
    \item Crossformer values on ETTh2 and ETTm2 are N/A due to configuration incompatibility. TimeMixer++ is excluded as its code is not publicly available.
  \end{tablenotes}
  \end{threeparttable}
  }
\end{table}

\end{document}